\documentclass{article} 
\usepackage{iclr2027_conference,times}

\usepackage[utf8]{inputenc}
\usepackage[T1]{fontenc}
\usepackage{hyperref}
\usepackage{url}
\usepackage{booktabs}
\usepackage{amsfonts}
\usepackage{amsmath}
\usepackage{amssymb}
\usepackage{nicefrac}
\usepackage{microtype}
\usepackage{xcolor}
\usepackage{graphicx}
\usepackage{subcaption}
\usepackage{multirow}
\usepackage{array}
\usepackage{placeins}
\usepackage{makecell}

\title{Channel-Dependent State Space Model for Multivariate Time Series Forecasting}

\author{%
  \makebox[\textwidth][c]{%
    \begin{tabular}{c}
      Yu-Cheng Wu\textsuperscript{1},
      Fan-Keng Sun\textsuperscript{1},
      Li-Chun Lu\textsuperscript{2},
      Duane S. Boning\textsuperscript{1}
      \\
      \normalfont\textsuperscript{1}Massachusetts Institute of Technology,
      \normalfont\textsuperscript{2}National Taiwan University
    \end{tabular}%
  }%
}

\iclrfinalcopy 

\begin{document}

\maketitle
\lhead{}

\begin{abstract}
Multivariate time series forecasting (MTSF) is critical across many real-world domains. Existing deep learning approaches fall into two paradigms with distinct limitations: channel-independent (CI) methods ignore cross-variable dependencies and model only temporal dynamics, while channel-dependent (CD) methods consider both but typically rely on architectural compromises to mitigate overfitting and computational overhead. We therefore propose Chameleon, a CD state space model (SSM) that enables data-dependent, fine-grained interactions across variables while scaling linearly with the number of variables. By connecting selective SSMs with the Kalman filter, we adapt the latter's measurement-update formulation for cross-variable modeling while preserving the SSM backbone for robust temporal modeling. We further identify favorable inductive biases of GatedDeltaNet for time series, adapt it as our backbone, and improve generalization through additional techniques, including a previously unexplored stochastic perturbation of reversible instance normalization. On strongly dependent ODE and PEMS datasets, Chameleon achieves the best MSE and MAE across all settings, while its CI ablation and prior CD methods incur 61--178\% higher MSE on average. Across 28 standard benchmark settings, Chameleon also achieves better MSE and MAE than each baseline in at least 27 and 22 cases, respectively. Training-time, FLOP, and peak-memory analyses on Traffic and ETT further demonstrate competitive efficiency and favorable scalability across different variable counts.
\end{abstract}

\section{Introduction}
\label{sec:intro}

Multivariate time series forecasting (MTSF) refers to predicting the future values of multiple continuous and potentially correlated variables based on their histories. Accurate forecasts are valuable across applications including energy demand prediction, traffic flow estimation, weather forecasting, financial asset prediction, and product quality estimation in manufacturing~\citep{suganthi2012, hernandez2014, jiang2022, lana2018, bi2023, price2025, elliott2008, sezer2020, ren2023}. As data volume and complexity grow, deep neural networks (DNNs) have become widely adopted for MTSF. However, designing a generalizable architecture across diverse multivariate systems remains a long-standing challenge because cross-variable dependencies vary widely in strength and structure. Moreover, time series data are generally noisier, less regular, and less abundant than text or images, while both the temporal and variable dimensions can span hundreds of elements, making unconstrained joint modeling prone to overfitting, computationally expensive, and difficult to scale.

One simplified paradigm, channel-independent (CI) modeling, has therefore recently emerged and proven unexpectedly effective on benchmark datasets. PatchTST~\citep{nie2023} pioneered this line by forecasting each variable, also called a channel or series, independently through a shared Transformer~\citep{vaswani2017} that treats temporal patches as tokens while avoiding interactions across variables, achieving state-of-the-art (SOTA) performance at its release. Subsequent approaches, such as FITS~\citep{xu2024} and SFNN~\citep{sun2025}, reinforced this trend by matching or even surpassing prior, more complex channel-dependent (CD) architectures through one or a few feedforward layers. Together, these results highlight the importance of effective intra-series modeling and suggest that many current benchmarks exhibit weak cross-variable dependencies.

Nevertheless, information from other variables can be crucial in some settings, motivating renewed interest in CD models. Inspired by CI, a prominent family of CD methods embeds each variable independently over time before modeling dependencies across variables. In addition, to mitigate overfitting, architectural compromises are commonly adopted, often at the cost of modeling granularity or capacity for temporal dynamics. iTransformer~\citep{liu2024} popularized these strategies by representing each variable's entire input window as a single token using only a linear layer, while reserving the more expressive Transformer for cross-variable interactions. This formulation is retained by approaches such as SOFTS~\citep{han2024} and S-Mamba~\citep{wang2025}, which replace the Transformer with more scalable designs. Other variants, including Crossformer~\citep{zhang2023}, Leddam~\citep{yu2024}, and TimeMachine~\citep{ahamed2024}, follow the same factorization but employ either self-attention or Mamba along both temporal and variable dimensions. However, this added capacity incurs higher computational cost without consistently translating into better benchmark performance than CI SFNN~\citep{sun2025} or later, more constrained CD methods such as DUET~\citep{qiu2025} and TimeFilter~\citep{hu2025}. To improve generalization, DUET and TimeFilter instead restrict interactions to selected variable or patch pairs, yet still inherit similar compromises or introduce greater scalability concerns by constructing pairwise dependencies across patches from all variables. These developments therefore leave an unresolved trade-off among modeling granularity, temporal capacity, generalization, efficiency, and scalability.

These observations raise two natural questions:
\begin{enumerate}
    \item Can we incorporate data-dependent and fine-grained cross-variable interactions into a strong temporal modeling backbone while remaining efficient and scalable?
    \item If so, can we avoid overfitting and achieve SOTA performance in multivariate systems with both weak and strong dependencies across variables?
\end{enumerate}

These questions become even more compelling in light of recent advances in selective state space models (SSMs), pioneered by Mamba~\citep{gu2024}. Mamba retains the recency-emphasizing inductive bias of recurrent neural networks while overcoming their long-range memory limitations. Among its successors, GatedDeltaNet~\citep{yang2025} further introduces delta-based memory updates~\citep{schmidhuber1992}, enabling both rapid forgetting and targeted memory rewrites, which are desirable properties for noisy time series. Despite these advantages, GatedDeltaNet has not yet been explored for MTSF, while existing Mamba-based attempts largely inherit established architectures by replacing their original, often attention-based, backbones with Mamba~\citep{ahamed2024, wang2025}, leaving specialized cross-variable interaction mechanisms underexplored.

This work therefore presents \textbf{Chameleon}, an MTSF model built on our proposed CD SSM blocks to address these gaps. Our contributions are:
\begin{enumerate}
    \item We draw a parallel between selective SSMs and the Kalman filter~\citep{kalman1960} for MTSF, exploiting this connection to design an explicit cross-variable measurement update while maintaining linear complexity in the number of variables.

    \item We identify favorable inductive biases of GatedDeltaNet for time series, tailor it into an MTSF-specialized backbone, and further introduce a previously unexplored stochastic perturbation of reversible instance normalization~\citep{kim2022} during training to improve generalization.
    
    \item Chameleon achieves SOTA performance across standard long-horizon benchmarks and datasets with strong dependencies across variables.
\end{enumerate}

\section{Preliminaries}
\label{sec:preliminaries}

\paragraph{Problem Formulation.}
Given a historical window $\mathbf{X} \in \mathbb{R}^{N \times T}$ of $N$ variables observed at $T$ regularly spaced time steps, MTSF aims to forecast the next $F$ steps $\mathbf{Y} \in \mathbb{R}^{N \times F}$. Exact timestamp information is not explicitly used, and stationarity is not assumed, as trends and seasonality are prevalent in real-world time series.

\paragraph{Selective SSMs and GatedDeltaNet.}
Selective SSMs~\citep{gu2024, dao2024} have emerged as strong sequence learners through efficient long-context modeling and data-dependent state updates. GatedDeltaNet~\citep{yang2025} further combines the global decay of Mamba2~\citep{dao2024} for rapid forgetting with the delta rule of DeltaNet~\citep{yang2024} for targeted memory rewrites:
\begin{equation}
\label{eq:GatedDeltaNet}
    \mathbf{S}_t = \alpha_t \mathbf{S}_{t-1} (\mathbf{I} - \beta_t \mathbf{k}_t \mathbf{k}_t^\top) + \beta_t \mathbf{v}_t \mathbf{k}_t^\top \in \mathbb{R}^{D_v \times D_k},
\end{equation}
where $\alpha_t,\beta_t \in (0,1)$ control global decay and targeted updating, respectively, and $\mathbf{v}_t \in \mathbb{R}^{D_v}$ and $\mathbf{k}_t \in \mathbb{R}^{D_k}$ are input-dependent projections. This adaptive memory management provides a favorable inductive bias for filtering irrelevant information in noisy time series.

\paragraph{The Kalman Filter.}
\label{sec:KF}
The Kalman filter~\citep{kalman1960} estimates dynamic system states through a recursive two-step cycle. The ``time update'' produces the prior state estimate $\mathbf{S}_t^- \in \mathbb{R}^{D_s}$ and error covariance $\mathbf{P}_t^- \in \mathbb{R}^{D_s \times D_s}$:
\begin{equation}
\label{eq:KF_prediction}
    \mathbf{S}_t^- = \mathbf{A}_t \mathbf{S}_{t-1} + \mathbf{B}_t \mathbf{u}_t, \quad
    \mathbf{P}_t^- = \mathbf{A}_t \mathbf{P}_{t-1} \mathbf{A}_t^\top + \mathbf{Q}_t,
\end{equation}
where $\mathbf{A}_t \in \mathbb{R}^{D_s \times D_s}$ is the state transition matrix, $\mathbf{u}_t \in \mathbb{R}^{D_u}$ is the control input, $\mathbf{B}_t \in \mathbb{R}^{D_s \times D_u}$ is the control matrix, and $\mathbf{Q}_t \in \mathbb{R}^{D_s \times D_s}$ is the process noise covariance. The ``measurement update'' corrects this predicted prior state using new observations $\mathbf{z}_t \in \mathbb{R}^{D_z}$:
\begin{equation}
\label{eq:KF_correction}
    \mathbf{K}_t = \mathbf{P}_t^- \mathbf{H}_t^\top
    (\mathbf{H}_t \mathbf{P}_t^- \mathbf{H}_t^\top + \mathbf{R}_t)^{-1}, \quad
    \mathbf{S}_t = \mathbf{S}_t^- + \mathbf{K}_t
    (\mathbf{z}_t - \mathbf{H}_t \mathbf{S}_t^-), \quad
    \mathbf{P}_t = (\mathbf{I} - \mathbf{K}_t \mathbf{H}_t) \mathbf{P}_t^-,
\end{equation}
where $\mathbf{H}_t \in \mathbb{R}^{D_z \times D_s}$ maps state to measurement space, $\mathbf{R}_t \in \mathbb{R}^{D_z \times D_z}$ is the measurement noise covariance, and $\mathbf{K}_t \in \mathbb{R}^{D_s \times D_z}$ is the Kalman gain that weights the innovation by the relative confidence between prediction and measurement. The recurrent state update of the aforementioned SSMs naturally parallels the time update, with $\mathbf{v}_t$ serving as the control input. However, explicit measurement updates for SSMs remain underexplored for domain-specific modeling. This work leverages this connection to introduce a learnable and interpretable cross-variable modeling mechanism in MTSF.

\section{Method}
\label{sec:method}
The proposed framework processes the multivariate input through the following stages: optional reversible instance normalization (RevIN)~\citep{kim2022} and decomposition, patching and embedding, an encoder built from our proposed Chameleon blocks, and an MLP-based decoder. Before entering the encoder, each variable is independently transformed into patch-level representations, which are then processed by the encoder and mapped to forecasts by the decoder.

\subsection{Input Processing}
\label{sec:input}

The input $\mathbf{X} \in \mathbb{R}^{N \times T}$ optionally undergoes RevIN and decomposition to mitigate distribution shifts and preserve prominent trend structure, respectively. We select their use by comparing validation loss with and without each step. Appendix~\ref{app:preprocessing_guidance} provides supporting ablations and useful diagnostic metrics for preliminary guidance on new datasets.

\textbf{Reversible Instance Normalization.}
When enabled, the mean $\mu_i$ and standard deviation $\sigma_i$ for each variable $i$ are computed over a predefined normalization window and used to normalize the input as $\tilde{x}_{i,t} = (x_{i,t} - \mu_i)/\sigma_i$. When disabled, we instead set $\mu_i=0$ and $\sigma_i=1$ for all instances and variables. These statistics are stored for denormalizing the final predictions. To further improve generalization under distribution shifts, we introduce a simple yet previously unexplored data augmentation that perturbs $\mu_i$ and $\sigma_i$ during training:
\begin{equation}
    \tilde{\mu}_i = \mu_i + \epsilon_{\mu,i}, \quad
    \tilde{\sigma}_i = \sigma_i \cdot \exp(\epsilon_{\sigma,i}), \quad
    \epsilon_{\mu,i} \sim \mathcal{N}(0, \delta_\mu^2), \quad
    \epsilon_{\sigma,i} \sim \mathcal{N}(0, \delta_\sigma^2),
\end{equation}
where $\delta_\mu$ and $\delta_\sigma$ control the perturbation magnitude. The multiplicative parameterization of $\tilde{\sigma}_i$ ensures positivity without additional constraints. During training, $\mu_i$ and $\sigma_i$ in both normalization and denormalization are replaced by $\tilde{\mu}_i$ and $\tilde{\sigma}_i$, respectively.

\textbf{Decomposition, Patching, and Embedding.}
Following Autoformer~\citep{wu2021}, when decomposition is applied, each series is separated into a trend component $\tilde{x}_{i,t}^{\text{trend}}$ via a length-preserving moving average kernel and a seasonal component $\tilde{x}_{i,t}^{\text{seasonal}} = \tilde{x}_{i,t} - \tilde{x}_{i,t}^{\text{trend}}$. Each component is then divided into $L = \lceil T/T_p \rceil$ non-overlapping patches of length $T_p$ following PatchTST~\citep{nie2023}, producing $\mathbf{X}_i^{\text{trend}}, \mathbf{X}_i^{\text{seasonal}} \in \mathbb{R}^{L \times T_p}$. Replication padding is used to preserve length during decomposition and complete the final patch when needed. Rather than processing the two components through separate forecasting networks and combining their outputs at the final stage, we embed them through separate linear projections and immediately sum the resulting representations, keeping the additional cost minimal:
\begin{equation}
\label{eq:embed}
    \mathbf{x}^{\text{embed}}_{i,l} = \mathbf{W}^{(\text{s})} \mathbf{x}^{\text{seasonal}}_{i,l} + \mathbf{W}^{(\text{t})} \mathbf{x}^{\text{trend}}_{i,l} + \mathbf{e}_l \in \mathbb{R}^{D_v},
\end{equation}
where $\mathbf{W}^{(\text{s})}, \mathbf{W}^{(\text{t})} \in \mathbb{R}^{D_v \times T_p}$ are shared across patches and variables, and $\mathbf{e}_l$ is the fixed sinusoidal positional encoding~\citep{vaswani2017}. When decomposition is disabled, $\tilde{x}_{i,t}^{\text{seasonal}} = \tilde{x}_{i,t}$, and the embedding reduces accordingly to
$\mathbf{x}^{\text{embed}}_{i,l} = \mathbf{W}^{(\text{s})} \mathbf{x}^{\text{seasonal}}_{i,l} + \mathbf{e}_l$.

\subsection{Chameleon Blocks and Encoder}
\label{sec:encoder}
\begin{figure}[t]
    \centering
    \includegraphics[width=1\textwidth]{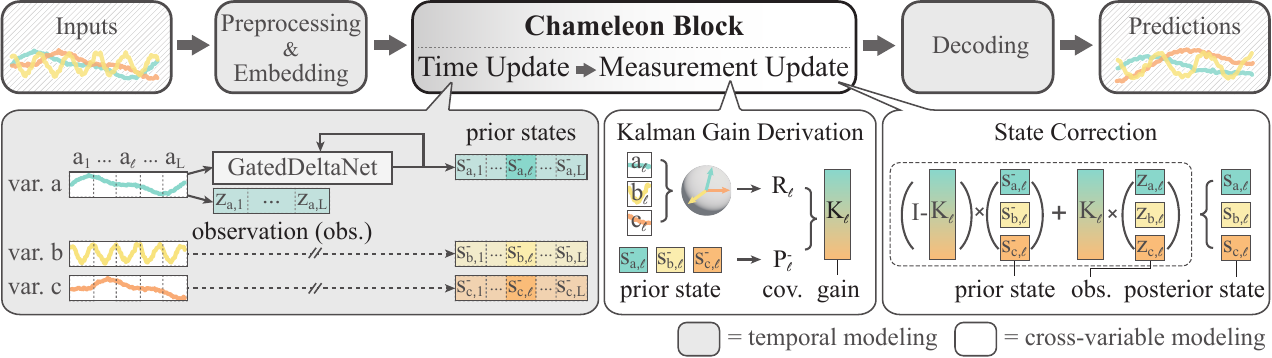}
    \caption{Overview of Chameleon, which adopts GatedDeltaNet for temporal modeling and incorporates a Kalman-filter-inspired measurement update for cross-variable modeling through data-dependent observation signals (obs.) and covariances (cov.).}
    \label{fig:model}
\end{figure}

The encoder is composed of repeated units, each consisting of root mean square layer normalization (RMSNorm)~\citep{zhang2019} followed by our Chameleon block and an optional feedforward network block. In the classical Kalman filter, both the time and measurement updates operate on the joint state of all variables and alternate recursively, but such joint modeling has been found suboptimal for Mamba in MTSF~\citep{weng2025}. Chameleon therefore follows recent factorized architectures to separate temporal and variable modeling, leveraging GatedDeltaNet for the former and the Kalman filter's measurement-update formulation for the latter, as illustrated in Figure~\ref{fig:model}.

\paragraph{Time Update.}
The GatedDeltaNet block is leveraged as the temporal modeling backbone due to its advantageous inductive biases. For variable $i$, given the RMSNorm-processed $l$-th patch token $\bar{\mathbf{x}}_{i,l} \in \mathbb{R}^{D_v}$, the SSM computes a prior state via Equation~\ref{eq:GatedDeltaNet} in a CI manner:
\begin{equation}
\label{eq:prediction}
\mathbf{S}_{i,l}^-
=
\alpha_{i,l}\mathbf{S}_{i,l-1}^-
\left(
\mathbf{I}
-\beta_{i,l}\mathbf{k}_{i,l}\mathbf{k}_{i,l}^{\top}
\right)
+
\beta_{i,l}\mathbf{v}_{i,l}\mathbf{k}_{i,l}^{\top}
\in\mathbb{R}^{D_v\times D_k},
\end{equation}
where $\alpha_{i,l}$, $\beta_{i,l}$, $\mathbf{k}_{i,l}$, and $\mathbf{v}_{i,l}$ are all projected from $\bar{\mathbf{x}}_{i,l}$. We omit the original convolutional layers for deriving $\mathbf{k}_{i,l}$ and $\mathbf{v}_{i,l}$ because patch embeddings already encode local temporal details. The ablation in CMamba~\citep{zeng2024} empirically supports this choice for an SSM backbone with patch tokens, and this removal also reduces computation. Structurally, $\alpha_{i,l}(\mathbf{I} - \beta_{i,l} \mathbf{k}_{i,l} \mathbf{k}_{i,l}^\top) \in \mathbb{R}^{D_k \times D_k}$, $\beta_{i,l} \mathbf{k}_{i,l}^\top$, and $\mathbf{v}_{i,l}$ are respectively analogous to the state transition matrix, control matrix, and control input in the Kalman filter's time update of Equation~\ref{eq:KF_prediction}, but the update here is performed independently within each variable's state without computation across variables.

\paragraph{Measurement Update.}
\label{sec:measurement}
To capture informative dependencies across variables, we adapt the Kalman filter's measurement update into a learnable state-correction mechanism. Crucially, existing work suggests that unconstrained cross-variable modeling, even with greater capacity, can often harm generalization, particularly when dependencies are weak~\citep{sun2025,abdelmalak2025}. We therefore base the following design choices on prior empirical findings, rather than assuming that adopting the full Kalman-filter formulation or allowing greater modeling flexibility would necessarily improve forecasting performance. Specifically, unlike the original Kalman filter, Chameleon corrects the states of all patch steps once after completing the temporal recurrence without feeding them back. This preserves CI temporal modeling in Equation~\ref{eq:prediction} within each block, making it easier to rely primarily on the temporal backbone when cross-variable dependencies are weak. Moreover, joint recurrence would incur higher computational overhead that can be unnecessary in such cases and disrupt GatedDeltaNet's original chunkwise parallel computation. Following common practice in PatchTST~\citep{nie2023} and other deep sequential models for MTSF, Chameleon passes the outputs from all corrected stepwise states to subsequent blocks, where they can still interact temporally and contribute to final predictions. 

The actual benefit we draw from the Kalman filter is its interpretable measurement-update formulation, which gives prior uncertainty, observation uncertainty, and cross-variable correlations distinct roles with corresponding constraints. The challenge is thus to construct an observation signal $\mathbf{Z}_l$ and a Kalman gain $\mathbf{K}_l\in\mathbb{R}^{N\times N}$ analogous to Equation~\ref{eq:KF_correction}, which involves specifying $\mathbf{H}_l,\mathbf{P}_l^-,\mathbf{R}_l\in\mathbb{R}^{N\times N}$. The same patch tokens $\bar{\mathbf{x}}_{i,l}$ are reused to construct $\mathbf{Z}_l$. Specifically, input projections $\mathbf{o}_{i,l}\in\mathbb{R}^{D_v}$ and $\mathbf{h}_{i,l}\in\mathbb{R}^{D_k}$ follow the same design as $\mathbf{v}_{i,l}$ and $\mathbf{k}_{i,l}$, and their outer products $\mathbf{o}_{i,l}\mathbf{h}_{i,l}^{\top}\in\mathbb{R}^{D_v\times D_k}$ are flattened and stacked across variables to form $\mathbf{Z}_l\in\mathbb{R}^{N\times(D_vD_k)}$. Unlike physical measurements supplied in a distinct space, these observations are derived directly in the hidden-state coordinates, naturally leading to $\mathbf{H}_l=\mathbf{I}_{N\times N}$ while their variable-wise content and scale remain learnable. In addition, a dense learnable $\mathbf{H}_l$ would introduce another unconstrained linear transformation across variables, and SFNN's ablation shows that such a design can degrade performance on weakly dependent benchmarks~\citep{sun2025}, motivating us to concentrate cross-variable interactions in the more structured Kalman gain.

Given that the temporal recurrence operates independently for each variable without feedback from corrected states, a diagonal prior error covariance $\mathbf{P}_l^-=\operatorname{diag}(\mathbf{p}_l)$, where $\mathbf{p}_l\in\mathbb{R}^N_+$, is a natural modeling assumption. Learning off-diagonal covariances independently in both $\mathbf{P}_l^-$ and $\mathbf{R}_l$ can also introduce redundancy because shared correlations can cancel in the Kalman gain, as suggested by Equation~\ref{eq:KF_correction}. $\mathbf{P}_l^-$ thus reflects variable-wise prior uncertainty, while $\mathbf{R}_l$ captures the patch-wise cross-variable covariance structure. To separate observation uncertainty magnitudes from correlation strengths, we parameterize $\mathbf{R}_l$ through the covariance-correlation decomposition $\mathbf{R}_l=\mathbf{\Lambda}_l^{1/2}\mathbf{C}_l\mathbf{\Lambda}_l^{1/2}$, where $\mathbf{\Lambda}_l=\operatorname{diag}(\mathbf{d}_l)$ with $\mathbf{d}_l\in\mathbb{R}^N_+$, and $\mathbf{C}_l\in\mathbb{R}^{N\times N}$ is a correlation matrix. Both $\mathbf{p}_l$ and $\mathbf{d}_l$ are designed to be data-dependent, with their parameterization elaborated later.

To build $\mathbf{C}_l$, an additional input projection $\boldsymbol{\eta}_{i,l} \in \mathbb{R}^{D_\eta}$ is normalized onto the unit sphere for each variable $i$. Stacking these projections yields $\boldsymbol{\eta}_l \in \mathbb{R}^{N \times D_\eta}$, whose Gram matrix $\boldsymbol{\eta}_l\boldsymbol{\eta}_l^\top$ contains pairwise cosine similarities that reflect dependencies among variables in the projected space. We further introduce a global learnable scale $\gamma \in (0,1)$ and define:
\begin{equation}
\label{eq:correlation}
\mathbf{C}_l = (1-\gamma)\mathbf{I} + \gamma\boldsymbol{\eta}_l\boldsymbol{\eta}_l^\top \in \mathbb{R}^{N \times N},
\end{equation}
which is symmetric positive semidefinite with a unit diagonal, forming a valid correlation matrix. The parameter $\gamma$ controls the strength of cross-variable interactions and can attenuate them when dependencies are weak. A small initial $\gamma$ also limits unreliable interactions early in training, before the learned dependencies become informative. Defining $\mathbf{U}_l = \sqrt{\gamma}\mathbf{\Lambda}_l^{1/2}\boldsymbol{\eta}_l$ gives $\mathbf{R}_l = (1-\gamma)\mathbf{\Lambda}_l + \mathbf{U}_l\mathbf{U}_l^\top$, and thus $\mathbf{P}_l^- + \mathbf{R}_l = \operatorname{diag}(\mathbf{p}_l + (1-\gamma)\mathbf{d}_l) + \mathbf{U}_l\mathbf{U}_l^\top$. The Kalman gain in Equation~\ref{eq:KF_correction} hence admits the Woodbury form:
\begin{equation}
\label{eq:woodbury_K}
    \mathbf{K}_l = \mathbf{P}_l^- (\mathbf{P}_l^- + \mathbf{R}_l)^{-1} = \operatorname{diag}(\mathbf{p}_l)\left[\mathbf{\Omega}_l - \mathbf{\Omega}_l\mathbf{U}_l\left(\mathbf{I} + \mathbf{U}_l^\top\mathbf{\Omega}_l\mathbf{U}_l\right)^{-1}\mathbf{U}_l^\top\mathbf{\Omega}_l\right] \in \mathbb{R}^{N \times N},
\end{equation}
where $\mathbf{\Omega}_l = \operatorname{diag}(\mathbf{p}_l + (1-\gamma)\mathbf{d}_l)^{-1}$. Although $\mathbf{C}_l$ and $\mathbf{K}_l$ are expressed as $N \times N$ matrices, neither is explicitly materialized for large $N$. Instead, the computation operates directly on the diagonal terms and the low-rank factor $\mathbf{U}_l$ when applying the gain to the subsequent state correction. This avoids $\mathcal{O}(N^2)$ storage and reduces the inversion complexity from $\mathcal{O}(N^3)$ to $\mathcal{O}(N D_\eta^2 + D_\eta^3)$, which is linear in $N$ for fixed $D_\eta$ and substantially improves scalability.

Since $\mathbf{p}_l$ and $\mathbf{d}_l$ reflect the uncertainty of the prior state $\mathbf{S}_{i,l}^-$ and observation signal, respectively, we make both dependent on the input tokens, with $\mathbf{p}_l$ additionally conditioned on the prior state. We therefore introduce another input projection to produce the log observation variance $\log d_{i,l}$ for each variable $i$. Because multiplying $\mathbf{p}_l$ and $\mathbf{d}_l$ by the same positive scalar leaves the Kalman gain unchanged, their common scale is unidentifiable. We remove this ambiguity by centering $\log d_{i,l}$ across variables, fixing the geometric mean of $\mathbf{d}_l$ to one. With this common scale fixed, rather than estimating $p_{i,l}$ directly, we model its ratio to $d_{i,l}$ through $\tau_{i,l}=\log(p_{i,l}/d_{i,l})$. Following the state readout design of GatedDeltaNet, an additional normalized query vector extracts a compact representation from $\mathbf{S}_{i,l}^-$, which is combined with the input token to produce $\tau_{i,l}$, and the prior variance is recovered as $p_{i,l}=d_{i,l}\exp(\tau_{i,l})$. Both the centered $\log d_{i,l}$ and $\tau_{i,l}$ are bounded by predefined values for numerical stability, while exponentiation guarantees $d_{i,l},p_{i,l}>0$.

The prior state $\mathbf{S}_l^- \in \mathbb{R}^{N \times D_v \times D_k}$ is then flattened into $\tilde{\mathbf{S}}_l^- \in \mathbb{R}^{N \times (D_vD_k)}$ and corrected via the data-dependent Kalman gain $\mathbf{K}_l$ and observation signal $\mathbf{Z}_l$:
\begin{equation}
\label{eq:correction}
    \tilde{\mathbf{S}}_l = (\mathbf{I} - \mathbf{K}_l)\tilde{\mathbf{S}}_l^- + \mathbf{K}_l \mathbf{Z}_l \in \mathbb{R}^{N \times (D_vD_k)}.
\end{equation}
After restoring the $D_v \times D_k$ state shape, the posterior $\mathbf{S}_{i,l}$ and prior $\mathbf{S}_{i,l}^-$ are fused and read out independently for each variable through another global learnable scalar gate $\omega \in (0,1)$:
\begin{equation}
\label{eq:out}
    \mathbf{y}^{\text{enc}}_{i,l} = [\omega \mathbf{S}_{i,l} + (1-\omega)\mathbf{S}_{i,l}^-]\,\mathbf{q}_{i,l} \in \mathbb{R}^{D_v},
\end{equation}
where $\mathbf{q}_{i,l} \in \mathbb{R}^{D_k}$ is the GatedDeltaNet output query vector projected from $\bar{\mathbf{x}}_{i,l}$. The prior branch provides a gated skip connection around the measurement update step to improve stability and flexibility. While $\gamma$ controls cross-variable interactions within the Kalman gain, $\omega$ gates the entire correction and can similarly suppress it when dependencies are weak. Initializing $\omega$ to a small value also helps keep the Chameleon block close to the CI backbone early in training.

\subsection{Decoder}
The decoder maps the encoded representations to the forecast horizon $F$ in a CI manner. For each variable $i$, the patch representations $\mathbf{Y}^{\text{enc}}_i \in \mathbb{R}^{L \times D_v}$ are flattened into a vector of dimension $LD_v$, followed by a SiLU activation and a linear projection to produce $\hat{\mathbf{Y}}' \in \mathbb{R}^{N \times F}$. Optional decoder blocks consisting of RMSNorm, SiLU, and a width-preserving linear layer can be appended but are typically not used. When RevIN is enabled, the final predictions $\hat{\mathbf{Y}} \in \mathbb{R}^{N \times F}$ are obtained by reversing the normalization with the stored statistics. Otherwise, $\hat{\mathbf{Y}}=\hat{\mathbf{Y}}'$.

\section{Experiments}
\label{sec:experiments}
Chameleon is trained with smooth L1 (Huber) loss using Adam~\citep{kingma2014} under a StepLR schedule, with parameters additionally smoothed via exponential moving average. All experiments are conducted on a single NVIDIA L40S (46GB) or H100 (80GB) GPU. Full hyperparameters and code are available at \url{https://github.com/anonymous/Chameleon-TSF}.

\subsection{Datasets}
\label{sec:datasets}
Based on Granger causality analysis and empirical comparisons of prior CI and CD methods, \citet{abdelmalak2025} argue that current long-horizon MTSF benchmarks exhibit weak dependencies, and accordingly suggest several simulation datasets with strong coupling governed by physical ordinary differential equations (ODEs)~\citep{gilpin2021}. To examine the proposed measurement update, we therefore select Double Pendulum and Lorenz Coupled, two of the strongest datasets identified in these analyses. As a complementary real-world setting with a different dependency structure and higher dimensionality, PEMS~\citep{song2020} contains traffic flow measurements from hundreds of spatially distributed sensors on California highways. Traffic propagation can induce approximately linear relationships between sensors at temporal offsets, making maximum pairwise lagged correlation particularly informative. Although Granger causality does not indicate stronger dependencies than in the benchmarks, our analysis shows that all PEMS datasets exhibit an average maximum pairwise lagged correlation above 0.8, compared with around 0.45 for the Traffic benchmark. PEMS03 and PEMS08 are selected because they show the strongest dependencies under both lagged-correlation and Granger-causality analyses. Although PEMS provides additional spatial information, neither Chameleon nor the baselines in Table~\ref{tab:strong_results} use it. Appendix~\ref{app:results} further compares Chameleon with MGCN~\citep{lin2025}, which incorporates this information.

Still, we test Chameleon on seven long-horizon benchmarks to evaluate whether it can achieve SOTA performance in these common settings, where more competitive baselines are available. These experiments further evaluate the effectiveness of the GatedDeltaNet backbone, the proposed RevIN perturbation, and other temporal modeling designs, and examine whether the measurement update can exploit relatively stronger dependencies without degrading performance when they are weak. The evaluated benchmarks include ETTm1, ETTm2, ETTh1, ETTh2~\citep{zhou2021}, Electricity, Solar Energy, and Traffic~\citep{lai2018}. Following SFNN~\citep{sun2025}, we exclude Weather, Illness, and Exchange Rate due to known issues, such as low predictability or insufficient samples.

We follow the standard practice of splitting all datasets chronologically into training, validation, and test partitions. Appendix~\ref{app:datasets} provides the split ratios and brief descriptions of all datasets.

\subsection{Setup}
\label{sec:protocol}
\paragraph{Protocol.}
We use the common forecast horizons $F \in \{96, 192, 336, 720\}$ for all benchmarks, while we evaluate the PEMS and ODE datasets at two horizons each. We use $F \in \{48,96\}$ for PEMS due to better baseline availability, and exclude longer horizons for the chaotic ODE systems, where nearly identical histories can diverge sharply farther into the future, as shown in Appendix~\ref{app:datasets}. Because the conventional look-back size $T=96$ can omit important periodic context and disadvantage methods with greater long-range temporal modeling capacity, as analyzed in Appendix~\ref{app:periodicity}, we follow recent works~\citep{abdelmalak2025,sun2025,liu2024,qiu2025,hu2025,xu2024} in evaluating multiple look-back sizes and adopt SFNN's grids~\citep{sun2025} for the benchmarks. All methods use the same candidate grid within each setting, with dataset-specific $T$ and $F$ values listed in Appendix~\ref{app:datasets}. For each dataset-horizon-model combination, we evaluate each $T$ over five independent runs and select the $T$ with the lowest validation loss under the same objective as in training. We report the corresponding test mean squared error (MSE) and mean absolute error (MAE) averaged across runs, and also assess statistical significance using two-sided Welch's $t$-tests.

\paragraph{Baselines.}
We compare against six competitive methods spanning the major modeling paradigms. S-Mamba~\citep{wang2025} with adaptive channel normalization (ACN)~\citep{lee2025} is, to our knowledge, currently the strongest CD SSM-based method, while TimeFilter~\citep{hu2025} and DUET~\citep{qiu2025} are SOTA CD approaches leading under both conventional ($T=96$) and longer look-back sizes. These two methods also share Chameleon's motivation to handle both strongly and weakly dependent systems. Crossformer~\citep{zhang2023}, the strongest model evaluated by \citet{abdelmalak2025} on Double Pendulum and Lorenz Coupled at shorter horizons, is included for the ODE and PEMS datasets. For the benchmarks, the primarily CI SFNN~\citep{sun2025} and CI FITS~\citep{xu2024} provide additional SOTA references.

\subsection{Results}
Across all ODE and PEMS datasets and horizons, Chameleon achieves the best MSE and MAE, with statistical significance over all baselines in seven of the eight settings, as shown in Table~\ref{tab:strong_results}. Removing the measurement update raises MSE and MAE by 61\% and 34\% on average, reaching 175\% and 96\%, respectively, on Lorenz Coupled ($F=48$), demonstrating its substantial benefit when cross-variable dependencies are strong. Prior CD methods incur even larger average increases of 62--178\% in MSE and 47--110\% in MAE relative to Chameleon. Appendix~\ref{app:results} includes an additional comparison against MGCN~\citep{lin2025} on PEMS, where Chameleon also achieves better forecasting accuracy without using extra spatial information.

Table~\ref{tab:main_results} complements these findings with horizon-averaged results on the standard benchmarks. Chameleon ranks first in MSE on all seven benchmarks and in MAE on six, while the baselines incur 3.2--11.3\% higher MSE and 1.9--9.8\% higher MAE. Notably, CD baselines frequently trail CI methods such as FITS~\citep{xu2024} and the primarily CI SFNN~\citep{sun2025} in Table~\ref{tab:main_results}, as well as Chameleon's CI variant in Table~\ref{tab:strong_results}. This pattern highlights the importance of preserving strong temporal modeling capacity, one of Chameleon's central goals. Complete per-horizon means, standard deviations, and statistical significance tests across runs are reported in Appendix~\ref{app:results}.



\newlength{\strongmetricwidth}
\setlength{\strongmetricwidth}{1cm}

\newlength{\strongmetricgap}
\setlength{\strongmetricgap}{3pt}

\newlength{\strongsegmentgap}
\setlength{\strongsegmentgap}{10pt}

\newlength{\stronghalfmodelgap}
\setlength{\stronghalfmodelgap}{5pt}

\newlength{\strongmodelblockwidth}
\setlength{\strongmodelblockwidth}{%
  \dimexpr 2\strongmetricwidth+\strongmetricgap\relax
}


\newsavebox{\strongmodelrefbox}
\newsavebox{\strongmodeltmpbox}
\newsavebox{\strongmodelscaledbox}

\sbox{\strongmodelrefbox}{\strut \textbf{Chameleon}}

\newcommand{\strongconsiderModel}[1]{%
  \sbox{\strongmodeltmpbox}{\strut #1}%
  \ifdim\wd\strongmodeltmpbox>\wd\strongmodelrefbox
    \sbox{\strongmodelrefbox}{\strut #1}%
  \fi
}

\strongconsiderModel{Chameleon (CI)}
\strongconsiderModel{S-Mamba+ACN}
\strongconsiderModel{TimeFilter}
\strongconsiderModel{DUET}
\strongconsiderModel{Crossformer}

\ifdim\wd\strongmodelrefbox>\strongmodelblockwidth
  \sbox{\strongmodelscaledbox}{%
    \resizebox{\strongmodelblockwidth}{!}{\usebox{\strongmodelrefbox}}%
  }%
\else
  \sbox{\strongmodelscaledbox}{\usebox{\strongmodelrefbox}}%
\fi

\newlength{\strongmodelheadheight}
\setlength{\strongmodelheadheight}{%
  \dimexpr
    \ht\strongmodelscaledbox+\dp\strongmodelscaledbox
  \relax
}

\newcommand{\stronguniformmodel}[1]{%
  \resizebox*{!}{\strongmodelheadheight}{\strut #1}%
}


\newcommand{\strongmodelhead}[1]{%
  \multicolumn{2}{@{}c@{}}{%
    \hspace*{-\stronghalfmodelgap}%
    \makebox[\strongmodelblockwidth][c]{%
      \stronguniformmodel{#1}%
    }%
    \hspace*{\stronghalfmodelgap}%
  }%
}

\newlength{\stronglastmodelshift}
\setlength{\stronglastmodelshift}{%
  \dimexpr\stronghalfmodelgap/2\relax
}

\newcommand{\stronglastmodelhead}[1]{%
  \multicolumn{2}{@{}c@{}}{%
    \hspace*{-\stronglastmodelshift}%
    \makebox[\strongmodelblockwidth][c]{%
      \stronguniformmodel{#1}%
    }%
    \hspace*{\stronglastmodelshift}%
  }%
}


\definecolor{stronghighlightcolor}{HTML}{9ECED0}
\colorlet{strongbestlight}{stronghighlightcolor!90}
\colorlet{strongsecondlight}{stronghighlightcolor!35}

\newcommand{\strongbestst}[1]{%
  {\setlength{\fboxsep}{0pt}%
  \colorbox{strongbestlight}{%
    \makebox[\strongmetricwidth][c]{%
      \strut \underline{#1}%
    }%
  }}%
}

\newcommand{\strongbest}[1]{%
  {\setlength{\fboxsep}{0pt}%
  \colorbox{strongbestlight}{%
    \makebox[\strongmetricwidth][c]{\strut #1}%
  }}%
}

\newcommand{\strongsecond}[1]{%
  {\setlength{\fboxsep}{0pt}%
  \colorbox{strongsecondlight}{%
    \makebox[\strongmetricwidth][c]{\strut #1}%
  }}%
}

\newcommand{\strongcaptionbest}[1]{%
  {\setlength{\fboxsep}{1pt}%
  \colorbox{strongbestlight}{#1}}%
}

\newcommand{\strongcaptionsecond}[1]{%
  {\setlength{\fboxsep}{1pt}%
  \colorbox{strongsecondlight}{#1}}%
}


\newcommand{\strongincpct}[1]{%
  \makebox[\strongmetricwidth][r]{#1\%}%
}


\begin{table*}[t]
\centering
\caption{Results on strongly dependent datasets.
Chameleon (CI) removes the proposed cross-variable measurement update.
\strongcaptionbest{Best} and \strongcaptionsecond{second-best} results are highlighted.
\strongcaptionbest{\underline{Underlining}} denotes statistical significance over all other baselines ($p < 0.05$).
The Avg./Max. increase rows report each baseline's average/maximum relative error increase over Chameleon across all settings, with larger values indicating worse performance.}
\label{tab:strong_results}

\resizebox{1.0\textwidth}{!}{%
\begin{tabular}{
l@{\hspace{5pt}}r|
@{\hspace{\stronghalfmodelgap}}
>{\centering\arraybackslash}p{\strongmetricwidth}
@{\hspace{\strongmetricgap}}
>{\centering\arraybackslash}p{\strongmetricwidth}
@{\hspace{\stronghalfmodelgap}}
@{\hspace{\stronghalfmodelgap}}
>{\centering\arraybackslash}p{\strongmetricwidth}
@{\hspace{\strongmetricgap}}
>{\centering\arraybackslash}p{\strongmetricwidth}
@{\hspace{\stronghalfmodelgap}}
@{\hspace{\stronghalfmodelgap}}
>{\centering\arraybackslash}p{\strongmetricwidth}
@{\hspace{\strongmetricgap}}
>{\centering\arraybackslash}p{\strongmetricwidth}
@{\hspace{\stronghalfmodelgap}}
@{\hspace{\stronghalfmodelgap}}
>{\centering\arraybackslash}p{\strongmetricwidth}
@{\hspace{\strongmetricgap}}
>{\centering\arraybackslash}p{\strongmetricwidth}
@{\hspace{\stronghalfmodelgap}}
@{\hspace{\stronghalfmodelgap}}
>{\centering\arraybackslash}p{\strongmetricwidth}
@{\hspace{\strongmetricgap}}
>{\centering\arraybackslash}p{\strongmetricwidth}
@{\hspace{\stronghalfmodelgap}}
@{\hspace{\stronghalfmodelgap}}
>{\centering\arraybackslash}p{\strongmetricwidth}
@{\hspace{\strongmetricgap}}
>{\centering\arraybackslash}p{\strongmetricwidth}
}

\toprule

\multicolumn{2}{c}{Model}
& \strongmodelhead{\textbf{Chameleon}}
& \strongmodelhead{Chameleon (CI)}
& \strongmodelhead{S-Mamba+ACN}
& \strongmodelhead{TimeFilter}
& \strongmodelhead{DUET}
& \stronglastmodelhead{Crossformer} \\[-0.5ex]

\cmidrule(lr){1-2}
\cmidrule(l{0pt}r{\strongsegmentgap}){3-4}
\cmidrule(l{0pt}r{\strongsegmentgap}){5-6}
\cmidrule(l{0pt}r{\strongsegmentgap}){7-8}
\cmidrule(l{0pt}r{\strongsegmentgap}){9-10}
\cmidrule(l{0pt}r{\strongsegmentgap}){11-12}
\cmidrule(l{0pt}r){13-14}

\noalign{\vskip 0.5ex}

Dataset & (F)
& MSE & MAE
& MSE & MAE
& MSE & MAE
& MSE & MAE
& MSE & MAE
& MSE & MAE \\

\midrule


\multirow{2}{*}{\shortstack[l]{Double\\Pendulum}}
& 96
& \strongbestst{0.0678} & \strongbestst{0.1176}
& 0.1258 & 0.1880
& 0.0927 & 0.1642
& \strongsecond{0.0870} & \strongsecond{0.1616}
& 0.2510 & 0.3271
& 0.1208 & 0.1975 \\

& 192
& \strongbestst{0.3053} & \strongbestst{0.3216}
& 0.3845 & 0.3836
& \strongsecond{0.3325} & \strongsecond{0.3701}
& 0.3522 & 0.3787
& 0.5044 & 0.5052
& 0.4774 & 0.4717 \\

\midrule


\multirow{2}{*}{\shortstack[l]{Lorenz\\Coupled}}
& 48
& \strongbestst{0.0179} & \strongbestst{0.0389}
& 0.0493 & \strongsecond{0.0761}
& 0.0708 & 0.1247
& \strongsecond{0.0424} & 0.0934
& 0.1570 & 0.2137
& 0.0581 & 0.1152 \\

& 96
& \strongbestst{0.1569} & \strongbestst{0.1842}
& 0.3368 & \strongsecond{0.3213}
& 0.3666 & 0.3837
& \strongsecond{0.2935} & 0.3230
& 0.5026 & 0.4758
& 0.3019 & 0.3299 \\

\midrule


\multirow{2}{*}{PEMS03}
& 48
& \strongbestst{0.0905} & \strongbestst{0.1920}
& 0.1045 & 0.1991
& \strongsecond{0.0927} & \strongsecond{0.1968}
& 0.1048 & 0.2099
& 0.1028 & 0.2050
& 0.0968 & 0.1989 \\

& 96
& \strongbest{0.1156} & \strongbest{0.2157}
& 0.1359 & 0.2229
& 0.2309 & 0.3330
& 0.1407 & 0.2389
& 0.1447 & 0.2407
& \strongsecond{0.1211} & \strongsecond{0.2208} \\

\midrule


\multirow{2}{*}{PEMS08}
& 48
& \strongbestst{0.0975} & \strongbestst{0.1783}
& \strongsecond{0.1187} & \strongsecond{0.1924}
& 0.1266 & 0.1985
& 0.1617 & 0.2047
& 0.1244 & 0.1929
& 0.1525 & 0.2111 \\

& 96
& \strongbestst{0.1307} & \strongbestst{0.1910}
& 0.1703 & 0.2115
& 0.1869 & \strongsecond{0.2064}
& 0.2960 & 0.3193
& \strongsecond{0.1663} & 0.2087
& 0.2201 & 0.2335 \\

\midrule


\multicolumn{2}{c}{Avg. Increase}
& -- & --
& \strongincpct{61} & \strongincpct{34}
& \strongincpct{81} & \strongincpct{57}
& \strongincpct{62} & \strongincpct{47}
& \strongincpct{178} & \strongincpct{110}
& \strongincpct{74} & \strongincpct{55} \\

\multicolumn{2}{c}{Max. Increase}
& -- & --
& \strongincpct{175} & \strongincpct{96}
& \strongincpct{296} & \strongincpct{221}
& \strongincpct{137} & \strongincpct{140}
& \strongincpct{777} & \strongincpct{449}
& \strongincpct{225} & \strongincpct{196} \\

\bottomrule

\end{tabular}%
}
\end{table*}


\newlength{\longmetricwidth}
\setlength{\longmetricwidth}{1cm}

\newlength{\longmetricgap}
\setlength{\longmetricgap}{3pt}

\newlength{\longsegmentgap}
\setlength{\longsegmentgap}{10pt}

\newlength{\longhalfmodelgap}
\setlength{\longhalfmodelgap}{5pt}

\newlength{\longmodelblockwidth}
\setlength{\longmodelblockwidth}{%
  \dimexpr 2\longmetricwidth+\longmetricgap\relax
}


\newsavebox{\longmodelrefbox}
\newsavebox{\longmodeltmpbox}
\newsavebox{\longmodelscaledbox}

\sbox{\longmodelrefbox}{\strut \textbf{Chameleon}}

\newcommand{\longconsiderModel}[1]{%
  \sbox{\longmodeltmpbox}{\strut #1}%
  \ifdim\wd\longmodeltmpbox>\wd\longmodelrefbox
    \sbox{\longmodelrefbox}{\strut #1}%
  \fi
}

\longconsiderModel{S-Mamba+ACN}
\longconsiderModel{TimeFilter}
\longconsiderModel{DUET}
\longconsiderModel{SFNN}
\longconsiderModel{FITS}

\ifdim\wd\longmodelrefbox>\longmodelblockwidth
  \sbox{\longmodelscaledbox}{%
    \resizebox{\longmodelblockwidth}{!}{\usebox{\longmodelrefbox}}%
  }%
\else
  \sbox{\longmodelscaledbox}{\usebox{\longmodelrefbox}}%
\fi

\newlength{\longmodelheadheight}
\setlength{\longmodelheadheight}{%
  \dimexpr
    \ht\longmodelscaledbox+\dp\longmodelscaledbox
  \relax
}

\newcommand{\longuniformmodel}[1]{%
  \resizebox*{!}{\longmodelheadheight}{\strut #1}%
}


\newcommand{\longmodelhead}[1]{%
  \multicolumn{2}{@{}c@{}}{%
    \hspace*{-\longhalfmodelgap}%
    \makebox[\longmodelblockwidth][c]{%
      \longuniformmodel{#1}%
    }%
    \hspace*{\longhalfmodelgap}%
  }%
}

\newlength{\longlastmodelshift}
\setlength{\longlastmodelshift}{%
  \dimexpr\longhalfmodelgap/2\relax
}

\newcommand{\longlastmodelhead}[1]{%
  \multicolumn{2}{@{}c@{}}{%
    \hspace*{-\longlastmodelshift}%
    \makebox[\longmodelblockwidth][c]{%
      \longuniformmodel{#1}%
    }%
    \hspace*{\longlastmodelshift}%
  }%
}


\definecolor{longhighlightcolor}{HTML}{9ECED0}
\colorlet{longbestlight}{longhighlightcolor!90}
\colorlet{longsecondlight}{longhighlightcolor!35}

\newcommand{\longbest}[1]{%
  {\setlength{\fboxsep}{0pt}%
  \colorbox{longbestlight}{%
    \makebox[\longmetricwidth][c]{\strut #1}%
  }}%
}

\newcommand{\longsecond}[1]{%
  {\setlength{\fboxsep}{0pt}%
  \colorbox{longsecondlight}{%
    \makebox[\longmetricwidth][c]{\strut #1}%
  }}%
}

\newcommand{\longcaptionbest}[1]{%
  {\setlength{\fboxsep}{1pt}%
  \colorbox{longbestlight}{#1}}%
}

\newcommand{\longcaptionsecond}[1]{%
  {\setlength{\fboxsep}{1pt}%
  \colorbox{longsecondlight}{#1}}%
}

%

\newcommand{\longincpct}[1]{%
  \makebox[\longmetricwidth][r]{#1\%}%
}


\begin{table*}[t]
\centering
\caption{Results on long-horizon benchmarks, averaged across four forecasting horizons.
Highlighting and Avg./Max. increase conventions follow Table~\ref{tab:strong_results}.}
\label{tab:main_results}

\resizebox{1.0\textwidth}{!}{%
\begin{tabular}{
l|
@{\hspace{\longhalfmodelgap}}
>{\centering\arraybackslash}p{\longmetricwidth}
@{\hspace{\longmetricgap}}
>{\centering\arraybackslash}p{\longmetricwidth}
@{\hspace{\longhalfmodelgap}}
@{\hspace{\longhalfmodelgap}}
>{\centering\arraybackslash}p{\longmetricwidth}
@{\hspace{\longmetricgap}}
>{\centering\arraybackslash}p{\longmetricwidth}
@{\hspace{\longhalfmodelgap}}
@{\hspace{\longhalfmodelgap}}
>{\centering\arraybackslash}p{\longmetricwidth}
@{\hspace{\longmetricgap}}
>{\centering\arraybackslash}p{\longmetricwidth}
@{\hspace{\longhalfmodelgap}}
@{\hspace{\longhalfmodelgap}}
>{\centering\arraybackslash}p{\longmetricwidth}
@{\hspace{\longmetricgap}}
>{\centering\arraybackslash}p{\longmetricwidth}
@{\hspace{\longhalfmodelgap}}
@{\hspace{\longhalfmodelgap}}
>{\centering\arraybackslash}p{\longmetricwidth}
@{\hspace{\longmetricgap}}
>{\centering\arraybackslash}p{\longmetricwidth}
@{\hspace{\longhalfmodelgap}}
@{\hspace{\longhalfmodelgap}}
>{\centering\arraybackslash}p{\longmetricwidth}
@{\hspace{\longmetricgap}}
>{\centering\arraybackslash}p{\longmetricwidth}
}

\toprule

\multicolumn{1}{c}{Model}
& \longmodelhead{\textbf{Chameleon}}
& \longmodelhead{S-Mamba+ACN}
& \longmodelhead{TimeFilter}
& \longmodelhead{DUET}
& \longmodelhead{SFNN}
& \longlastmodelhead{FITS} \\[-0.5ex]

\cmidrule(lr){1-1}
\cmidrule(l{0pt}r{\longsegmentgap}){2-3}
\cmidrule(l{0pt}r{\longsegmentgap}){4-5}
\cmidrule(l{0pt}r{\longsegmentgap}){6-7}
\cmidrule(l{0pt}r{\longsegmentgap}){8-9}
\cmidrule(l{0pt}r{\longsegmentgap}){10-11}
\cmidrule(l{0pt}r){12-13}

\noalign{\vskip 0.5ex}

Dataset
& MSE & MAE
& MSE & MAE
& MSE & MAE
& MSE & MAE
& MSE & MAE
& MSE & MAE \\

\midrule

Traffic
& \longbest{0.3636} & \longbest{0.2409}
& 0.4012 & 0.2744
& \longsecond{0.3738} & 0.2580
& 0.3868 & \longsecond{0.2548}
& 0.3836 & 0.2656
& 0.4074 & 0.2783 \\

\midrule

Solar
& \longbest{0.1783} & \longsecond{0.2204}
& 0.1936 & 0.2531
& 0.2020 & 0.2438
& 0.1884 & \longbest{0.2150}
& \longsecond{0.1831} & 0.2434
& 0.2086 & 0.2495 \\

\midrule

Electricity
& \longbest{0.1522} & \longbest{0.2418}
& 0.1590 & 0.2576
& 0.1573 & 0.2541
& 0.1575 & 0.2469
& \longsecond{0.1554} & \longsecond{0.2452}
& 0.1609 & 0.2552 \\

\midrule

ETTh1
& \longbest{0.3870} & \longbest{0.4168}
& 0.4484 & 0.4538
& 0.4477 & 0.4431
& 0.4078 & 0.4214
& \longsecond{0.3964} & 0.4208
& 0.4087 & \longsecond{0.4207} \\

\midrule

ETTh2
& \longbest{0.3310} & \longbest{0.3773}
& 0.3850 & 0.4074
& 0.3928 & 0.4155
& 0.3440 & 0.3864
& 0.3481 & 0.3832
& \longsecond{0.3331} & \longsecond{0.3820} \\

\midrule

ETTm1
& \longbest{0.3298} & \longbest{0.3673}
& 0.3666 & 0.3952
& 0.3509 & 0.3812
& 0.3365 & 0.3694
& \longsecond{0.3329} & \longsecond{0.3687}
& 0.3501 & 0.3759 \\

\midrule

ETTm2
& \longbest{0.2356} & \longbest{0.2995}
& 0.2654 & 0.3252
& 0.2673 & 0.3213
& 0.2512 & 0.3108
& 0.2438 & \longsecond{0.3034}
& \longsecond{0.2416} & 0.3098 \\

\midrule


\multicolumn{1}{c}{Avg. Increase}
& -- & --
& \longincpct{11.3} & \longincpct{9.8}
& \longincpct{10.5} & \longincpct{7.2}
& \longincpct{4.8} & \longincpct{1.9}
& \longincpct{3.2} & \longincpct{3.8}
& \longincpct{7.1} & \longincpct{6.0} \\

\multicolumn{1}{c}{Max. Increase}
& -- & --
& \longincpct{16.3} & \longincpct{14.8}
& \longincpct{18.7} & \longincpct{10.6}
& \longincpct{6.6} & \longincpct{5.8}
& \longincpct{5.5} & \longincpct{10.4}
& \longincpct{17.0} & \longincpct{15.5} \\

\bottomrule

\end{tabular}%
}
\end{table*}


\newlength{\abIncreaseWidth}

\newcommand{\inc}[1]{%
  \makebox[\abIncreaseWidth][r]{+#1\%}%
}


\newlength{\abDatasetWidth}
\setlength{\abDatasetWidth}{1.20cm}

\newlength{\abSepGap}
\setlength{\abSepGap}{4pt}

\newlength{\abModelWidth}
\setlength{\abModelWidth}{\textwidth}

\addtolength{\abModelWidth}{-\abDatasetWidth}
\addtolength{\abModelWidth}{-\arrayrulewidth}
\addtolength{\abModelWidth}{-2\abSepGap}

\divide\abModelWidth by 6

\newcolumntype{D}{%
  >{\raggedright\arraybackslash}p{\abDatasetWidth}%
}

\newcolumntype{A}{%
  >{\centering\arraybackslash}p{\abModelWidth}%
}


\newcommand{\abhead}[1]{%
  \makebox[0pt][c]{%
    \scalebox{0.95}{\strut #1}%
  }%
}


\begin{table*}[t]
\centering
\caption{Ablations of the temporal backbone, preprocessing, and cross-variable measurement update. Chameleon reports horizon-averaged MSE. Other columns report the relative increase in this average over Chameleon. Chameleon (CI) removes the measurement update. w/o RevIN Pert and w/o Decomp remove stochastic RevIN perturbation and decomposition, respectively. GatedDeltaNet denotes the vanilla backbone without these three designs and with its original convolutional layers. w/ Mamba2 replaces GatedDeltaNet with Mamba2.}
\label{tab:ablation_main}

\begingroup
\small
\settowidth{\abIncreaseWidth}{+1.00\%}

\setlength{\tabcolsep}{0pt}

\begin{tabular}{
@{}
D
@{\hspace{\abSepGap}}|
@{\hspace{\abSepGap}}
A
A
A
A
A
A
@{}
}

\toprule

Dataset
& \abhead{\textbf{Chameleon}}
& \abhead{GatedDeltaNet}
& \abhead{Chameleon (CI)}
& \abhead{w/o RevIN Pert}
& \abhead{w/o Decomp}
& \abhead{w/ Mamba2} \\

\midrule

PEMS08
& 0.1141
& \inc{26.80}
& \inc{26.58}
& \inc{2.16}
& \inc{1.09}
& \inc{0.38} \\

Traffic
& 0.3636
& \inc{2.47}
& \inc{2.31}
& \inc{2.67}
& \inc{2.53}
& \inc{1.82} \\

Solar
& 0.1783
& \inc{7.20}
& \inc{3.89}
& \inc{1.07}
& \inc{3.05}
& \inc{3.76} \\

ETTh2
& 0.3310
& \inc{0.77}
& \inc{0.55}
& \inc{0.77}
& \inc{0.80}
& \inc{0.85} \\

ETTm2
& 0.2356
& \inc{1.00}
& \inc{0.04}
& \inc{0.77}
& \inc{0.27}
& \inc{0.83} \\

\bottomrule

\end{tabular}

\endgroup
\end{table*}

We further compare against vanilla GatedDeltaNet~\citep{yang2025}, replace GatedDeltaNet with Mamba2~\citep{dao2024}, and ablate the measurement update, RevIN perturbation, and decomposition in Table~\ref{tab:ablation_main}. Together with Table~\ref{tab:strong_results}, these results suggest that the measurement update's benefit broadly aligns with dependency strength. Specifically, its removal increases MSE by 26.58\% on PEMS08 and by larger margins on the ODE datasets. Among the benchmarks, Traffic and Solar have higher average Granger F-scores at lag 24 (6.3 and 5.5) than ETTh2 and ETTm2 (2.2 and 1.4). Removing the measurement update correspondingly leads to larger MSE increases of around 2--4\% on these two datasets. Even on ETTh2 and ETTm2, it still provides incremental improvements over the CI variant, with only a 3.3\% average increase in parameter count. These results support our objective of exploiting strong cross-variable dependencies without degrading the CI temporal backbone's performance when they are weak. Achieving this balance remains a longstanding challenge in MTSF~\citep{sun2025,abdelmalak2025}.

Meanwhile, removing RevIN perturbation or decomposition, or replacing GatedDeltaNet with Mamba2, also increases MSE by around 1--4\% on Traffic and Solar. Although RevIN perturbation and decomposition sometimes improve MSE only slightly, especially on ETTh2 and ETTm2, they reduce cross-run standard deviations by 64\% and 72\% on average, respectively, with modest computational overhead. Specifically, decomposition and using GatedDeltaNet instead of Mamba2 increase forward FLOPs by an average of 8.5\% and 4.4\% across all 36 settings, respectively, while perturbation introduces negligible computation. In addition, the comparison against vanilla GatedDeltaNet demonstrates that combining these designs provides benefits beyond backbone selection alone.

\begin{table*}[t]
\centering
\caption{Ablations of the cross-variable measurement update and comparisons with four alternatives. Chameleon, Chameleon (CI), and the reporting format follow Table~\ref{tab:ablation_main}. The $\gamma=0$ variant only retains learnable variable-wise correction with a diagonal gain. Attention, SSM, and static low-rank transformation replace the entire correction in Equation~\ref{eq:correction}, while dynamic low-rank replaces only the Kalman gain. Horizons follow Tables~\ref{tab:strong_results} and~\ref{tab:main_results} for Lorenz Coupled (LorenzC), PEMS08, and ETTh2. Solar uses $F\in\{96,336\}$.}
\label{tab:correction_ablation}
\begingroup
\small
\setlength{\tabcolsep}{0pt}
\newlength{\caDatasetWidth}
\setlength{\caDatasetWidth}{1.65cm}
\newlength{\caSepGap}
\setlength{\caSepGap}{4pt}
\newlength{\caModelWidth}
\setlength{\caModelWidth}{\dimexpr(\textwidth-\caDatasetWidth-2\caSepGap-\arrayrulewidth)/7\relax}
\newlength{\caIncreaseWidth}
\settowidth{\caIncreaseWidth}{+10.00\%}
\newcommand{\cainc}[1]{\makebox[\caIncreaseWidth][r]{+#1\%}}
\begin{tabular}{@{}>{\raggedright\arraybackslash}p{\caDatasetWidth}@{\hspace{\caSepGap}}|@{\hspace{\caSepGap}}*{7}{>{\centering\arraybackslash}p{\caModelWidth}}@{}}
\toprule
\makecell[l]{Dataset}
 & \makecell[c]{\textbf{Chameleon}} & \makecell[c]{Chameleon\\($\gamma=0$)} & \makecell[c]{Chameleon\\(CI)} & \makecell[c]{Attention} & \makecell[c]{SSM} & \makecell[c]{Static\\low-rank} & \makecell[c]{Dynamic\\low-rank} \\
\midrule
LorenzC
& 0.0874
& \cainc{113.53}
& \cainc{120.86}
& \cainc{19.87}
& \cainc{13.40}
& \cainc{20.96}
& \cainc{16.11} \\

PEMS08
& 0.1141
& \cainc{25.29}
& \cainc{26.58}
& \cainc{26.55}
& \cainc{8.75}
& \cainc{34.06}
& \cainc{3.80} \\

Solar
& 0.1717
& \cainc{4.77}
& \cainc{5.50}
& \cainc{3.53}
& \cainc{7.88}
& \cainc{3.20}
& \cainc{3.29} \\

ETTh2
& 0.3310
& \cainc{0.55}
& \cainc{0.55}
& \cainc{0.23}
& \cainc{1.33}
& \cainc{0.78}
& \cainc{0.38} \\
\bottomrule
\end{tabular}
\endgroup
\end{table*}

To isolate the contribution of cross-variable interactions, we fix $\gamma=0$, making $\mathbf{K}_l$ diagonal while retaining learnable variable-wise correction. We also replace the entire state correction in Equation~\ref{eq:correction} with variable-wise attention, a bidirectional Mamba2-style SSM, or a static low-rank transformation on the same GatedDeltaNet backbone. Each alternative mechanism receives the patch-wise prior states $\mathbf{S}_{l}^-$ and produces corrected states $\mathbf{S}_{l}$. The states $\mathbf{S}_{l}^-$ and $\mathbf{S}_{l}$ are then combined through the same learned $\omega$ as in Equation~\ref{eq:out}. The dynamic low-rank alternative instead replaces only $\mathbf{K}_l$ with an input- and prior-dependent diagonal-plus-low-rank operator, preserving the construction and use of $\mathbf{Z}_l$ in addition to $\omega$. We select alternative-specific dimensions for comparable computational costs, especially forward FLOPs. All other shared configurations and hyperparameters remain identical within each setting. Implementation details, computational costs, per-horizon MSE results, and statistical significance tests are reported in Appendix~\ref{app:results}.

Table~\ref{tab:correction_ablation} shows that Chameleon's MSE remains the lowest across four datasets with varying dependency strengths. The small differences on ETTh2 are expected given its weak dependencies. Consistently, the ETTh2 example in Appendix~\ref{app:results} has a small learned $\omega$ that suppresses the state correction. Variable-wise correction alone provides mostly incremental gains over the CI variant, while cross-variable interactions account for the major improvements on the more strongly dependent datasets. Chameleon's advantage over the alternatives further supports the empirical value of its structured measurement update, which separates variable-wise uncertainty from correlation and applies corresponding constraints.

\begin{figure*}[t]
\centering
\begin{subfigure}[t]{0.49\textwidth}
\centering
\includegraphics[width=\linewidth]{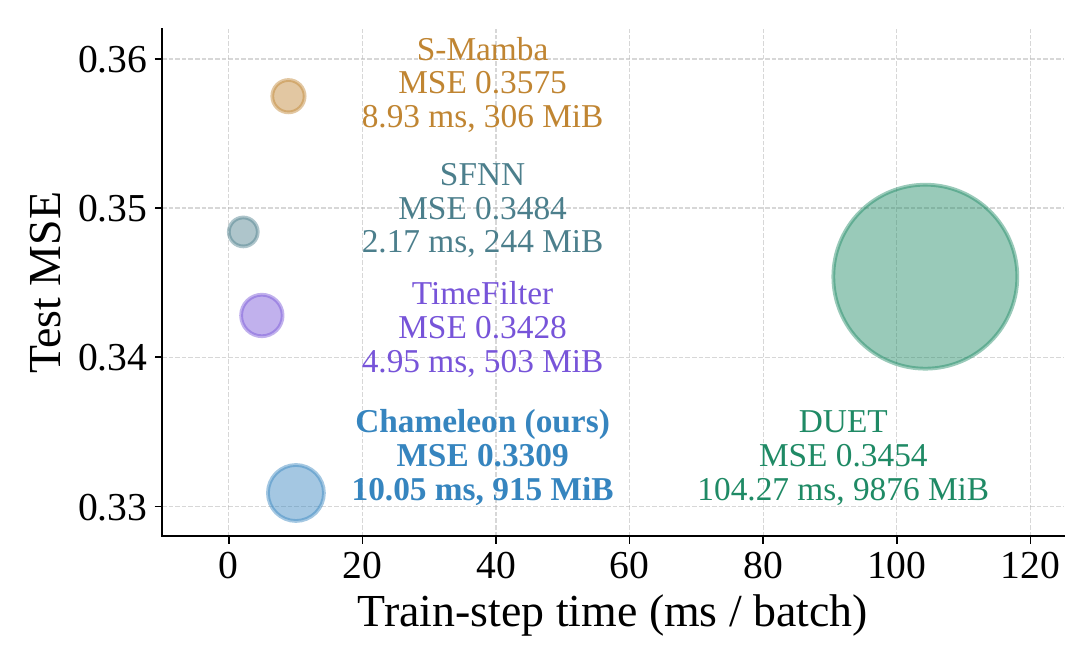}
\caption{Test MSE versus training time on Traffic.}
\label{fig:traffic_efficiency}
\end{subfigure}\hfill
\begin{subfigure}[t]{0.49\textwidth}
\centering
\includegraphics[width=\linewidth]{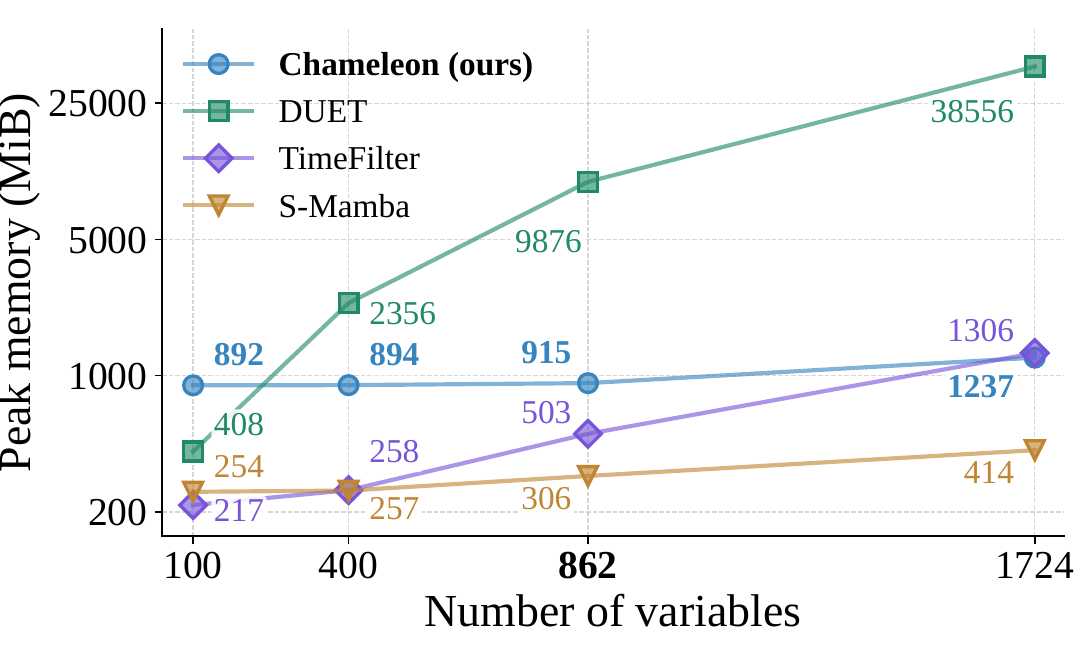}
\caption{Full-model peak memory versus variable count on Traffic.}
\label{fig:traffic_scalability}
\end{subfigure}

\begin{subfigure}[t]{0.49\textwidth}
\centering
\includegraphics[width=\linewidth]{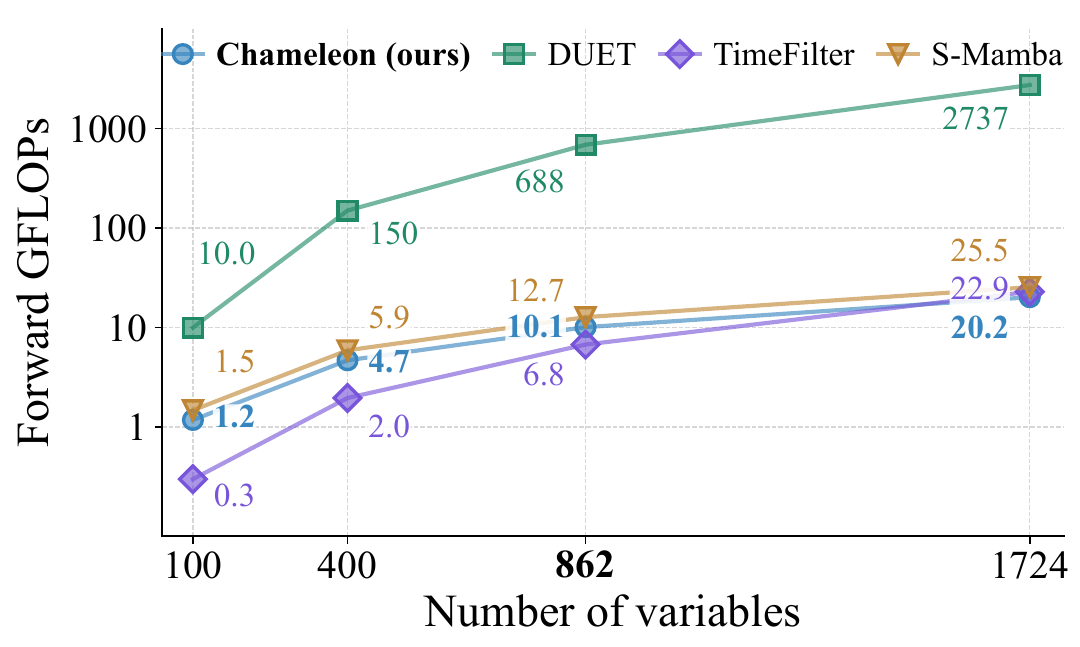}
\caption{Component FLOPs versus variable count on Traffic.}
\label{fig:traffic_component_flops}
\end{subfigure}\hfill
\begin{subfigure}[t]{0.49\textwidth}
\centering
\includegraphics[width=\linewidth]{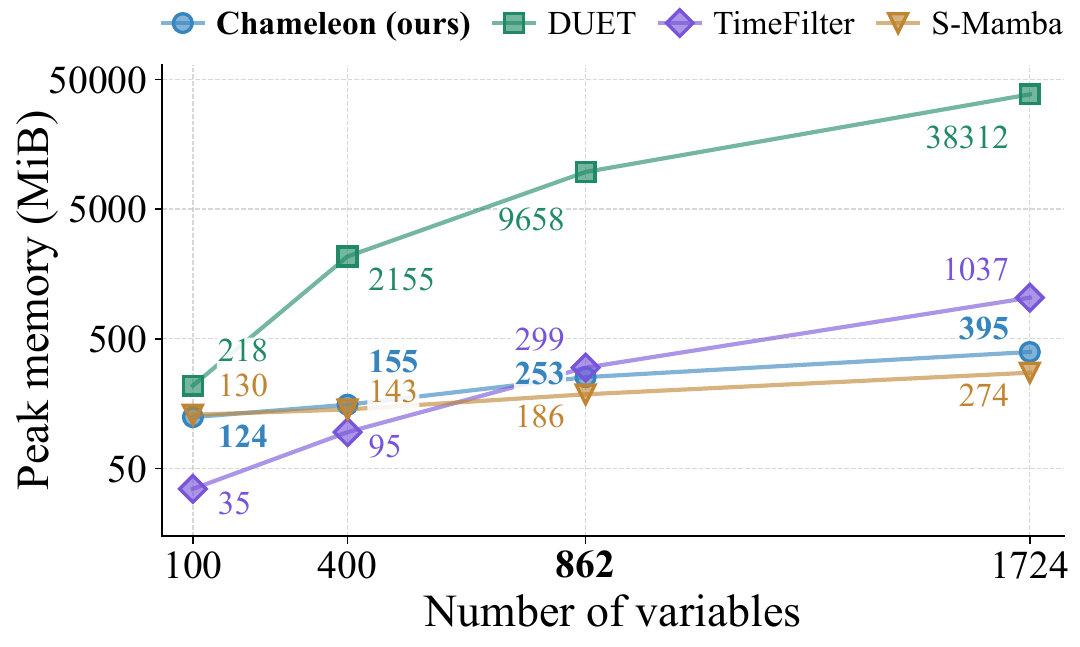}
\caption{Component peak memory versus variable count on Traffic.}
\label{fig:traffic_component_memory}
\end{subfigure}

\begin{subfigure}[t]{0.49\textwidth}
\centering
\includegraphics[width=\linewidth]{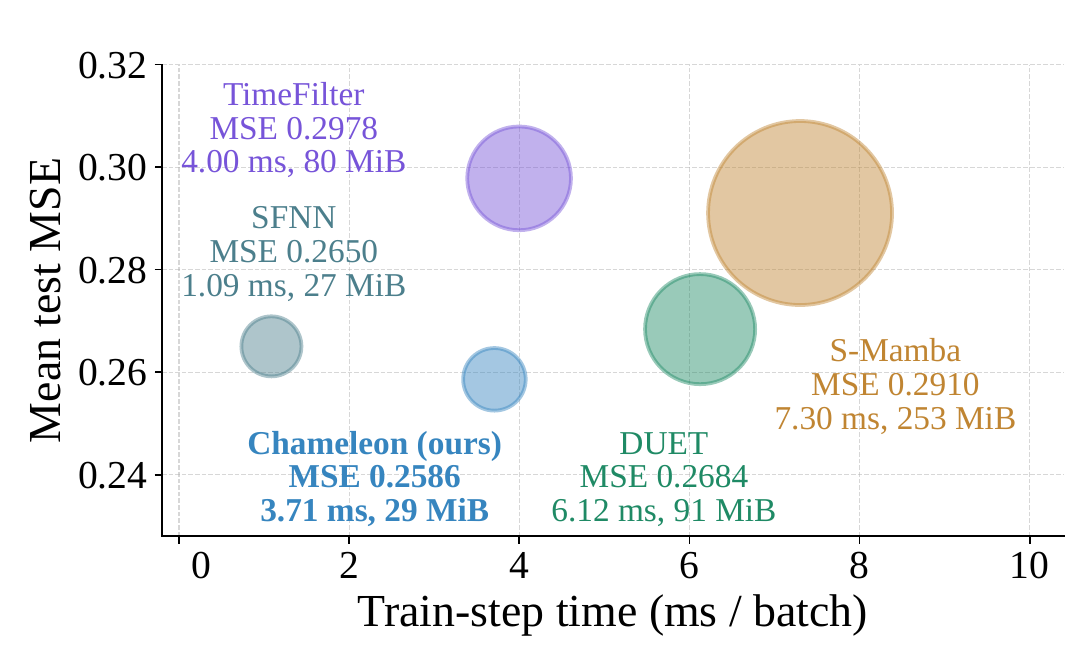}
\caption{Test MSE versus training time, averaged across four ETT datasets.}
\label{fig:ett_efficiency}
\end{subfigure}\hfill
\begin{subfigure}[t]{0.49\textwidth}
\centering
\includegraphics[width=\linewidth]{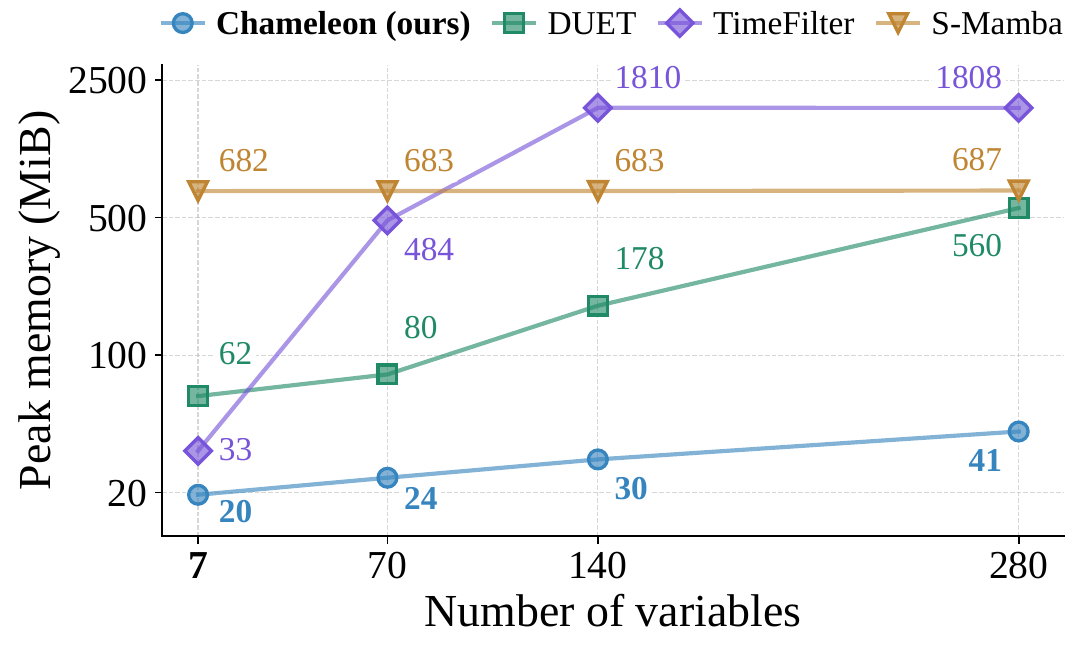}
\caption{Full-model peak memory versus variable count on ETTh2.}
\label{fig:ett_scalability}
\end{subfigure}
\caption{Training efficiency and scalability on Traffic and ETT with $F=96$ and a batch size of one. Bubble areas in (a) and (e) are proportional to peak memory. Subfigures (c) and (d) measure cross-variable modeling components across all layers in each model. Variable counts are varied by retaining subsets or duplicating variables.}
\label{fig:efficiency_scalability}
\end{figure*}

Finally, Figure~\ref{fig:efficiency_scalability} evaluates training efficiency and scalability on Traffic (862 variables) and ETT (seven variables) with $F=96$. On Traffic, Chameleon achieves the best MSE while keeping training time and peak memory within the same order of magnitude as TimeFilter~\citep{hu2025} and S-Mamba~\citep{wang2025}, and substantially below DUET~\citep{qiu2025}. Across the four ETT datasets, it also obtains the best average MSE with the least training time and peak memory among all evaluated CD methods. More importantly, the scalability analyses vary the number of variables by retaining subsets or duplicating all variables, reaching 1724 variables on Traffic. Figures~\ref{fig:traffic_scalability} and~\ref{fig:ett_scalability} show that Chameleon's peak memory scales comparably to S-Mamba on both Traffic and ETTh2, consistent with their theoretical $\mathcal{O}(N)$ complexity, whereas TimeFilter and DUET grow considerably faster. The additional Traffic analyses in Figures~\ref{fig:traffic_component_flops} and~\ref{fig:traffic_component_memory} isolate each model's cross-variable modeling components and reveal the same scaling pattern across methods in both forward FLOPs and peak memory. Parameter counts are further compared in Appendix~\ref{app:results}.

\section{Conclusion}
We presented Chameleon, a CD SSM that integrates a Kalman-filter-inspired cross-variable measurement update into a strong GatedDeltaNet-based temporal backbone, enabling data-dependent, fine-grained interactions with linear complexity in the number of variables. On strongly dependent ODE and PEMS datasets, Chameleon achieves the best MSE and MAE in every case, while removing the update raises MSE by 61\% on average. Across 28 standard benchmark settings, it also ranks first in 25 for MSE and 20 for MAE. Comparisons with alternative update mechanisms support the empirical value of the structured measurement-update formulation. Traffic and ETT analyses further show competitive training efficiency and favorable scalability across different variable counts. Together, these results suggest that Chameleon exploits strong cross-variable dependencies without sacrificing temporal modeling capacity or scalability across diverse MTSF settings.

\subsection*{AI use statement}
In this work, we used generative AI tools for partial code revision. We have not used generative AI tools for developing theoretical models or conceptual frameworks, proposing or refining hypotheses, designing or providing feedback on research methodology or experiments, implementing the full method, supporting qualitative or thematic data analysis, or interpreting the results, and assistance in formulating or proving mathematical claims and cleaning or reformatting datasets is not applicable to this work. Additionally, we used generative AI tools for visualizing the analyses and polishing the writing. We have reviewed all AI-assisted work and take responsibility for the final content of this work, including text, claims, or artifacts produced with the aid of generative AI.
%

\bibliographystyle{iclr2027_conference}
\bibliography{reference}
\clearpage

\appendix

\section{Full Experimental Results and Additional Analyses}
\label{app:results}

Tables~\ref{tab:benchmark_std}, \ref{tab:strong_results_std}, and \ref{tab:correction_ablation_std} report the complete means and standard deviations across runs, together with statistical significance for each dataset-horizon combination, corresponding to Tables~\ref{tab:main_results}, \ref{tab:strong_results}, and \ref{tab:correction_ablation}, respectively. In addition to achieving the best mean MSE and MAE in most cases, Chameleon generally exhibits lower or comparable cross-run standard deviations than the other CD baselines.


\newlength{\metricwidth}
\setlength{\metricwidth}{1.5cm}
\newlength{\metricgap}
\setlength{\metricgap}{4pt}
\newlength{\segmentgap}
\setlength{\segmentgap}{14pt}
\newlength{\halfmodelgap}
\setlength{\halfmodelgap}{7pt}
\newcolumntype{M}{>{\centering\arraybackslash}p{\metricwidth}}
\definecolor{highlightcolor}{HTML}{9ECED0}
\colorlet{bestlight}{highlightcolor!80}
\colorlet{secondlight}{highlightcolor!35}

\newlength{\highlightwidth}
\settowidth{\highlightwidth}{0.0000}

\begin{table*}[h]
\centering
\caption{Detailed MSE and MAE results (mean $\pm$ std) on long-horizon benchmarks. Highlighting and underlining follow Table~\ref{tab:strong_results}.}
\label{tab:benchmark_std}
\resizebox{1.0\textwidth}{!}{
\begin{tabular}{lr|MM@{\hspace{\halfmodelgap}}MM@{\hspace{\halfmodelgap}}MM@{\hspace{\halfmodelgap}}MM@{\hspace{\halfmodelgap}}MM@{\hspace{\halfmodelgap}}MM}
\toprule

\multicolumn{2}{c}{Model}
& \multicolumn{2}{c}{\textbf{Chameleon}}
& \multicolumn{2}{c}{S-Mamba+ACN}
& \multicolumn{2}{c}{TimeFilter}
& \multicolumn{2}{c}{DUET}
& \multicolumn{2}{c}{SFNN}
& \multicolumn{2}{c}{FITS} \\

\cmidrule(lr){1-2}
\cmidrule(l{0pt}r{\segmentgap}){3-4}
\cmidrule(l{0pt}r{\segmentgap}){5-6}
\cmidrule(l{0pt}r{\segmentgap}){7-8}
\cmidrule(l{0pt}r{\segmentgap}){9-10}
\cmidrule(l{0pt}r{\segmentgap}){11-12}
\cmidrule(l{0pt}r){13-14}

Dataset & (F)
& MSE & MAE
& MSE & MAE
& MSE & MAE
& MSE & MAE
& MSE & MAE
& MSE & MAE \\
\midrule

\multirow{8}{*}{Traffic} & 96 & \makebox[\metricwidth][c]{\phantom{$\pm$}\colorbox{bestlight}{\strut\underline{0.3309}}} & \makebox[\metricwidth][c]{\phantom{$\pm$}\colorbox{bestlight}{\strut\underline{0.2247}}} & \makebox[\metricwidth][c]{\phantom{$\pm$}0.3653} & \makebox[\metricwidth][c]{\phantom{$\pm$}0.2563} & \makebox[\metricwidth][c]{\phantom{$\pm$}\colorbox{secondlight}{\strut 0.3428}} & \makebox[\metricwidth][c]{\phantom{$\pm$}0.2412} & \makebox[\metricwidth][c]{\phantom{$\pm$}0.3493} & \makebox[\metricwidth][c]{\phantom{$\pm$}\colorbox{secondlight}{\strut 0.2367}} & \makebox[\metricwidth][c]{\phantom{$\pm$}0.3484} & \makebox[\metricwidth][c]{\phantom{$\pm$}0.2478} & \makebox[\metricwidth][c]{\phantom{$\pm$}0.3820} & \makebox[\metricwidth][c]{\phantom{$\pm$}0.2674} \\
& & \makebox[\metricwidth][c]{$\pm$0.0018} & \makebox[\metricwidth][c]{$\pm$0.0021} & \makebox[\metricwidth][c]{$\pm$0.0009} & \makebox[\metricwidth][c]{$\pm$0.0007} & \makebox[\metricwidth][c]{$\pm$0.0003} & \makebox[\metricwidth][c]{$\pm$0.0003} & \makebox[\metricwidth][c]{$\pm$0.0008} & \makebox[\metricwidth][c]{$\pm$0.0003} & \makebox[\metricwidth][c]{$\pm$0.0002} & \makebox[\metricwidth][c]{$\pm$0.0003} & \makebox[\metricwidth][c]{$\pm$0.0008} & \makebox[\metricwidth][c]{$\pm$0.0011} \\

& 192 & \makebox[\metricwidth][c]{\phantom{$\pm$}\colorbox{bestlight}{\strut\underline{0.3485}}} & \makebox[\metricwidth][c]{\phantom{$\pm$}\colorbox{bestlight}{\strut\underline{0.2328}}} & \makebox[\metricwidth][c]{\phantom{$\pm$}0.3755} & \makebox[\metricwidth][c]{\phantom{$\pm$}0.2664} & \makebox[\metricwidth][c]{\phantom{$\pm$}\colorbox{secondlight}{\strut 0.3609}} & \makebox[\metricwidth][c]{\phantom{$\pm$}0.2505} & \makebox[\metricwidth][c]{\phantom{$\pm$}0.3691} & \makebox[\metricwidth][c]{\phantom{$\pm$}\colorbox{secondlight}{\strut 0.2456}} & \makebox[\metricwidth][c]{\phantom{$\pm$}0.3697} & \makebox[\metricwidth][c]{\phantom{$\pm$}0.2582} & \makebox[\metricwidth][c]{\phantom{$\pm$}0.3944} & \makebox[\metricwidth][c]{\phantom{$\pm$}0.2713} \\
& & \makebox[\metricwidth][c]{$\pm$0.0026} & \makebox[\metricwidth][c]{$\pm$0.0014} & \makebox[\metricwidth][c]{$\pm$0.0029} & \makebox[\metricwidth][c]{$\pm$0.0019} & \makebox[\metricwidth][c]{$\pm$0.0003} & \makebox[\metricwidth][c]{$\pm$0.0002} & \makebox[\metricwidth][c]{$\pm$0.0010} & \makebox[\metricwidth][c]{$\pm$0.0002} & \makebox[\metricwidth][c]{$\pm$0.0031} & \makebox[\metricwidth][c]{$\pm$0.0013} & \makebox[\metricwidth][c]{$\pm$0.0003} & \makebox[\metricwidth][c]{$\pm$0.0003} \\

& 336 & \makebox[\metricwidth][c]{\phantom{$\pm$}\colorbox{bestlight}{\strut\underline{0.3651}}} & \makebox[\metricwidth][c]{\phantom{$\pm$}\colorbox{bestlight}{\strut\underline{0.2418}}} & \makebox[\metricwidth][c]{\phantom{$\pm$}0.4055} & \makebox[\metricwidth][c]{\phantom{$\pm$}0.2746} & \makebox[\metricwidth][c]{\phantom{$\pm$}\colorbox{secondlight}{\strut 0.3775}} & \makebox[\metricwidth][c]{\phantom{$\pm$}0.2600} & \makebox[\metricwidth][c]{\phantom{$\pm$}0.3922} & \makebox[\metricwidth][c]{\phantom{$\pm$}\colorbox{secondlight}{\strut 0.2592}} & \makebox[\metricwidth][c]{\phantom{$\pm$}0.3853} & \makebox[\metricwidth][c]{\phantom{$\pm$}0.2661} & \makebox[\metricwidth][c]{\phantom{$\pm$}0.4077} & \makebox[\metricwidth][c]{\phantom{$\pm$}0.2777} \\
& & \makebox[\metricwidth][c]{$\pm$0.0039} & \makebox[\metricwidth][c]{$\pm$0.0009} & \makebox[\metricwidth][c]{$\pm$0.0014} & \makebox[\metricwidth][c]{$\pm$0.0003} & \makebox[\metricwidth][c]{$\pm$0.0003} & \makebox[\metricwidth][c]{$\pm$0.0002} & \makebox[\metricwidth][c]{$\pm$0.0003} & \makebox[\metricwidth][c]{$\pm$0.0002} & \makebox[\metricwidth][c]{$\pm$0.0014} & \makebox[\metricwidth][c]{$\pm$0.0006} & \makebox[\metricwidth][c]{$\pm$0.0003} & \makebox[\metricwidth][c]{$\pm$0.0004} \\

& 720 & \makebox[\metricwidth][c]{\phantom{$\pm$}\colorbox{bestlight}{\strut\underline{0.4105}}} & \makebox[\metricwidth][c]{\phantom{$\pm$}\colorbox{bestlight}{\strut\underline{0.2647}}} & \makebox[\metricwidth][c]{\phantom{$\pm$}0.4587} & \makebox[\metricwidth][c]{\phantom{$\pm$}0.3002} & \makebox[\metricwidth][c]{\phantom{$\pm$}\colorbox{secondlight}{\strut 0.4141}} & \makebox[\metricwidth][c]{\phantom{$\pm$}0.2804} & \makebox[\metricwidth][c]{\phantom{$\pm$}0.4366} & \makebox[\metricwidth][c]{\phantom{$\pm$}\colorbox{secondlight}{\strut 0.2778}} & \makebox[\metricwidth][c]{\phantom{$\pm$}0.4310} & \makebox[\metricwidth][c]{\phantom{$\pm$}0.2903} & \makebox[\metricwidth][c]{\phantom{$\pm$}0.4454} & \makebox[\metricwidth][c]{\phantom{$\pm$}0.2969} \\
& & \makebox[\metricwidth][c]{$\pm$0.0047} & \makebox[\metricwidth][c]{$\pm$0.0026} & \makebox[\metricwidth][c]{$\pm$0.0045} & \makebox[\metricwidth][c]{$\pm$0.0016} & \makebox[\metricwidth][c]{$\pm$0.0004} & \makebox[\metricwidth][c]{$\pm$0.0003} & \makebox[\metricwidth][c]{$\pm$0.0038} & \makebox[\metricwidth][c]{$\pm$0.0002} & \makebox[\metricwidth][c]{$\pm$0.0010} & \makebox[\metricwidth][c]{$\pm$0.0007} & \makebox[\metricwidth][c]{$\pm$0.0004} & \makebox[\metricwidth][c]{$\pm$0.0006} \\

\midrule


\multirow{8}{*}{Solar} & 96 & \makebox[\metricwidth][c]{\phantom{$\pm$}\colorbox{bestlight}{\strut 0.1533}} & \makebox[\metricwidth][c]{\phantom{$\pm$}\colorbox{secondlight}{\strut 0.1979}} & \makebox[\metricwidth][c]{\phantom{$\pm$}0.1743} & \makebox[\metricwidth][c]{\phantom{$\pm$}0.2218} & \makebox[\metricwidth][c]{\phantom{$\pm$}0.1827} & \makebox[\metricwidth][c]{\phantom{$\pm$}0.2282} & \makebox[\metricwidth][c]{\phantom{$\pm$}0.1647} & \makebox[\metricwidth][c]{\phantom{$\pm$}\colorbox{bestlight}{\strut\underline{0.1875}}} & \makebox[\metricwidth][c]{\phantom{$\pm$}\colorbox{secondlight}{\strut 0.1553}} & \makebox[\metricwidth][c]{\phantom{$\pm$}0.2001} & \makebox[\metricwidth][c]{\phantom{$\pm$}0.1880} & \makebox[\metricwidth][c]{\phantom{$\pm$}0.2400} \\
& & \makebox[\metricwidth][c]{$\pm$0.0017} & \makebox[\metricwidth][c]{$\pm$0.0020} & \makebox[\metricwidth][c]{$\pm$0.0027} & \makebox[\metricwidth][c]{$\pm$0.0019} & \makebox[\metricwidth][c]{$\pm$0.0100} & \makebox[\metricwidth][c]{$\pm$0.0058} & \makebox[\metricwidth][c]{$\pm$0.0013} & \makebox[\metricwidth][c]{$\pm$0.0007} & \makebox[\metricwidth][c]{$\pm$0.0027} & \makebox[\metricwidth][c]{$\pm$0.0010} & \makebox[\metricwidth][c]{$\pm$0.0004} & \makebox[\metricwidth][c]{$\pm$0.0008} \\

& 192 & \makebox[\metricwidth][c]{\phantom{$\pm$}\colorbox{bestlight}{\strut\underline{0.1750}}} & \makebox[\metricwidth][c]{\phantom{$\pm$}\colorbox{secondlight}{\strut 0.2198}} & \makebox[\metricwidth][c]{\phantom{$\pm$}0.1867} & \makebox[\metricwidth][c]{\phantom{$\pm$}0.2519} & \makebox[\metricwidth][c]{\phantom{$\pm$}0.1974} & \makebox[\metricwidth][c]{\phantom{$\pm$}0.2396} & \makebox[\metricwidth][c]{\phantom{$\pm$}0.1864} & \makebox[\metricwidth][c]{\phantom{$\pm$}\colorbox{bestlight}{\strut 0.2190}} & \makebox[\metricwidth][c]{\phantom{$\pm$}\colorbox{secondlight}{\strut 0.1855}} & \makebox[\metricwidth][c]{\phantom{$\pm$}0.2521} & \makebox[\metricwidth][c]{\phantom{$\pm$}0.2059} & \makebox[\metricwidth][c]{\phantom{$\pm$}0.2483} \\
& & \makebox[\metricwidth][c]{$\pm$0.0029} & \makebox[\metricwidth][c]{$\pm$0.0019} & \makebox[\metricwidth][c]{$\pm$0.0035} & \makebox[\metricwidth][c]{$\pm$0.0046} & \makebox[\metricwidth][c]{$\pm$0.0140} & \makebox[\metricwidth][c]{$\pm$0.0097} & \makebox[\metricwidth][c]{$\pm$0.0044} & \makebox[\metricwidth][c]{$\pm$0.0002} & \makebox[\metricwidth][c]{$\pm$0.0019} & \makebox[\metricwidth][c]{$\pm$0.0045} & \makebox[\metricwidth][c]{$\pm$0.0002} & \makebox[\metricwidth][c]{$\pm$0.0006} \\

& 336 & \makebox[\metricwidth][c]{\phantom{$\pm$}\colorbox{bestlight}{\strut\underline{0.1899}}} & \makebox[\metricwidth][c]{\phantom{$\pm$}\colorbox{bestlight}{\strut\underline{0.2291}}} & \makebox[\metricwidth][c]{\phantom{$\pm$}0.1993} & \makebox[\metricwidth][c]{\phantom{$\pm$}0.2617} & \makebox[\metricwidth][c]{\phantom{$\pm$}0.2056} & \makebox[\metricwidth][c]{\phantom{$\pm$}0.2466} & \makebox[\metricwidth][c]{\phantom{$\pm$}0.1950} & \makebox[\metricwidth][c]{\phantom{$\pm$}\colorbox{secondlight}{\strut 0.2345}} & \makebox[\metricwidth][c]{\phantom{$\pm$}\colorbox{secondlight}{\strut 0.1935}} & \makebox[\metricwidth][c]{\phantom{$\pm$}0.2638} & \makebox[\metricwidth][c]{\phantom{$\pm$}0.2194} & \makebox[\metricwidth][c]{\phantom{$\pm$}0.2546} \\
& & \makebox[\metricwidth][c]{$\pm$0.0024} & \makebox[\metricwidth][c]{$\pm$0.0022} & \makebox[\metricwidth][c]{$\pm$0.0026} & \makebox[\metricwidth][c]{$\pm$0.0019} & \makebox[\metricwidth][c]{$\pm$0.0059} & \makebox[\metricwidth][c]{$\pm$0.0056} & \makebox[\metricwidth][c]{$\pm$0.0017} & \makebox[\metricwidth][c]{$\pm$0.0023} & \makebox[\metricwidth][c]{$\pm$0.0024} & \makebox[\metricwidth][c]{$\pm$0.0031} & \makebox[\metricwidth][c]{$\pm$0.0003} & \makebox[\metricwidth][c]{$\pm$0.0003} \\

& 720 & \makebox[\metricwidth][c]{\phantom{$\pm$}\colorbox{bestlight}{\strut\underline{0.1951}}} & \makebox[\metricwidth][c]{\phantom{$\pm$}\colorbox{secondlight}{\strut 0.2346}} & \makebox[\metricwidth][c]{\phantom{$\pm$}0.2141} & \makebox[\metricwidth][c]{\phantom{$\pm$}0.2771} & \makebox[\metricwidth][c]{\phantom{$\pm$}0.2225} & \makebox[\metricwidth][c]{\phantom{$\pm$}0.2606} & \makebox[\metricwidth][c]{\phantom{$\pm$}0.2077} & \makebox[\metricwidth][c]{\phantom{$\pm$}\colorbox{bestlight}{\strut\underline{0.2189}}} & \makebox[\metricwidth][c]{\phantom{$\pm$}\colorbox{secondlight}{\strut 0.1980}} & \makebox[\metricwidth][c]{\phantom{$\pm$}0.2574} & \makebox[\metricwidth][c]{\phantom{$\pm$}0.2211} & \makebox[\metricwidth][c]{\phantom{$\pm$}0.2552} \\
& & \makebox[\metricwidth][c]{$\pm$0.0019} & \makebox[\metricwidth][c]{$\pm$0.0018} & \makebox[\metricwidth][c]{$\pm$0.0021} & \makebox[\metricwidth][c]{$\pm$0.0032} & \makebox[\metricwidth][c]{$\pm$0.0151} & \makebox[\metricwidth][c]{$\pm$0.0096} & \makebox[\metricwidth][c]{$\pm$0.0015} & \makebox[\metricwidth][c]{$\pm$0.0006} & \makebox[\metricwidth][c]{$\pm$0.0011} & \makebox[\metricwidth][c]{$\pm$0.0015} & \makebox[\metricwidth][c]{$\pm$0.0001} & \makebox[\metricwidth][c]{$\pm$0.0003} \\

\midrule


\multirow{8}{*}{Electricity} & 96 & \makebox[\metricwidth][c]{\phantom{$\pm$}\colorbox{bestlight}{\strut\underline{0.1246}}} & \makebox[\metricwidth][c]{\phantom{$\pm$}\colorbox{bestlight}{\strut\underline{0.2146}}} & \makebox[\metricwidth][c]{\phantom{$\pm$}0.1339} & \makebox[\metricwidth][c]{\phantom{$\pm$}0.2311} & \makebox[\metricwidth][c]{\phantom{$\pm$}0.1282} & \makebox[\metricwidth][c]{\phantom{$\pm$}0.2220} & \makebox[\metricwidth][c]{\phantom{$\pm$}0.1275} & \makebox[\metricwidth][c]{\phantom{$\pm$}0.2179} & \makebox[\metricwidth][c]{\phantom{$\pm$}\colorbox{secondlight}{\strut 0.1261}} & \makebox[\metricwidth][c]{\phantom{$\pm$}\colorbox{secondlight}{\strut 0.2166}} & \makebox[\metricwidth][c]{\phantom{$\pm$}0.1323} & \makebox[\metricwidth][c]{\phantom{$\pm$}0.2291} \\
& & \makebox[\metricwidth][c]{$\pm$0.0001} & \makebox[\metricwidth][c]{$\pm$0.0001} & \makebox[\metricwidth][c]{$\pm$0.0026} & \makebox[\metricwidth][c]{$\pm$0.0017} & \makebox[\metricwidth][c]{$\pm$0.0002} & \makebox[\metricwidth][c]{$\pm$0.0001} & \makebox[\metricwidth][c]{$\pm$0.0002} & \makebox[\metricwidth][c]{$\pm$0.0002} & \makebox[\metricwidth][c]{$\pm$0.0003} & \makebox[\metricwidth][c]{$\pm$0.0003} & \makebox[\metricwidth][c]{$\pm$0.0001} & \makebox[\metricwidth][c]{$\pm$0.0001} \\

& 192 & \makebox[\metricwidth][c]{\phantom{$\pm$}\colorbox{bestlight}{\strut\underline{0.1432}}} & \makebox[\metricwidth][c]{\phantom{$\pm$}\colorbox{bestlight}{\strut\underline{0.2324}}} & \makebox[\metricwidth][c]{\phantom{$\pm$}0.1534} & \makebox[\metricwidth][c]{\phantom{$\pm$}0.2486} & \makebox[\metricwidth][c]{\phantom{$\pm$}0.1484} & \makebox[\metricwidth][c]{\phantom{$\pm$}0.2413} & \makebox[\metricwidth][c]{\phantom{$\pm$}0.1467} & \makebox[\metricwidth][c]{\phantom{$\pm$}0.2355} & \makebox[\metricwidth][c]{\phantom{$\pm$}\colorbox{secondlight}{\strut 0.1448}} & \makebox[\metricwidth][c]{\phantom{$\pm$}\colorbox{secondlight}{\strut 0.2337}} & \makebox[\metricwidth][c]{\phantom{$\pm$}0.1472} & \makebox[\metricwidth][c]{\phantom{$\pm$}0.2424} \\
& & \makebox[\metricwidth][c]{$\pm$0.0001} & \makebox[\metricwidth][c]{$\pm$0.0001} & \makebox[\metricwidth][c]{$\pm$0.0012} & \makebox[\metricwidth][c]{$\pm$0.0010} & \makebox[\metricwidth][c]{$\pm$0.0006} & \makebox[\metricwidth][c]{$\pm$0.0001} & \makebox[\metricwidth][c]{$\pm$0.0002} & \makebox[\metricwidth][c]{$\pm$0.0002} & \makebox[\metricwidth][c]{$\pm$0.0002} & \makebox[\metricwidth][c]{$\pm$0.0002} & \makebox[\metricwidth][c]{$\pm$0.0000} & \makebox[\metricwidth][c]{$\pm$0.0001} \\

& 336 & \makebox[\metricwidth][c]{\phantom{$\pm$}\colorbox{secondlight}{\strut 0.1566}} & \makebox[\metricwidth][c]{\phantom{$\pm$}\colorbox{bestlight}{\strut\underline{0.2475}}} & \makebox[\metricwidth][c]{\phantom{$\pm$}0.1637} & \makebox[\metricwidth][c]{\phantom{$\pm$}0.2645} & \makebox[\metricwidth][c]{\phantom{$\pm$}\colorbox{bestlight}{\strut\underline{0.1554}}} & \makebox[\metricwidth][c]{\phantom{$\pm$}0.2598} & \makebox[\metricwidth][c]{\phantom{$\pm$}0.1623} & \makebox[\metricwidth][c]{\phantom{$\pm$}0.2518} & \makebox[\metricwidth][c]{\phantom{$\pm$}0.1601} & \makebox[\metricwidth][c]{\phantom{$\pm$}\colorbox{secondlight}{\strut 0.2503}} & \makebox[\metricwidth][c]{\phantom{$\pm$}0.1631} & \makebox[\metricwidth][c]{\phantom{$\pm$}0.2586} \\
& & \makebox[\metricwidth][c]{$\pm$0.0001} & \makebox[\metricwidth][c]{$\pm$0.0000} & \makebox[\metricwidth][c]{$\pm$0.0023} & \makebox[\metricwidth][c]{$\pm$0.0027} & \makebox[\metricwidth][c]{$\pm$0.0006} & \makebox[\metricwidth][c]{$\pm$0.0001} & \makebox[\metricwidth][c]{$\pm$0.0005} & \makebox[\metricwidth][c]{$\pm$0.0006} & \makebox[\metricwidth][c]{$\pm$0.0003} & \makebox[\metricwidth][c]{$\pm$0.0001} & \makebox[\metricwidth][c]{$\pm$0.0000} & \makebox[\metricwidth][c]{$\pm$0.0001} \\

& 720 & \makebox[\metricwidth][c]{\phantom{$\pm$}\colorbox{bestlight}{\strut 0.1842}} & \makebox[\metricwidth][c]{\phantom{$\pm$}\colorbox{bestlight}{\strut\underline{0.2727}}} & \makebox[\metricwidth][c]{\phantom{$\pm$}\colorbox{secondlight}{\strut 0.1849}} & \makebox[\metricwidth][c]{\phantom{$\pm$}0.2861} & \makebox[\metricwidth][c]{\phantom{$\pm$}0.1972} & \makebox[\metricwidth][c]{\phantom{$\pm$}0.2932} & \makebox[\metricwidth][c]{\phantom{$\pm$}0.1936} & \makebox[\metricwidth][c]{\phantom{$\pm$}0.2825} & \makebox[\metricwidth][c]{\phantom{$\pm$}0.1908} & \makebox[\metricwidth][c]{\phantom{$\pm$}\colorbox{secondlight}{\strut 0.2802}} & \makebox[\metricwidth][c]{\phantom{$\pm$}0.2012} & \makebox[\metricwidth][c]{\phantom{$\pm$}0.2906} \\
& & \makebox[\metricwidth][c]{$\pm$0.0029} & \makebox[\metricwidth][c]{$\pm$0.0026} & \makebox[\metricwidth][c]{$\pm$0.0026} & \makebox[\metricwidth][c]{$\pm$0.0024} & \makebox[\metricwidth][c]{$\pm$0.0005} & \makebox[\metricwidth][c]{$\pm$0.0006} & \makebox[\metricwidth][c]{$\pm$0.0010} & \makebox[\metricwidth][c]{$\pm$0.0009} & \makebox[\metricwidth][c]{$\pm$0.0002} & \makebox[\metricwidth][c]{$\pm$0.0004} & \makebox[\metricwidth][c]{$\pm$0.0000} & \makebox[\metricwidth][c]{$\pm$0.0001} \\

\midrule


\multirow{8}{*}{ETTh1} & 96 & \makebox[\metricwidth][c]{\phantom{$\pm$}\colorbox{bestlight}{\strut\underline{0.3428}}} & \makebox[\metricwidth][c]{\phantom{$\pm$}\colorbox{secondlight}{\strut 0.3844}} & \makebox[\metricwidth][c]{\phantom{$\pm$}0.3889} & \makebox[\metricwidth][c]{\phantom{$\pm$}0.4082} & \makebox[\metricwidth][c]{\phantom{$\pm$}0.3801} & \makebox[\metricwidth][c]{\phantom{$\pm$}0.3992} & \makebox[\metricwidth][c]{\phantom{$\pm$}0.3579} & \makebox[\metricwidth][c]{\phantom{$\pm$}\colorbox{bestlight}{\strut\underline{0.3830}}} & \makebox[\metricwidth][c]{\phantom{$\pm$}\colorbox{secondlight}{\strut 0.3503}} & \makebox[\metricwidth][c]{\phantom{$\pm$}0.3846} & \makebox[\metricwidth][c]{\phantom{$\pm$}0.3739} & \makebox[\metricwidth][c]{\phantom{$\pm$}0.3952} \\
& & \makebox[\metricwidth][c]{$\pm$0.0003} & \makebox[\metricwidth][c]{$\pm$0.0003} & \makebox[\metricwidth][c]{$\pm$0.0015} & \makebox[\metricwidth][c]{$\pm$0.0005} & \makebox[\metricwidth][c]{$\pm$0.0040} & \makebox[\metricwidth][c]{$\pm$0.0022} & \makebox[\metricwidth][c]{$\pm$0.0010} & \makebox[\metricwidth][c]{$\pm$0.0009} & \makebox[\metricwidth][c]{$\pm$0.0004} & \makebox[\metricwidth][c]{$\pm$0.0004} & \makebox[\metricwidth][c]{$\pm$0.0001} & \makebox[\metricwidth][c]{$\pm$0.0001} \\

& 192 & \makebox[\metricwidth][c]{\phantom{$\pm$}\colorbox{bestlight}{\strut\underline{0.3808}}} & \makebox[\metricwidth][c]{\phantom{$\pm$}\colorbox{secondlight}{\strut 0.4095}} & \makebox[\metricwidth][c]{\phantom{$\pm$}0.4242} & \makebox[\metricwidth][c]{\phantom{$\pm$}0.4470} & \makebox[\metricwidth][c]{\phantom{$\pm$}0.4365} & \makebox[\metricwidth][c]{\phantom{$\pm$}0.4333} & \makebox[\metricwidth][c]{\phantom{$\pm$}0.3986} & \makebox[\metricwidth][c]{\phantom{$\pm$}0.4098} & \makebox[\metricwidth][c]{\phantom{$\pm$}\colorbox{secondlight}{\strut 0.3880}} & \makebox[\metricwidth][c]{\phantom{$\pm$}\colorbox{bestlight}{\strut\underline{0.4087}}} & \makebox[\metricwidth][c]{\phantom{$\pm$}0.4065} & \makebox[\metricwidth][c]{\phantom{$\pm$}0.4138} \\
& & \makebox[\metricwidth][c]{$\pm$0.0006} & \makebox[\metricwidth][c]{$\pm$0.0004} & \makebox[\metricwidth][c]{$\pm$0.0011} & \makebox[\metricwidth][c]{$\pm$0.0007} & \makebox[\metricwidth][c]{$\pm$0.0078} & \makebox[\metricwidth][c]{$\pm$0.0039} & \makebox[\metricwidth][c]{$\pm$0.0010} & \makebox[\metricwidth][c]{$\pm$0.0010} & \makebox[\metricwidth][c]{$\pm$0.0005} & \makebox[\metricwidth][c]{$\pm$0.0003} & \makebox[\metricwidth][c]{$\pm$0.0001} & \makebox[\metricwidth][c]{$\pm$0.0001} \\

& 336 & \makebox[\metricwidth][c]{\phantom{$\pm$}\colorbox{bestlight}{\strut\underline{0.4083}}} & \makebox[\metricwidth][c]{\phantom{$\pm$}\colorbox{secondlight}{\strut 0.4267}} & \makebox[\metricwidth][c]{\phantom{$\pm$}0.4754} & \makebox[\metricwidth][c]{\phantom{$\pm$}0.4586} & \makebox[\metricwidth][c]{\phantom{$\pm$}0.4859} & \makebox[\metricwidth][c]{\phantom{$\pm$}0.4612} & \makebox[\metricwidth][c]{\phantom{$\pm$}0.4245} & \makebox[\metricwidth][c]{\phantom{$\pm$}0.4274} & \makebox[\metricwidth][c]{\phantom{$\pm$}\colorbox{secondlight}{\strut 0.4119}} & \makebox[\metricwidth][c]{\phantom{$\pm$}\colorbox{bestlight}{\strut\underline{0.4241}}} & \makebox[\metricwidth][c]{\phantom{$\pm$}0.4294} & \makebox[\metricwidth][c]{\phantom{$\pm$}0.4276} \\
& & \makebox[\metricwidth][c]{$\pm$0.0006} & \makebox[\metricwidth][c]{$\pm$0.0005} & \makebox[\metricwidth][c]{$\pm$0.0028} & \makebox[\metricwidth][c]{$\pm$0.0014} & \makebox[\metricwidth][c]{$\pm$0.0112} & \makebox[\metricwidth][c]{$\pm$0.0052} & \makebox[\metricwidth][c]{$\pm$0.0016} & \makebox[\metricwidth][c]{$\pm$0.0018} & \makebox[\metricwidth][c]{$\pm$0.0025} & \makebox[\metricwidth][c]{$\pm$0.0012} & \makebox[\metricwidth][c]{$\pm$0.0001} & \makebox[\metricwidth][c]{$\pm$0.0002} \\

& 720 & \makebox[\metricwidth][c]{\phantom{$\pm$}\colorbox{bestlight}{\strut\underline{0.4162}}} & \makebox[\metricwidth][c]{\phantom{$\pm$}\colorbox{secondlight}{\strut 0.4468}} & \makebox[\metricwidth][c]{\phantom{$\pm$}0.5051} & \makebox[\metricwidth][c]{\phantom{$\pm$}0.5016} & \makebox[\metricwidth][c]{\phantom{$\pm$}0.4884} & \makebox[\metricwidth][c]{\phantom{$\pm$}0.4788} & \makebox[\metricwidth][c]{\phantom{$\pm$}0.4503} & \makebox[\metricwidth][c]{\phantom{$\pm$}0.4656} & \makebox[\metricwidth][c]{\phantom{$\pm$}0.4355} & \makebox[\metricwidth][c]{\phantom{$\pm$}0.4660} & \makebox[\metricwidth][c]{\phantom{$\pm$}\colorbox{secondlight}{\strut 0.4249}} & \makebox[\metricwidth][c]{\phantom{$\pm$}\colorbox{bestlight}{\strut\underline{0.4462}}} \\
& & \makebox[\metricwidth][c]{$\pm$0.0010} & \makebox[\metricwidth][c]{$\pm$0.0004} & \makebox[\metricwidth][c]{$\pm$0.0147} & \makebox[\metricwidth][c]{$\pm$0.0083} & \makebox[\metricwidth][c]{$\pm$0.0274} & \makebox[\metricwidth][c]{$\pm$0.0135} & \makebox[\metricwidth][c]{$\pm$0.0159} & \makebox[\metricwidth][c]{$\pm$0.0104} & \makebox[\metricwidth][c]{$\pm$0.0010} & \makebox[\metricwidth][c]{$\pm$0.0016} & \makebox[\metricwidth][c]{$\pm$0.0002} & \makebox[\metricwidth][c]{$\pm$0.0003} \\

\midrule


\multirow{8}{*}{ETTh2} & 96 & \makebox[\metricwidth][c]{\phantom{$\pm$}\colorbox{bestlight}{\strut\underline{0.2628}}} & \makebox[\metricwidth][c]{\phantom{$\pm$}\colorbox{bestlight}{\strut\underline{0.3271}}} & \makebox[\metricwidth][c]{\phantom{$\pm$}0.3016} & \makebox[\metricwidth][c]{\phantom{$\pm$}0.3510} & \makebox[\metricwidth][c]{\phantom{$\pm$}0.3110} & \makebox[\metricwidth][c]{\phantom{$\pm$}0.3551} & \makebox[\metricwidth][c]{\phantom{$\pm$}0.2846} & \makebox[\metricwidth][c]{\phantom{$\pm$}0.3356} & \makebox[\metricwidth][c]{\phantom{$\pm$}0.2734} & \makebox[\metricwidth][c]{\phantom{$\pm$}\colorbox{secondlight}{\strut 0.3315}} & \makebox[\metricwidth][c]{\phantom{$\pm$}\colorbox{secondlight}{\strut 0.2707}} & \makebox[\metricwidth][c]{\phantom{$\pm$}0.3359} \\
& & \makebox[\metricwidth][c]{$\pm$0.0004} & \makebox[\metricwidth][c]{$\pm$0.0005} & \makebox[\metricwidth][c]{$\pm$0.0082} & \makebox[\metricwidth][c]{$\pm$0.0045} & \makebox[\metricwidth][c]{$\pm$0.0173} & \makebox[\metricwidth][c]{$\pm$0.0148} & \makebox[\metricwidth][c]{$\pm$0.0022} & \makebox[\metricwidth][c]{$\pm$0.0016} & \makebox[\metricwidth][c]{$\pm$0.0032} & \makebox[\metricwidth][c]{$\pm$0.0019} & \makebox[\metricwidth][c]{$\pm$0.0002} & \makebox[\metricwidth][c]{$\pm$0.0001} \\

& 192 & \makebox[\metricwidth][c]{\phantom{$\pm$}\colorbox{bestlight}{\strut\underline{0.3250}}} & \makebox[\metricwidth][c]{\phantom{$\pm$}\colorbox{bestlight}{\strut\underline{0.3688}}} & \makebox[\metricwidth][c]{\phantom{$\pm$}0.3887} & \makebox[\metricwidth][c]{\phantom{$\pm$}0.4028} & \makebox[\metricwidth][c]{\phantom{$\pm$}0.3824} & \makebox[\metricwidth][c]{\phantom{$\pm$}0.4018} & \makebox[\metricwidth][c]{\phantom{$\pm$}0.3333} & \makebox[\metricwidth][c]{\phantom{$\pm$}0.3758} & \makebox[\metricwidth][c]{\phantom{$\pm$}0.3392} & \makebox[\metricwidth][c]{\phantom{$\pm$}0.3743} & \makebox[\metricwidth][c]{\phantom{$\pm$}\colorbox{secondlight}{\strut 0.3305}} & \makebox[\metricwidth][c]{\phantom{$\pm$}\colorbox{secondlight}{\strut 0.3742}} \\
& & \makebox[\metricwidth][c]{$\pm$0.0007} & \makebox[\metricwidth][c]{$\pm$0.0003} & \makebox[\metricwidth][c]{$\pm$0.0061} & \makebox[\metricwidth][c]{$\pm$0.0028} & \makebox[\metricwidth][c]{$\pm$0.0096} & \makebox[\metricwidth][c]{$\pm$0.0062} & \makebox[\metricwidth][c]{$\pm$0.0012} & \makebox[\metricwidth][c]{$\pm$0.0007} & \makebox[\metricwidth][c]{$\pm$0.0052} & \makebox[\metricwidth][c]{$\pm$0.0032} & \makebox[\metricwidth][c]{$\pm$0.0001} & \makebox[\metricwidth][c]{$\pm$0.0001} \\

& 336 & \makebox[\metricwidth][c]{\phantom{$\pm$}\colorbox{bestlight}{\strut\underline{0.3523}}} & \makebox[\metricwidth][c]{\phantom{$\pm$}\colorbox{bestlight}{\strut\underline{0.3915}}} & \makebox[\metricwidth][c]{\phantom{$\pm$}0.4089} & \makebox[\metricwidth][c]{\phantom{$\pm$}0.4254} & \makebox[\metricwidth][c]{\phantom{$\pm$}0.4584} & \makebox[\metricwidth][c]{\phantom{$\pm$}0.4639} & \makebox[\metricwidth][c]{\phantom{$\pm$}0.3662} & \makebox[\metricwidth][c]{\phantom{$\pm$}0.4038} & \makebox[\metricwidth][c]{\phantom{$\pm$}0.3817} & \makebox[\metricwidth][c]{\phantom{$\pm$}0.4036} & \makebox[\metricwidth][c]{\phantom{$\pm$}\colorbox{secondlight}{\strut 0.3541}} & \makebox[\metricwidth][c]{\phantom{$\pm$}\colorbox{secondlight}{\strut 0.3955}} \\
& & \makebox[\metricwidth][c]{$\pm$0.0005} & \makebox[\metricwidth][c]{$\pm$0.0004} & \makebox[\metricwidth][c]{$\pm$0.0080} & \makebox[\metricwidth][c]{$\pm$0.0037} & \makebox[\metricwidth][c]{$\pm$0.0149} & \makebox[\metricwidth][c]{$\pm$0.0107} & \makebox[\metricwidth][c]{$\pm$0.0014} & \makebox[\metricwidth][c]{$\pm$0.0011} & \makebox[\metricwidth][c]{$\pm$0.0014} & \makebox[\metricwidth][c]{$\pm$0.0010} & \makebox[\metricwidth][c]{$\pm$0.0001} & \makebox[\metricwidth][c]{$\pm$0.0000} \\

& 720 & \makebox[\metricwidth][c]{\phantom{$\pm$}\colorbox{secondlight}{\strut 0.3837}} & \makebox[\metricwidth][c]{\phantom{$\pm$}\colorbox{bestlight}{\strut 0.4219}} & \makebox[\metricwidth][c]{\phantom{$\pm$}0.4410} & \makebox[\metricwidth][c]{\phantom{$\pm$}0.4505} & \makebox[\metricwidth][c]{\phantom{$\pm$}0.4195} & \makebox[\metricwidth][c]{\phantom{$\pm$}0.4411} & \makebox[\metricwidth][c]{\phantom{$\pm$}0.3917} & \makebox[\metricwidth][c]{\phantom{$\pm$}0.4306} & \makebox[\metricwidth][c]{\phantom{$\pm$}0.3981} & \makebox[\metricwidth][c]{\phantom{$\pm$}0.4235} & \makebox[\metricwidth][c]{\phantom{$\pm$}\colorbox{bestlight}{\strut\underline{0.3772}}} & \makebox[\metricwidth][c]{\phantom{$\pm$}\colorbox{secondlight}{\strut 0.4222}} \\
& & \makebox[\metricwidth][c]{$\pm$0.0023} & \makebox[\metricwidth][c]{$\pm$0.0010} & \makebox[\metricwidth][c]{$\pm$0.0077} & \makebox[\metricwidth][c]{$\pm$0.0043} & \makebox[\metricwidth][c]{$\pm$0.0047} & \makebox[\metricwidth][c]{$\pm$0.0030} & \makebox[\metricwidth][c]{$\pm$0.0049} & \makebox[\metricwidth][c]{$\pm$0.0029} & \makebox[\metricwidth][c]{$\pm$0.0023} & \makebox[\metricwidth][c]{$\pm$0.0015} & \makebox[\metricwidth][c]{$\pm$0.0002} & \makebox[\metricwidth][c]{$\pm$0.0001} \\

\midrule


\multirow{8}{*}{ETTm1} & 96 & \makebox[\metricwidth][c]{\phantom{$\pm$}\colorbox{bestlight}{\strut\underline{0.2755}}} & \makebox[\metricwidth][c]{\phantom{$\pm$}\colorbox{secondlight}{\strut 0.3309}} & \makebox[\metricwidth][c]{\phantom{$\pm$}0.3067} & \makebox[\metricwidth][c]{\phantom{$\pm$}0.3578} & \makebox[\metricwidth][c]{\phantom{$\pm$}0.2942} & \makebox[\metricwidth][c]{\phantom{$\pm$}0.3432} & \makebox[\metricwidth][c]{\phantom{$\pm$}0.2905} & \makebox[\metricwidth][c]{\phantom{$\pm$}0.3332} & \makebox[\metricwidth][c]{\phantom{$\pm$}\colorbox{secondlight}{\strut 0.2790}} & \makebox[\metricwidth][c]{\phantom{$\pm$}\colorbox{bestlight}{\strut 0.3306}} & \makebox[\metricwidth][c]{\phantom{$\pm$}0.3039} & \makebox[\metricwidth][c]{\phantom{$\pm$}0.3455} \\
& & \makebox[\metricwidth][c]{$\pm$0.0007} & \makebox[\metricwidth][c]{$\pm$0.0004} & \makebox[\metricwidth][c]{$\pm$0.0046} & \makebox[\metricwidth][c]{$\pm$0.0027} & \makebox[\metricwidth][c]{$\pm$0.0009} & \makebox[\metricwidth][c]{$\pm$0.0006} & \makebox[\metricwidth][c]{$\pm$0.0003} & \makebox[\metricwidth][c]{$\pm$0.0003} & \makebox[\metricwidth][c]{$\pm$0.0003} & \makebox[\metricwidth][c]{$\pm$0.0002} & \makebox[\metricwidth][c]{$\pm$0.0002} & \makebox[\metricwidth][c]{$\pm$0.0002} \\

& 192 & \makebox[\metricwidth][c]{\phantom{$\pm$}\colorbox{bestlight}{\strut\underline{0.3133}}} & \makebox[\metricwidth][c]{\phantom{$\pm$}\colorbox{bestlight}{\strut\underline{0.3570}}} & \makebox[\metricwidth][c]{\phantom{$\pm$}0.3451} & \makebox[\metricwidth][c]{\phantom{$\pm$}0.3807} & \makebox[\metricwidth][c]{\phantom{$\pm$}0.3289} & \makebox[\metricwidth][c]{\phantom{$\pm$}0.3680} & \makebox[\metricwidth][c]{\phantom{$\pm$}\colorbox{secondlight}{\strut 0.3188}} & \makebox[\metricwidth][c]{\phantom{$\pm$}0.3592} & \makebox[\metricwidth][c]{\phantom{$\pm$}0.3197} & \makebox[\metricwidth][c]{\phantom{$\pm$}\colorbox{secondlight}{\strut 0.3585}} & \makebox[\metricwidth][c]{\phantom{$\pm$}0.3372} & \makebox[\metricwidth][c]{\phantom{$\pm$}0.3649} \\
& & \makebox[\metricwidth][c]{$\pm$0.0014} & \makebox[\metricwidth][c]{$\pm$0.0005} & \makebox[\metricwidth][c]{$\pm$0.0010} & \makebox[\metricwidth][c]{$\pm$0.0009} & \makebox[\metricwidth][c]{$\pm$0.0017} & \makebox[\metricwidth][c]{$\pm$0.0015} & \makebox[\metricwidth][c]{$\pm$0.0001} & \makebox[\metricwidth][c]{$\pm$0.0006} & \makebox[\metricwidth][c]{$\pm$0.0007} & \makebox[\metricwidth][c]{$\pm$0.0003} & \makebox[\metricwidth][c]{$\pm$0.0001} & \makebox[\metricwidth][c]{$\pm$0.0001} \\

& 336 & \makebox[\metricwidth][c]{\phantom{$\pm$}\colorbox{bestlight}{\strut\underline{0.3413}}} & \makebox[\metricwidth][c]{\phantom{$\pm$}\colorbox{bestlight}{\strut\underline{0.3768}}} & \makebox[\metricwidth][c]{\phantom{$\pm$}0.3769} & \makebox[\metricwidth][c]{\phantom{$\pm$}0.4026} & \makebox[\metricwidth][c]{\phantom{$\pm$}0.3597} & \makebox[\metricwidth][c]{\phantom{$\pm$}0.3905} & \makebox[\metricwidth][c]{\phantom{$\pm$}\colorbox{secondlight}{\strut 0.3463}} & \makebox[\metricwidth][c]{\phantom{$\pm$}\colorbox{secondlight}{\strut 0.3776}} & \makebox[\metricwidth][c]{\phantom{$\pm$}0.3514} & \makebox[\metricwidth][c]{\phantom{$\pm$}0.3807} & \makebox[\metricwidth][c]{\phantom{$\pm$}0.3611} & \makebox[\metricwidth][c]{\phantom{$\pm$}0.3860} \\
& & \makebox[\metricwidth][c]{$\pm$0.0008} & \makebox[\metricwidth][c]{$\pm$0.0005} & \makebox[\metricwidth][c]{$\pm$0.0006} & \makebox[\metricwidth][c]{$\pm$0.0002} & \makebox[\metricwidth][c]{$\pm$0.0012} & \makebox[\metricwidth][c]{$\pm$0.0007} & \makebox[\metricwidth][c]{$\pm$0.0013} & \makebox[\metricwidth][c]{$\pm$0.0002} & \makebox[\metricwidth][c]{$\pm$0.0002} & \makebox[\metricwidth][c]{$\pm$0.0003} & \makebox[\metricwidth][c]{$\pm$0.0001} & \makebox[\metricwidth][c]{$\pm$0.0001} \\

& 720 & \makebox[\metricwidth][c]{\phantom{$\pm$}\colorbox{secondlight}{\strut 0.3891}} & \makebox[\metricwidth][c]{\phantom{$\pm$}\colorbox{bestlight}{\strut 0.4044}} & \makebox[\metricwidth][c]{\phantom{$\pm$}0.4378} & \makebox[\metricwidth][c]{\phantom{$\pm$}0.4398} & \makebox[\metricwidth][c]{\phantom{$\pm$}0.4208} & \makebox[\metricwidth][c]{\phantom{$\pm$}0.4231} & \makebox[\metricwidth][c]{\phantom{$\pm$}0.3902} & \makebox[\metricwidth][c]{\phantom{$\pm$}0.4076} & \makebox[\metricwidth][c]{\phantom{$\pm$}\colorbox{bestlight}{\strut\underline{0.3814}}} & \makebox[\metricwidth][c]{\phantom{$\pm$}\colorbox{secondlight}{\strut 0.4050}} & \makebox[\metricwidth][c]{\phantom{$\pm$}0.3981} & \makebox[\metricwidth][c]{\phantom{$\pm$}0.4071} \\
& & \makebox[\metricwidth][c]{$\pm$0.0003} & \makebox[\metricwidth][c]{$\pm$0.0004} & \makebox[\metricwidth][c]{$\pm$0.0024} & \makebox[\metricwidth][c]{$\pm$0.0013} & \makebox[\metricwidth][c]{$\pm$0.0014} & \makebox[\metricwidth][c]{$\pm$0.0009} & \makebox[\metricwidth][c]{$\pm$0.0001} & \makebox[\metricwidth][c]{$\pm$0.0003} & \makebox[\metricwidth][c]{$\pm$0.0003} & \makebox[\metricwidth][c]{$\pm$0.0008} & \makebox[\metricwidth][c]{$\pm$0.0000} & \makebox[\metricwidth][c]{$\pm$0.0000} \\

\midrule


\multirow{8}{*}{ETTm2} & 96 & \makebox[\metricwidth][c]{\phantom{$\pm$}\colorbox{bestlight}{\strut\underline{0.1528}}} & \makebox[\metricwidth][c]{\phantom{$\pm$}\colorbox{bestlight}{\strut\underline{0.2409}}} & \makebox[\metricwidth][c]{\phantom{$\pm$}0.1759} & \makebox[\metricwidth][c]{\phantom{$\pm$}0.2661} & \makebox[\metricwidth][c]{\phantom{$\pm$}0.1706} & \makebox[\metricwidth][c]{\phantom{$\pm$}0.2545} & \makebox[\metricwidth][c]{\phantom{$\pm$}0.1677} & \makebox[\metricwidth][c]{\phantom{$\pm$}0.2513} & \makebox[\metricwidth][c]{\phantom{$\pm$}\colorbox{secondlight}{\strut 0.1568}} & \makebox[\metricwidth][c]{\phantom{$\pm$}\colorbox{secondlight}{\strut 0.2423}} & \makebox[\metricwidth][c]{\phantom{$\pm$}0.1617} & \makebox[\metricwidth][c]{\phantom{$\pm$}0.2553} \\
& & \makebox[\metricwidth][c]{$\pm$0.0004} & \makebox[\metricwidth][c]{$\pm$0.0003} & \makebox[\metricwidth][c]{$\pm$0.0013} & \makebox[\metricwidth][c]{$\pm$0.0024} & \makebox[\metricwidth][c]{$\pm$0.0013} & \makebox[\metricwidth][c]{$\pm$0.0019} & \makebox[\metricwidth][c]{$\pm$0.0004} & \makebox[\metricwidth][c]{$\pm$0.0006} & \makebox[\metricwidth][c]{$\pm$0.0007} & \makebox[\metricwidth][c]{$\pm$0.0005} & \makebox[\metricwidth][c]{$\pm$0.0003} & \makebox[\metricwidth][c]{$\pm$0.0002} \\

& 192 & \makebox[\metricwidth][c]{\phantom{$\pm$}\colorbox{bestlight}{\strut\underline{0.2075}}} & \makebox[\metricwidth][c]{\phantom{$\pm$}\colorbox{bestlight}{\strut\underline{0.2805}}} & \makebox[\metricwidth][c]{\phantom{$\pm$}0.2360} & \makebox[\metricwidth][c]{\phantom{$\pm$}0.3066} & \makebox[\metricwidth][c]{\phantom{$\pm$}0.2306} & \makebox[\metricwidth][c]{\phantom{$\pm$}0.2968} & \makebox[\metricwidth][c]{\phantom{$\pm$}0.2264} & \makebox[\metricwidth][c]{\phantom{$\pm$}0.2891} & \makebox[\metricwidth][c]{\phantom{$\pm$}\colorbox{secondlight}{\strut 0.2118}} & \makebox[\metricwidth][c]{\phantom{$\pm$}\colorbox{secondlight}{\strut 0.2819}} & \makebox[\metricwidth][c]{\phantom{$\pm$}0.2170} & \makebox[\metricwidth][c]{\phantom{$\pm$}0.2905} \\
& & \makebox[\metricwidth][c]{$\pm$0.0002} & \makebox[\metricwidth][c]{$\pm$0.0004} & \makebox[\metricwidth][c]{$\pm$0.0041} & \makebox[\metricwidth][c]{$\pm$0.0029} & \makebox[\metricwidth][c]{$\pm$0.0021} & \makebox[\metricwidth][c]{$\pm$0.0022} & \makebox[\metricwidth][c]{$\pm$0.0000} & \makebox[\metricwidth][c]{$\pm$0.0001} & \makebox[\metricwidth][c]{$\pm$0.0018} & \makebox[\metricwidth][c]{$\pm$0.0020} & \makebox[\metricwidth][c]{$\pm$0.0001} & \makebox[\metricwidth][c]{$\pm$0.0001} \\

& 336 & \makebox[\metricwidth][c]{\phantom{$\pm$}\colorbox{bestlight}{\strut\underline{0.2554}}} & \makebox[\metricwidth][c]{\phantom{$\pm$}\colorbox{bestlight}{\strut\underline{0.3141}}} & \makebox[\metricwidth][c]{\phantom{$\pm$}0.2800} & \makebox[\metricwidth][c]{\phantom{$\pm$}0.3365} & \makebox[\metricwidth][c]{\phantom{$\pm$}0.2860} & \makebox[\metricwidth][c]{\phantom{$\pm$}0.3349} & \makebox[\metricwidth][c]{\phantom{$\pm$}0.2704} & \makebox[\metricwidth][c]{\phantom{$\pm$}0.3258} & \makebox[\metricwidth][c]{\phantom{$\pm$}0.2642} & \makebox[\metricwidth][c]{\phantom{$\pm$}\colorbox{secondlight}{\strut 0.3185}} & \makebox[\metricwidth][c]{\phantom{$\pm$}\colorbox{secondlight}{\strut 0.2593}} & \makebox[\metricwidth][c]{\phantom{$\pm$}0.3242} \\
& & \makebox[\metricwidth][c]{$\pm$0.0006} & \makebox[\metricwidth][c]{$\pm$0.0004} & \makebox[\metricwidth][c]{$\pm$0.0007} & \makebox[\metricwidth][c]{$\pm$0.0005} & \makebox[\metricwidth][c]{$\pm$0.0028} & \makebox[\metricwidth][c]{$\pm$0.0028} & \makebox[\metricwidth][c]{$\pm$0.0001} & \makebox[\metricwidth][c]{$\pm$0.0002} & \makebox[\metricwidth][c]{$\pm$0.0012} & \makebox[\metricwidth][c]{$\pm$0.0011} & \makebox[\metricwidth][c]{$\pm$0.0002} & \makebox[\metricwidth][c]{$\pm$0.0001} \\

& 720 & \makebox[\metricwidth][c]{\phantom{$\pm$}\colorbox{bestlight}{\strut\underline{0.3268}}} & \makebox[\metricwidth][c]{\phantom{$\pm$}\colorbox{bestlight}{\strut\underline{0.3623}}} & \makebox[\metricwidth][c]{\phantom{$\pm$}0.3698} & \makebox[\metricwidth][c]{\phantom{$\pm$}0.3915} & \makebox[\metricwidth][c]{\phantom{$\pm$}0.3820} & \makebox[\metricwidth][c]{\phantom{$\pm$}0.3991} & \makebox[\metricwidth][c]{\phantom{$\pm$}0.3401} & \makebox[\metricwidth][c]{\phantom{$\pm$}0.3771} & \makebox[\metricwidth][c]{\phantom{$\pm$}0.3422} & \makebox[\metricwidth][c]{\phantom{$\pm$}0.3711} & \makebox[\metricwidth][c]{\phantom{$\pm$}\colorbox{secondlight}{\strut 0.3283}} & \makebox[\metricwidth][c]{\phantom{$\pm$}\colorbox{secondlight}{\strut 0.3691}} \\
& & \makebox[\metricwidth][c]{$\pm$0.0004} & \makebox[\metricwidth][c]{$\pm$0.0002} & \makebox[\metricwidth][c]{$\pm$0.0023} & \makebox[\metricwidth][c]{$\pm$0.0010} & \makebox[\metricwidth][c]{$\pm$0.0075} & \makebox[\metricwidth][c]{$\pm$0.0089} & \makebox[\metricwidth][c]{$\pm$0.0004} & \makebox[\metricwidth][c]{$\pm$0.0007} & \makebox[\metricwidth][c]{$\pm$0.0050} & \makebox[\metricwidth][c]{$\pm$0.0023} & \makebox[\metricwidth][c]{$\pm$0.0004} & \makebox[\metricwidth][c]{$\pm$0.0001} \\

\midrule
\multicolumn{2}{c}{$1^{st}$ Count}
& \makebox[\metricwidth][c]{\phantom{$\pm$}\colorbox{bestlight}{\makebox[\highlightwidth][c]{\strut 25}}} & \makebox[\metricwidth][c]{\phantom{$\pm$}\colorbox{bestlight}{\makebox[\highlightwidth][c]{\strut 20}}} & \makebox[\metricwidth][c]{\phantom{$\pm$}\makebox[\highlightwidth][c]{\strut 0}} & \makebox[\metricwidth][c]{\phantom{$\pm$}\makebox[\highlightwidth][c]{\strut 0}} & \makebox[\metricwidth][c]{\phantom{$\pm$}\makebox[\highlightwidth][c]{\strut 1}} & \makebox[\metricwidth][c]{\phantom{$\pm$}\makebox[\highlightwidth][c]{\strut 0}} & \makebox[\metricwidth][c]{\phantom{$\pm$}\makebox[\highlightwidth][c]{\strut 0}} & \makebox[\metricwidth][c]{\phantom{$\pm$}\colorbox{secondlight}{\makebox[\highlightwidth][c]{\strut 4}}} & \makebox[\metricwidth][c]{\phantom{$\pm$}\makebox[\highlightwidth][c]{\strut 1}} & \makebox[\metricwidth][c]{\phantom{$\pm$}\makebox[\highlightwidth][c]{\strut 3}} & \makebox[\metricwidth][c]{\phantom{$\pm$}\makebox[\highlightwidth][c]{\strut 1}} & \makebox[\metricwidth][c]{\phantom{$\pm$}\makebox[\highlightwidth][c]{\strut 1}} \\

\multicolumn{2}{c}{w/ $p<0.05$}
& \makebox[\metricwidth][c]{\phantom{$\pm$}\colorbox{bestlight}{\makebox[\highlightwidth][c]{\strut 23}}} & \makebox[\metricwidth][c]{\phantom{$\pm$}\colorbox{bestlight}{\makebox[\highlightwidth][c]{\strut 18}}} & \makebox[\metricwidth][c]{\phantom{$\pm$}\makebox[\highlightwidth][c]{\strut 0}} & \makebox[\metricwidth][c]{\phantom{$\pm$}\makebox[\highlightwidth][c]{\strut 0}} & \makebox[\metricwidth][c]{\phantom{$\pm$}\makebox[\highlightwidth][c]{\strut 1}} & \makebox[\metricwidth][c]{\phantom{$\pm$}\makebox[\highlightwidth][c]{\strut 0}} & \makebox[\metricwidth][c]{\phantom{$\pm$}\makebox[\highlightwidth][c]{\strut 0}} & \makebox[\metricwidth][c]{\phantom{$\pm$}\colorbox{secondlight}{\makebox[\highlightwidth][c]{\strut 3}}} & \makebox[\metricwidth][c]{\phantom{$\pm$}\makebox[\highlightwidth][c]{\strut 1}} & \makebox[\metricwidth][c]{\phantom{$\pm$}\makebox[\highlightwidth][c]{\strut 2}} & \makebox[\metricwidth][c]{\phantom{$\pm$}\makebox[\highlightwidth][c]{\strut 1}} & \makebox[\metricwidth][c]{\phantom{$\pm$}\makebox[\highlightwidth][c]{\strut 1}} \\
\bottomrule
\end{tabular}
}
\end{table*}


\newlength{\strongstdmetricwidth}
\setlength{\strongstdmetricwidth}{1.5cm}

\newlength{\strongstdhalfmodelgap}
\setlength{\strongstdhalfmodelgap}{7pt}

\definecolor{strongstdhighlightcolor}{HTML}{9ECED0}
\colorlet{strongstdbestlight}{strongstdhighlightcolor!80}
\colorlet{strongstdsecondlight}{strongstdhighlightcolor!35}

\newcommand{\strongstdmean}[1]{%
  \makebox[\strongstdmetricwidth][c]{%
    \phantom{$\pm$}#1%
  }%
}

\newcommand{\strongstdbestst}[1]{%
  \makebox[\strongstdmetricwidth][c]{%
    \phantom{$\pm$}%
    \colorbox{strongstdbestlight}{%
      \strut\underline{#1}%
    }%
  }%
}

\newcommand{\strongstdbest}[1]{%
  \makebox[\strongstdmetricwidth][c]{%
    \phantom{$\pm$}%
    \colorbox{strongstdbestlight}{%
      \strut #1%
    }%
  }%
}

\newcommand{\strongstdsecond}[1]{%
  \makebox[\strongstdmetricwidth][c]{%
    \phantom{$\pm$}%
    \colorbox{strongstdsecondlight}{%
      \strut #1%
    }%
  }%
}

\newcommand{\strongstddev}[1]{%
  \makebox[\strongstdmetricwidth][c]{%
    $\pm$#1%
  }%
}

\newlength{\strongstdhighlightwidth}
\settowidth{\strongstdhighlightwidth}{0.0000}


\begin{table*}[h]
\centering
\caption{Detailed MSE and MAE results (mean $\pm$ std) on strongly dependent datasets.
Highlighting and underlining follow Table~\ref{tab:strong_results}.}
\label{tab:strong_results_std}

\resizebox{1.0\textwidth}{!}{%
\begin{tabular}{
lr|
>{\centering\arraybackslash}p{\strongstdmetricwidth}
>{\centering\arraybackslash}p{\strongstdmetricwidth}
@{\hspace{\strongstdhalfmodelgap}}
>{\centering\arraybackslash}p{\strongstdmetricwidth}
>{\centering\arraybackslash}p{\strongstdmetricwidth}
@{\hspace{\strongstdhalfmodelgap}}
>{\centering\arraybackslash}p{\strongstdmetricwidth}
>{\centering\arraybackslash}p{\strongstdmetricwidth}
@{\hspace{\strongstdhalfmodelgap}}
>{\centering\arraybackslash}p{\strongstdmetricwidth}
>{\centering\arraybackslash}p{\strongstdmetricwidth}
@{\hspace{\strongstdhalfmodelgap}}
>{\centering\arraybackslash}p{\strongstdmetricwidth}
>{\centering\arraybackslash}p{\strongstdmetricwidth}
@{\hspace{\strongstdhalfmodelgap}}
>{\centering\arraybackslash}p{\strongstdmetricwidth}
>{\centering\arraybackslash}p{\strongstdmetricwidth}
}

\toprule

\multicolumn{2}{c}{Model}
& \multicolumn{2}{c}{\textbf{Chameleon}}
& \multicolumn{2}{c}{Chameleon (CI)}
& \multicolumn{2}{c}{S-Mamba+ACN}
& \multicolumn{2}{c}{TimeFilter}
& \multicolumn{2}{c}{DUET}
& \multicolumn{2}{c}{Crossformer} \\

\cmidrule(lr){1-2}
\cmidrule(lr){3-4}
\cmidrule(lr){5-6}
\cmidrule(lr){7-8}
\cmidrule(lr){9-10}
\cmidrule(lr){11-12}
\cmidrule(lr){13-14}

Dataset & (F)
& MSE & MAE
& MSE & MAE
& MSE & MAE
& MSE & MAE
& MSE & MAE
& MSE & MAE \\

\midrule


\multirow{4}{*}{\shortstack[l]{Double\\Pendulum}}
& 96
& \strongstdbestst{0.0678}
& \strongstdbestst{0.1176}
& \strongstdmean{0.1258}
& \strongstdmean{0.1880}
& \strongstdmean{0.0927}
& \strongstdmean{0.1642}
& \strongstdsecond{0.0870}
& \strongstdsecond{0.1616}
& \strongstdmean{0.2510}
& \strongstdmean{0.3271}
& \strongstdmean{0.1208}
& \strongstdmean{0.1975} \\

&
& \strongstddev{0.0029}
& \strongstddev{0.0026}
& \strongstddev{0.0043}
& \strongstddev{0.0040}
& \strongstddev{0.0019}
& \strongstddev{0.0013}
& \strongstddev{0.0097}
& \strongstddev{0.0101}
& \strongstddev{0.0053}
& \strongstddev{0.0046}
& \strongstddev{0.0074}
& \strongstddev{0.0039} \\

& 192
& \strongstdbestst{0.3053}
& \strongstdbestst{0.3216}
& \strongstdmean{0.3845}
& \strongstdmean{0.3836}
& \strongstdsecond{0.3325}
& \strongstdsecond{0.3701}
& \strongstdmean{0.3522}
& \strongstdmean{0.3787}
& \strongstdmean{0.5044}
& \strongstdmean{0.5052}
& \strongstdmean{0.4774}
& \strongstdmean{0.4717} \\

&
& \strongstddev{0.0051}
& \strongstddev{0.0033}
& \strongstddev{0.0035}
& \strongstddev{0.0027}
& \strongstddev{0.0032}
& \strongstddev{0.0024}
& \strongstddev{0.0144}
& \strongstddev{0.0123}
& \strongstddev{0.0050}
& \strongstddev{0.0036}
& \strongstddev{0.0209}
& \strongstddev{0.0217} \\

\midrule


\multirow{4}{*}{\shortstack[l]{Lorenz\\Coupled}}
& 48
& \strongstdbestst{0.0179}
& \strongstdbestst{0.0389}
& \strongstdmean{0.0493}
& \strongstdsecond{0.0761}
& \strongstdmean{0.0708}
& \strongstdmean{0.1247}
& \strongstdsecond{0.0424}
& \strongstdmean{0.0934}
& \strongstdmean{0.1570}
& \strongstdmean{0.2137}
& \strongstdmean{0.0581}
& \strongstdmean{0.1152} \\

&
& \strongstddev{0.0016}
& \strongstddev{0.0014}
& \strongstddev{0.0015}
& \strongstddev{0.0017}
& \strongstddev{0.0004}
& \strongstddev{0.0005}
& \strongstddev{0.0025}
& \strongstddev{0.0059}
& \strongstddev{0.0036}
& \strongstddev{0.0037}
& \strongstddev{0.0027}
& \strongstddev{0.0014} \\

& 96
& \strongstdbestst{0.1569}
& \strongstdbestst{0.1842}
& \strongstdmean{0.3368}
& \strongstdsecond{0.3213}
& \strongstdmean{0.3666}
& \strongstdmean{0.3837}
& \strongstdsecond{0.2935}
& \strongstdmean{0.3230}
& \strongstdmean{0.5026}
& \strongstdmean{0.4758}
& \strongstdmean{0.3019}
& \strongstdmean{0.3299} \\

&
& \strongstddev{0.0075}
& \strongstddev{0.0040}
& \strongstddev{0.0052}
& \strongstddev{0.0042}
& \strongstddev{0.0013}
& \strongstddev{0.0014}
& \strongstddev{0.0098}
& \strongstddev{0.0075}
& \strongstddev{0.0043}
& \strongstddev{0.0025}
& \strongstddev{0.0085}
& \strongstddev{0.0045} \\

\midrule


\multirow{4}{*}{PEMS03}
& 48
& \strongstdbestst{0.0905}
& \strongstdbestst{0.1920}
& \strongstdmean{0.1045}
& \strongstdmean{0.1991}
& \strongstdsecond{0.0927}
& \strongstdsecond{0.1968}
& \strongstdmean{0.1048}
& \strongstdmean{0.2099}
& \strongstdmean{0.1028}
& \strongstdmean{0.2050}
& \strongstdmean{0.0968}
& \strongstdmean{0.1989} \\

&
& \strongstddev{0.0007}
& \strongstddev{0.0007}
& \strongstddev{0.0007}
& \strongstddev{0.0003}
& \strongstddev{0.0012}
& \strongstddev{0.0013}
& \strongstddev{0.0004}
& \strongstddev{0.0004}
& \strongstddev{0.0054}
& \strongstddev{0.0062}
& \strongstddev{0.0028}
& \strongstddev{0.0020} \\

& 96
& \strongstdbest{0.1156}
& \strongstdbest{0.2157}
& \strongstdmean{0.1359}
& \strongstdmean{0.2229}
& \strongstdmean{0.2309}
& \strongstdmean{0.3330}
& \strongstdmean{0.1407}
& \strongstdmean{0.2389}
& \strongstdmean{0.1447}
& \strongstdmean{0.2407}
& \strongstdsecond{0.1211}
& \strongstdsecond{0.2208} \\

&
& \strongstddev{0.0015}
& \strongstddev{0.0014}
& \strongstddev{0.0005}
& \strongstddev{0.0005}
& \strongstddev{0.0385}
& \strongstddev{0.0337}
& \strongstddev{0.0011}
& \strongstddev{0.0008}
& \strongstddev{0.0045}
& \strongstddev{0.0044}
& \strongstddev{0.0074}
& \strongstddev{0.0050} \\

\midrule


\multirow{4}{*}{PEMS08}
& 48
& \strongstdbestst{0.0975}
& \strongstdbestst{0.1783}
& \strongstdsecond{0.1187}
& \strongstdsecond{0.1924}
& \strongstdmean{0.1266}
& \strongstdmean{0.1985}
& \strongstdmean{0.1617}
& \strongstdmean{0.2047}
& \strongstdmean{0.1244}
& \strongstdmean{0.1929}
& \strongstdmean{0.1525}
& \strongstdmean{0.2111} \\

&
& \strongstddev{0.0019}
& \strongstddev{0.0008}
& \strongstddev{0.0039}
& \strongstddev{0.0008}
& \strongstddev{0.0067}
& \strongstddev{0.0097}
& \strongstddev{0.0094}
& \strongstddev{0.0063}
& \strongstddev{0.0076}
& \strongstddev{0.0017}
& \strongstddev{0.0049}
& \strongstddev{0.0043} \\

& 96
& \strongstdbestst{0.1307}
& \strongstdbestst{0.1910}
& \strongstdmean{0.1703}
& \strongstdmean{0.2115}
& \strongstdmean{0.1869}
& \strongstdsecond{0.2064}
& \strongstdmean{0.2960}
& \strongstdmean{0.3193}
& \strongstdsecond{0.1663}
& \strongstdmean{0.2087}
& \strongstdmean{0.2201}
& \strongstdmean{0.2335} \\

&
& \strongstddev{0.0024}
& \strongstddev{0.0003}
& \strongstddev{0.0034}
& \strongstddev{0.0006}
& \strongstddev{0.0041}
& \strongstddev{0.0026}
& \strongstddev{0.0258}
& \strongstddev{0.0167}
& \strongstddev{0.0068}
& \strongstddev{0.0011}
& \strongstddev{0.0104}
& \strongstddev{0.0078} \\

\midrule

\multicolumn{2}{c}{$1^{st}$ Count}
& \makebox[\strongstdmetricwidth][c]{%
    \phantom{$\pm$}%
    \colorbox{strongstdbestlight}{%
      \makebox[\strongstdhighlightwidth][c]{\strut 8}%
    }%
  }
& \makebox[\strongstdmetricwidth][c]{%
    \phantom{$\pm$}%
    \colorbox{strongstdbestlight}{%
      \makebox[\strongstdhighlightwidth][c]{\strut 8}%
    }%
  }
& \makebox[\strongstdmetricwidth][c]{%
    \phantom{$\pm$}\makebox[\strongstdhighlightwidth][c]{\strut 0}%
  }
& \makebox[\strongstdmetricwidth][c]{%
    \phantom{$\pm$}\makebox[\strongstdhighlightwidth][c]{\strut 0}%
  }
& \makebox[\strongstdmetricwidth][c]{%
    \phantom{$\pm$}\makebox[\strongstdhighlightwidth][c]{\strut 0}%
  }
& \makebox[\strongstdmetricwidth][c]{%
    \phantom{$\pm$}\makebox[\strongstdhighlightwidth][c]{\strut 0}%
  }
& \makebox[\strongstdmetricwidth][c]{%
    \phantom{$\pm$}\makebox[\strongstdhighlightwidth][c]{\strut 0}%
  }
& \makebox[\strongstdmetricwidth][c]{%
    \phantom{$\pm$}\makebox[\strongstdhighlightwidth][c]{\strut 0}%
  }
& \makebox[\strongstdmetricwidth][c]{%
    \phantom{$\pm$}\makebox[\strongstdhighlightwidth][c]{\strut 0}%
  }
& \makebox[\strongstdmetricwidth][c]{%
    \phantom{$\pm$}\makebox[\strongstdhighlightwidth][c]{\strut 0}%
  }
& \makebox[\strongstdmetricwidth][c]{%
    \phantom{$\pm$}\makebox[\strongstdhighlightwidth][c]{\strut 0}%
  }
& \makebox[\strongstdmetricwidth][c]{%
    \phantom{$\pm$}\makebox[\strongstdhighlightwidth][c]{\strut 0}%
  } \\

\multicolumn{2}{c}{w/ $p<0.05$}
& \makebox[\strongstdmetricwidth][c]{%
    \phantom{$\pm$}%
    \colorbox{strongstdbestlight}{%
      \makebox[\strongstdhighlightwidth][c]{\strut 7}%
    }%
  }
& \makebox[\strongstdmetricwidth][c]{%
    \phantom{$\pm$}%
    \colorbox{strongstdbestlight}{%
      \makebox[\strongstdhighlightwidth][c]{\strut 7}%
    }%
  }
& \makebox[\strongstdmetricwidth][c]{%
    \phantom{$\pm$}\makebox[\strongstdhighlightwidth][c]{\strut 0}%
  }
& \makebox[\strongstdmetricwidth][c]{%
    \phantom{$\pm$}\makebox[\strongstdhighlightwidth][c]{\strut 0}%
  }
& \makebox[\strongstdmetricwidth][c]{%
    \phantom{$\pm$}\makebox[\strongstdhighlightwidth][c]{\strut 0}%
  }
& \makebox[\strongstdmetricwidth][c]{%
    \phantom{$\pm$}\makebox[\strongstdhighlightwidth][c]{\strut 0}%
  }
& \makebox[\strongstdmetricwidth][c]{%
    \phantom{$\pm$}\makebox[\strongstdhighlightwidth][c]{\strut 0}%
  }
& \makebox[\strongstdmetricwidth][c]{%
    \phantom{$\pm$}\makebox[\strongstdhighlightwidth][c]{\strut 0}%
  }
& \makebox[\strongstdmetricwidth][c]{%
    \phantom{$\pm$}\makebox[\strongstdhighlightwidth][c]{\strut 0}%
  }
& \makebox[\strongstdmetricwidth][c]{%
    \phantom{$\pm$}\makebox[\strongstdhighlightwidth][c]{\strut 0}%
  }
& \makebox[\strongstdmetricwidth][c]{%
    \phantom{$\pm$}\makebox[\strongstdhighlightwidth][c]{\strut 0}%
  }
& \makebox[\strongstdmetricwidth][c]{%
    \phantom{$\pm$}\makebox[\strongstdhighlightwidth][c]{\strut 0}%
  } \\

\bottomrule
\end{tabular}%
}
\end{table*}
\begingroup
\definecolor{castdhighlight}{HTML}{9ECED0}
\colorlet{castdbestlight}{castdhighlight!80}
\colorlet{castdsecondlight}{castdhighlight!35}
\newlength{\castdwidth}
\setlength{\castdwidth}{1.55cm}
\newcommand{\castdmean}[1]{\makebox[\castdwidth][c]{\phantom{$\pm$}#1}}
\newcommand{\castdbest}[1]{\makebox[\castdwidth][c]{\phantom{$\pm$}\colorbox{castdbestlight}{\strut #1}}}
\newcommand{\castdbestst}[1]{\makebox[\castdwidth][c]{\phantom{$\pm$}\colorbox{castdbestlight}{\strut\underline{#1}}}}
\newcommand{\castdsecond}[1]{\makebox[\castdwidth][c]{\phantom{$\pm$}\colorbox{castdsecondlight}{\strut #1}}}
\newcommand{\castddev}[1]{\makebox[\castdwidth][c]{$\pm$#1}}
\newcommand{\castdhead}[2][\castdwidth]{\parbox[c][3.4em][c]{#1}{\centering #2}}
\begin{table*}[t]
\centering
\caption{Detailed MSE results (mean $\pm$ std) for ablations of the measurement update and comparisons with alternative mechanisms. Highlighting and underlining follow Table~\ref{tab:strong_results}.}
\label{tab:correction_ablation_std}
\setlength{\tabcolsep}{3pt}
\resizebox{\textwidth}{!}{%
\begin{tabular}{lr|*{7}{>{\centering\arraybackslash}p{\castdwidth}}}
\toprule
\multicolumn{1}{c}{\castdhead[1.35cm]{Dataset}} & \multicolumn{1}{c|}{\castdhead[0.45cm]{$(F)$}}
 & \multicolumn{1}{c}{\castdhead{\textbf{Chameleon}}} & \multicolumn{1}{c}{\castdhead{Chameleon\\($\gamma=0$)}} & \multicolumn{1}{c}{\castdhead{Chameleon\\(CI)}} & \multicolumn{1}{c}{\castdhead{Attention}} & \multicolumn{1}{c}{\castdhead{SSM}} & \multicolumn{1}{c}{\castdhead{Static\\low-rank}} & \multicolumn{1}{c}{\castdhead{Dynamic\\low-rank}} \\
\midrule
\multirow{4}{*}{LorenzC} & 48 & \castdbest{0.0179} & \castdmean{0.0460} & \castdmean{0.0493} & \castdmean{0.0212} & \castdsecond{0.0197} & \castdmean{0.0202} & \castdmean{0.0204} \\
 &  & \castddev{0.0016} & \castddev{0.0008} & \castddev{0.0015} & \castddev{0.0011} & \castddev{0.0016} & \castddev{0.0013} & \castddev{0.0016} \\
 & 96 & \castdbestst{0.1569} & \castdmean{0.3273} & \castdmean{0.3368} & \castdmean{0.1883} & \castdsecond{0.1785} & \castdmean{0.1912} & \castdmean{0.1826} \\
 &  & \castddev{0.0075} & \castddev{0.0078} & \castddev{0.0052} & \castddev{0.0136} & \castddev{0.0109} & \castddev{0.0160} & \castddev{0.0125} \\
\midrule
\multirow{4}{*}{PEMS08} & 48 & \castdbest{0.0975} & \castdmean{0.1185} & \castdmean{0.1187} & \castdmean{0.1163} & \castdmean{0.1023} & \castdmean{0.1233} & \castdsecond{0.0989} \\
 &  & \castddev{0.0019} & \castddev{0.0011} & \castddev{0.0039} & \castddev{0.0047} & \castddev{0.0023} & \castddev{0.0091} & \castddev{0.0008} \\
 & 96 & \castdbestst{0.1307} & \castdmean{0.1675} & \castdmean{0.1703} & \castdmean{0.1726} & \castdmean{0.1460} & \castdmean{0.1827} & \castdsecond{0.1380} \\
 &  & \castddev{0.0024} & \castddev{0.0049} & \castddev{0.0034} & \castddev{0.0083} & \castddev{0.0029} & \castddev{0.0114} & \castddev{0.0056} \\
\midrule
\multirow{4}{*}{Solar} & 96 & \castdbestst{0.1534} & \castdmean{0.1672} & \castdmean{0.1681} & \castdsecond{0.1609} & \castdmean{0.1627} & \castdmean{0.1624} & \castdmean{0.1619} \\
 &  & \castddev{0.0024} & \castddev{0.0019} & \castddev{0.0014} & \castddev{0.0023} & \castddev{0.0040} & \castddev{0.0037} & \castddev{0.0041} \\
 & 336 & \castdbest{0.1899} & \castdmean{0.1925} & \castdmean{0.1941} & \castdmean{0.1946} & \castdmean{0.2078} & \castdsecond{0.1919} & \castdmean{0.1927} \\
 &  & \castddev{0.0024} & \castddev{0.0009} & \castddev{0.0025} & \castddev{0.0019} & \castddev{0.0149} & \castddev{0.0028} & \castddev{0.0026} \\
\midrule
\multirow{8}{*}{ETTh2} & 96 & \castdbest{0.2628} & \castdmean{0.2630} & \castdmean{0.2630} & \castdsecond{0.2629} & \castdmean{0.2649} & \castdmean{0.2631} & \castdmean{0.2629} \\
 &  & \castddev{0.0004} & \castddev{0.0003} & \castddev{0.0006} & \castddev{0.0003} & \castddev{0.0016} & \castddev{0.0008} & \castddev{0.0001} \\
 & 192 & \castdbest{0.3250} & \castdmean{0.3262} & \castdsecond{0.3251} & \castdmean{0.3251} & \castdmean{0.3292} & \castdmean{0.3258} & \castdmean{0.3251} \\
 &  & \castddev{0.0007} & \castddev{0.0024} & \castddev{0.0013} & \castddev{0.0010} & \castddev{0.0015} & \castddev{0.0009} & \castddev{0.0008} \\
 & 336 & \castdbest{0.3523} & \castdmean{0.3523} & \castdmean{0.3530} & \castdsecond{0.3523} & \castdmean{0.3584} & \castdmean{0.3535} & \castdmean{0.3528} \\
 &  & \castddev{0.0005} & \castddev{0.0012} & \castddev{0.0010} & \castddev{0.0015} & \castddev{0.0038} & \castddev{0.0005} & \castddev{0.0009} \\
 & 720 & \castdbest{0.3837} & \castdmean{0.3896} & \castdmean{0.3901} & \castdsecond{0.3865} & \castdmean{0.3890} & \castdmean{0.3918} & \castdmean{0.3882} \\
 &  & \castddev{0.0023} & \castddev{0.0059} & \castddev{0.0057} & \castddev{0.0051} & \castddev{0.0058} & \castddev{0.0058} & \castddev{0.0077} \\
\bottomrule
\end{tabular}%
}
\end{table*}
\endgroup

For the alternative mechanisms in Table~\ref{tab:correction_ablation}, attention computes queries and keys from RMS-normalized, flattened prior states and uses the original prior states as values. The bidirectional selective SSM scans along the variable dimension with input-dependent transition gates, state injection vectors, and state readout vectors, then averages the outputs from both directions. The static low-rank transformation takes the residual form $\tilde{\mathbf{S}}_l=\tilde{\mathbf{S}}_l^-+\mathbf{E}\mathbf{G}^\top\tilde{\mathbf{S}}_l^-$, where $\mathbf{E},\mathbf{G}\in\mathbb{R}^{N\times r}$ are learned parameters shared across samples and patch steps, with $\mathbf{E}$ initialized to zero so that the transformation initially preserves the prior states. All three alternatives replace the entire correction in Equation~\ref{eq:correction} and preserve the same learned $\omega$.

The dynamic low-rank alternative replaces only the Kalman gain, additionally preserving the observation construction and the correction form in Equation~\ref{eq:correction}. Its diagonal logits and factor rows are generated independently for each variable from the input patch features and the same prior-state readout used by Chameleon, with projection parameters shared across variables:
\begin{equation}
\mathbf{K}'_l=\operatorname{diag}\!\left(\text{sigmoid}(\mathbf{a}_l)\right)+\frac{1}{N}\mathbf{E}_l\mathbf{G}_l^\top,
\end{equation}
where $\mathbf{a}_l\in\mathbb{R}^N$ contains the logits for the diagonal term, and $\mathbf{E}_l,\mathbf{G}_l\in\mathbb{R}^{N\times r}$ are the low-rank factors. The logits follow the same bounded parameterization as Chameleon's log variance ratio. Each row $\mathbf{e}$ of either factor is smoothly normalized as $\mathbf{e}/\sqrt{1+\|\mathbf{e}\|_2^2}$. Together with the $1/N$ scaling, this ensures the low-rank term has a spectral norm of at most one, independent of $N$.

Tables~\ref{tab:correction_dimensions} and~\ref{tab:correction_budget} summarize the alternative-specific dimensions and per-layer computational costs, respectively. The costs include observation construction where applicable and the $\omega$ combination. Peak training memory is measured on an L40S GPU using FP32 tensors, with the same batch size across methods within each setting.

\begin{table}[t]
\centering
\small
\caption{Alternative-specific dimensions. We report the shared query and key dimension for attention, the projection/state dimensions for SSM, and factor widths for the low-rank (LR) alternatives.}
\label{tab:correction_dimensions}
\begin{tabular}{llcccc}
\toprule
Dataset & $F$ & Attention & SSM & Static LR & Dynamic LR \\
\midrule
LorenzC & 48/96 & 36 & 48/6 & 36 & 16 \\
PEMS08 & 48/96 & 16 & 32/16 & 64 & 32 \\
Solar & 96 & 12 & 32/12 & 48 & 32 \\
 & 336 & 12 & 24/12 & 48 & 32 \\
ETTh2 & 96/192/336 & 32 & 48/7 & 32 & 8 \\
 & 720 & 96 & 144/7 & 96 & 24 \\
\bottomrule
\end{tabular}
\end{table}
\begin{table*}[t]
\centering
\small
\setlength{\tabcolsep}{3pt}

\newcommand{\cbcost}[3]{\textbf{#1} / #2 / #3}
\newlength{\cbHeadOffset}
\newlength{\cbMiddleWidth}
\newcommand{\cbhead}[3]{%
  \settowidth{\cbHeadOffset}{\textbf{#2} / }%
  \settowidth{\cbMiddleWidth}{#3}%
  \addtolength{\cbHeadOffset}{0.5\cbMiddleWidth}%
  \makebox[0pt][l]{\hspace*{\cbHeadOffset}\makebox[0pt][c]{#1}}%
}
\caption{Computational costs of each cross-variable state correction mechanism per layer. Each entry reports \textbf{forward GFLOPs} / peak training memory (GiB) / parameters (K). Alternative-specific dimensions are selected for comparable costs, prioritizing forward FLOPs while avoiding excessively small dimensions. LR denotes low-rank.}
\label{tab:correction_budget}
\resizebox{\textwidth}{!}{%
\begin{tabular}{lllllll}
\toprule
Dataset & $F$ & \cbhead{Chameleon}{1.06}{0.23} & \cbhead{Attention}{1.01}{0.12} & \cbhead{SSM}{1.01}{0.27} & \cbhead{Static LR}{0.93}{0.32} & \cbhead{Dynamic LR}{1.22}{0.24} \\
\midrule
LorenzC & 48/96 & \cbcost{0.74}{0.24}{136} & \cbcost{0.77}{0.16}{147} & \cbcost{0.77}{0.41}{99.0} & \cbcost{0.70}{0.33}{0.43} & \cbcost{0.75}{0.27}{86.5} \\
PEMS08 & 48/96 & \cbcost{5.12}{0.77}{78.6} & \cbcost{9.20}{0.70}{65.5} & \cbcost{5.15}{2.14}{66.7} & \cbcost{5.79}{0.73}{21.8} & \cbcost{5.33}{0.79}{103} \\
Solar & 96 & \cbcost{14.5}{2.40}{28.1} & \cbcost{23.4}{2.17}{27.6} & \cbcost{13.5}{5.66}{37.7} & \cbcost{13.9}{2.25}{13.2} & \cbcost{14.9}{2.47}{41.8} \\
 & 336 & \cbcost{24.6}{4.61}{10.0} & \cbcost{46.9}{4.29}{27.6} & \cbcost{24.8}{11.3}{28.3} & \cbcost{27.7}{4.45}{13.2} & \cbcost{23.8}{4.75}{16.9} \\
ETTh2 & 96/192/336 & \cbcost{0.21}{0.11}{3.32} & \cbcost{0.19}{0.06}{3.07} & \cbcost{0.25}{0.14}{3.09} & \cbcost{0.16}{0.09}{0.45} & \cbcost{0.26}{0.11}{4.03} \\
 & 720 & \cbcost{1.06}{0.23}{18.9} & \cbcost{1.01}{0.12}{18.4} & \cbcost{1.01}{0.27}{16.1} & \cbcost{0.93}{0.32}{1.35} & \cbcost{1.22}{0.24}{18.8} \\
\bottomrule
\end{tabular}
}
\end{table*}

We also compare Chameleon against MGCN~\citep{lin2025}, which uses sensor spatial relationships through graph convolution and models temporal dependencies with bidirectional Mamba. We evaluate its released architecture and training configurations under the same protocol and $T$ grids as our PEMS experiments in Table~\ref{tab:strong_results}. As shown in Table~\ref{tab:mgcn_comparison}, Chameleon achieves average MSE and MAE reductions of 33.81\% and 14.91\%, respectively, with statistically significant improvements across all settings.

\begin{table}[t]
\centering
\small
\setlength{\tabcolsep}{6pt}
\newlength{\mgcnmetricwidth}
\setlength{\mgcnmetricwidth}{1cm}
\caption{Comparison with MGCN on PEMS03 and PEMS08. Underlining follows Table~\ref{tab:strong_results}.}
\label{tab:mgcn_comparison}
\begin{tabular}{lr|*{4}{>{\centering\arraybackslash}p{\mgcnmetricwidth}}}
\toprule
\multicolumn{2}{c}{Model}
& \multicolumn{2}{c}{\textbf{Chameleon}}
& \multicolumn{2}{c}{MGCN} \\
\cmidrule(lr){1-2}
\cmidrule(lr){3-4}
\cmidrule(lr){5-6}
Dataset & $(F)$ & MSE & MAE & MSE & MAE \\
\midrule
\multirow{2}{*}{PEMS03} & 48 & \underline{0.0905} & \underline{0.1920} & 0.1147 & 0.2132 \\
 & 96 & \underline{0.1156} & \underline{0.2157} & 0.1439 & 0.2370 \\
\midrule
\multirow{2}{*}{PEMS08} & 48 & \underline{0.0975} & \underline{0.1783} & 0.2103 & 0.2315 \\
 & 96 & \underline{0.1307} & \underline{0.1910} & 0.2208 & 0.2322 \\
\bottomrule
\end{tabular}
\end{table}

\begin{figure}[t]
    \centering
    \includegraphics[width=1.0\textwidth]{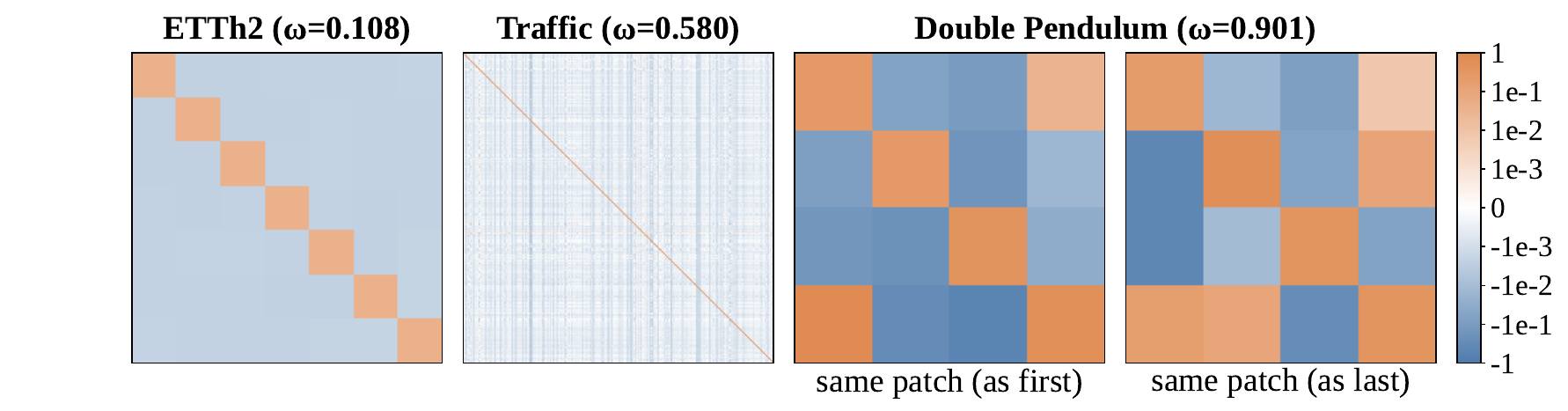}
    \caption{Learned effective Kalman gains $\omega\mathbf{K}_l$ on ETTh2, Traffic, and Double Pendulum. Larger off-diagonal magnitudes indicate stronger interactions across variables, while $\omega \in (0,1)$ is a learned gate controlling the contribution of the measurement update. The same Double Pendulum patch yields different gains when serving as the first and last patch in two temporally offset input windows.}
    \label{fig:kalman_gain}
\end{figure}

Figure~\ref{fig:kalman_gain} gives additional insights into Chameleon's measurement update through the learned effective Kalman gains $\omega\mathbf{K}_l$. For ETTh2 and Traffic, we visualize the gains with the largest mean off-diagonal magnitudes during testing, whereas we randomly sample the Double Pendulum example. Even under this selection, Double Pendulum shows substantially stronger cross-variable interactions and greater reliance on the measurement update, with $\omega=0.901$, compared with $0.580$ for Traffic and $0.108$ for ETTh2. In addition, the same Double Pendulum patch yields distinct gains in temporally offset input windows with different prior states, confirming that conditioning $\mathbf{P}_l^-$ on the prior state adapts the resulting correction.

\begin{figure*}[t]
    \centering
    \begin{subfigure}[t]{0.49\textwidth}
        \centering
        \includegraphics[width=\linewidth]{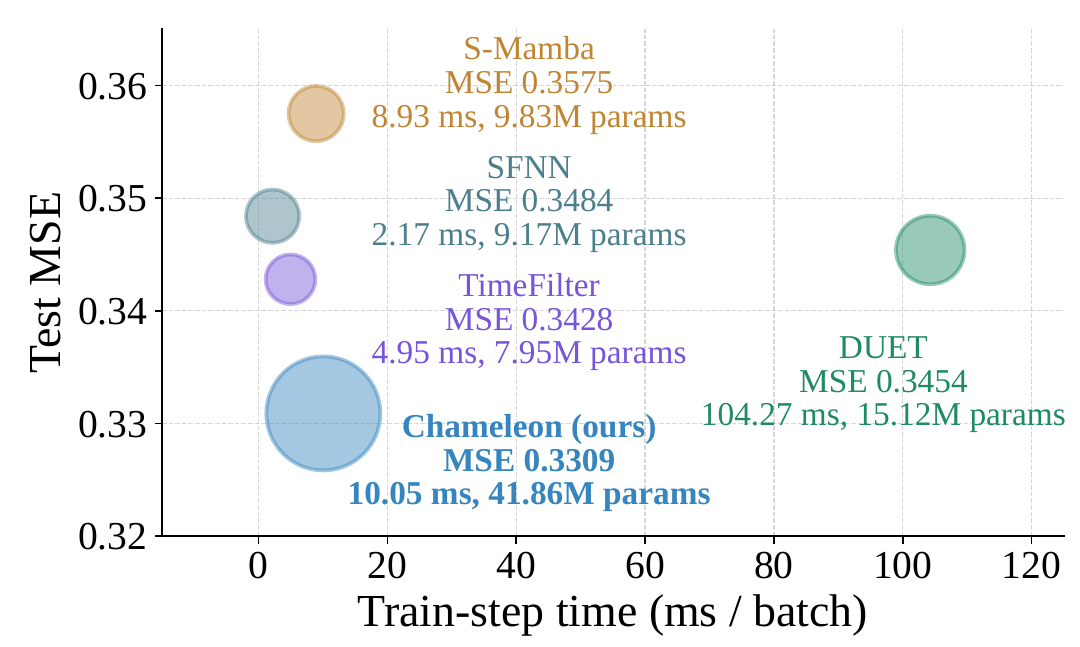}
        \captionsetup{width=0.97\linewidth}
        \caption{Test MSE versus training time on Traffic.}
        \label{fig:traffic_efficiency_params}
    \end{subfigure}\hfill
    \begin{subfigure}[t]{0.49\textwidth}
        \centering
        \includegraphics[width=\linewidth]{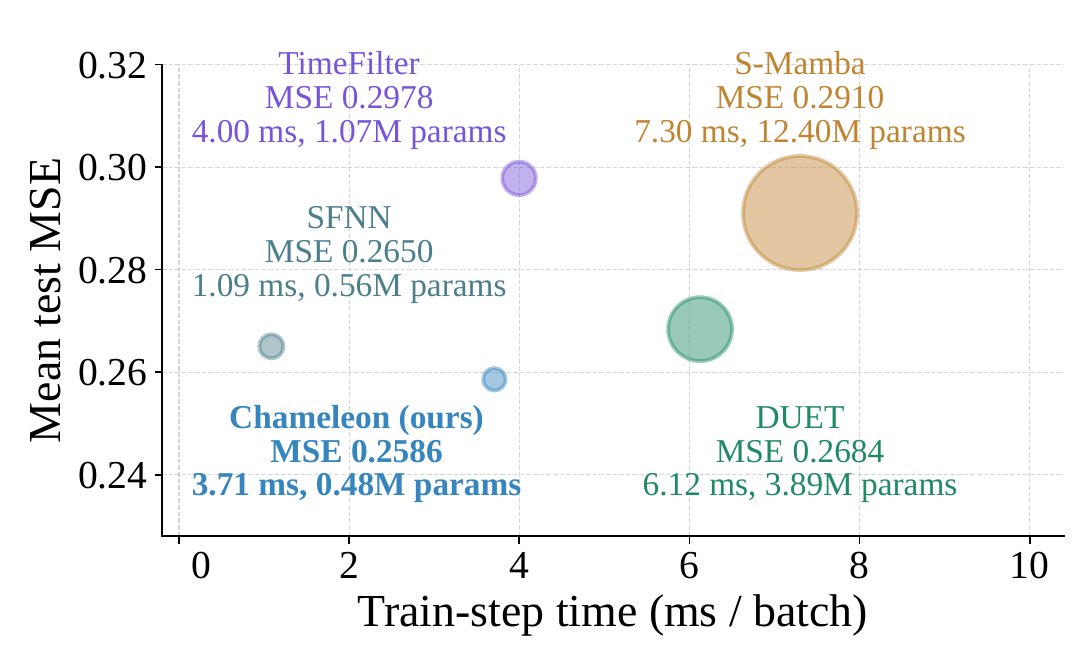}
        \captionsetup{width=0.97\linewidth}
        \caption{Test MSE versus training time, averaged across four ETT datasets.}
        \label{fig:ett_efficiency_params}
    \end{subfigure}
    \caption{Additional parameter-count analyses with $F=96$ and a batch size of one. Bubble areas are proportional to parameter counts.}
    \label{fig:param}
\end{figure*}

Figure~\ref{fig:param} complements the efficiency comparisons in Figures~\ref{fig:traffic_efficiency} and~\ref{fig:ett_efficiency} by using bubble areas to represent parameter counts instead of peak memory. Chameleon remains competitive on both Traffic and ETT, especially on the latter.

\section{Dataset Details}
\label{app:datasets}

\begin{table}[t]
\centering
\caption{Dataset descriptions. The last four are standard long-horizon benchmarks.}
\label{tab:dataset_info}
\resizebox{1.0\textwidth}{!}{
\begin{tabular}{lp{0.78\textwidth}@{\;}r}
\toprule
Dataset & Description & \\
\midrule
Double Pendulum & Simulated double pendulum dynamics with angles and angular velocities. & \citep{gilpin2021} \\
Lorenz Coupled  & Two coupled simulated Lorenz systems. & \citep{gilpin2021} \\
PEMS            & Traffic flow collected from hundreds of sensors in the Caltrans PeMS. & \citep{song2020} \\
Traffic         & Road occupancy rates from 862 sensors on Bay Area freeways. & \citep{lai2018} \\
Solar           & Solar power production from 137 photovoltaic plants in Alabama. & \citep{lai2018} \\
Electricity     & Electricity consumption of 321 clients recorded in kWh from 2012 to 2014. & \citep{lai2018} \\
ETT             & Oil temperature and six load indicators of electricity transformers. & \citep{zhou2021} \\
\bottomrule
\end{tabular}
}
\end{table}


\newlength{\freqnumwidth}
\setlength{\freqnumwidth}{1.4em}

\newlength{\frequnitwidth}
\setlength{\frequnitwidth}{2.6em}

\newlength{\freqsepwidth}
\setlength{\freqsepwidth}{0.3em}

\newlength{\freqtotalwidth}
\setlength{\freqtotalwidth}{%
  \dimexpr\freqnumwidth+\freqsepwidth+\frequnitwidth\relax
}

\newcommand{\freq}[2]{%
  \makebox[\freqnumwidth][r]{#1}%
  \makebox[\freqsepwidth][c]{}%
  \makebox[\frequnitwidth][l]{#2}%
}

\newcommand{\freqna}{%
  \makebox[\freqtotalwidth][c]{N/A}%
}

\begin{table}[t]
\centering
\caption{Experimental setup. Split: train/validation/test ratio by time; Var.: number of variables; Freq.: sampling frequency.}
\label{tab:setup}

\resizebox{1.0\textwidth}{!}{%
\begin{tabular}{l c c c r r r}
\toprule

Dataset
& Total Length
& Split
& Freq.
& Var.
& Look-back Window Size ($T$)
& Horizon ($F$) \\

\midrule

Double Pendulum
& 60,000
& 7:1:2
& \freqna
& 4
& \{96, 192, 336\}
& \{96, 192\} \\

Lorenz Coupled
& 60,000
& 7:1:2
& \freqna
& 6
& \{96, 192, 336\}
& \{48, 96\} \\

PEMS03
& 26,208
& 6:2:2
& \freq{5}{min}
& 358
& \{96, 288\}
& \{48, 96\} \\

PEMS08
& 17,856
& 6:2:2
& \freq{5}{min}
& 170
& \{96, 288\}
& \{48, 96\} \\

Traffic
& 17,544
& 7:1:2
& \freq{1}{hour}
& 862
& \{168, 336, 672, 1344\}
& \{96, 192, 336, 720\} \\

Solar
& 52,560
& 7:1:2
& \freq{10}{min}
& 137
& \{144, 288, 576, 1008\}
& \{96, 192, 336, 720\} \\

Electricity
& 26,304
& 7:1:2
& \freq{1}{hour}
& 321
& \{168, 336, 672, 1344\}
& \{96, 192, 336, 720\} \\

ETTh
& 14,400
& 6:2:2
& \freq{1}{hour}
& 7
& \{168, 336, 672, 1344\}
& \{96, 192, 336, 720\} \\

ETTm
& 57,600
& 6:2:2
& \freq{15}{min}
& 7
& \{96, 192, 384, 672, 1344\}
& \{96, 192, 336, 720\} \\

\bottomrule
\end{tabular}%
}
\end{table}

Tables~\ref{tab:dataset_info} and~\ref{tab:setup} summarize the datasets and experimental settings, respectively. The $T$ grids are based on \citet{sun2025} for the benchmarks and \citet{abdelmalak2025} for the ODE datasets. For the latter, we omit $T\in\{512,720\}$ because neither was selected in our pilot experiments for Chameleon and the tested baselines.

\begin{figure}[t]
    \centering
    \includegraphics[width=\textwidth]{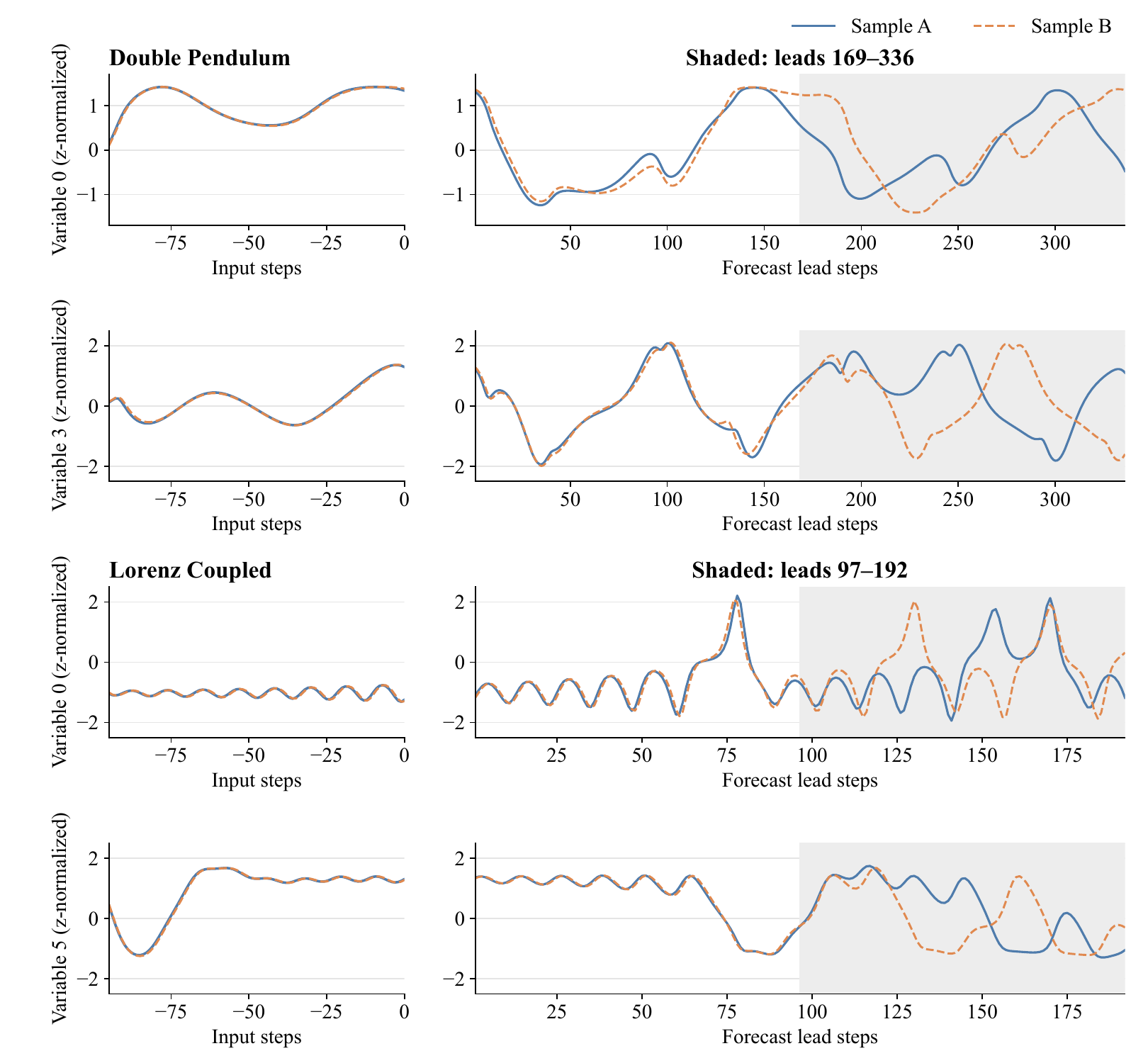}
    \caption{Nearly identical input histories with suddenly diverging futures in Double Pendulum and Lorenz Coupled. Each dataset is illustrated using its first and last variables. The left column shows 96-step input histories, and the right column shows the corresponding future trajectories, with the latter half of the forecast intervals shaded. Values are z-normalized using training-set statistics.}
    \label{fig:ode_divergent_futures}
\end{figure}

For the forecast horizons, we use shorter ranges on the chaotic ODE datasets because nearly identical input histories can lead to similar near-term trajectories that diverge sharply at longer horizons. Figure~\ref{fig:ode_divergent_futures} presents samples from Double Pendulum and Lorenz Coupled where trajectories separate sharply, particularly beyond 192 and 96 steps ahead, respectively. These examples reflect a prediction difficulty inherent in the underlying dynamics rather than specific to Chameleon. Consistently, \citet{abdelmalak2025} report normalized MSE approaching one across competitive methods on both datasets with $F=720$. These observations motivate our focus on horizons where forecasts remain informative and the benefits of cross-variable modeling can be meaningfully assessed.

Publicly released training configurations are available for most baselines on the standard benchmarks and PEMS. When no configuration is available, primarily for the two ODE datasets, we align shared dimensions with Chameleon to keep computational costs comparable. We tune the remaining hyperparameters using validation loss, considering each baseline's defaults and values commonly used on datasets with similar cross-variable dependency strengths.

\section{Preprocessing Guidance for Reversible Instance Normalization and Decomposition}
\label{app:preprocessing_guidance}



\definecolor{diaghighlightcolor}{HTML}{9ECED0}
\colorlet{diagbestlight}{diaghighlightcolor!90}

\newcommand{\diagbest}[1]{%
  {\setlength{\fboxsep}{3pt}%
  \colorbox{diagbestlight}{\strut #1}}%
}


\begin{table}[h]
\centering
\small
\caption{Trend and distribution-shift diagnostics computed from the training and validation splits. All values report the 95th percentile across variables. Values exceeding the example thresholds (Trend/Mean Shift/KS $>0.2$; Std. Shift $>1.2$) are highlighted.}
\label{tab:preprocessing_diagnostics}

\begin{tabular}{l|cccc}
\toprule
Dataset & Trend & Mean Shift & Std. Shift & KS \\
\midrule

Double Pendulum
& 0.058
& 0.093
& 1.105
& 0.074 \\

Lorenz Coupled
& 0.089
& 0.076
& 1.010
& 0.043 \\

PEMS03
& \diagbest{0.471}
& \diagbest{0.213}
& 1.173
& 0.174 \\

PEMS08
& \diagbest{0.815}
& \diagbest{0.467}
& \diagbest{1.283}
& \diagbest{0.270} \\

Traffic
& \diagbest{0.782}
& \diagbest{0.507}
& \diagbest{1.604}
& \diagbest{0.303} \\

Solar
& \diagbest{0.222}
& 0.056
& 1.058
& 0.072 \\

Electricity
& \diagbest{1.271}
& \diagbest{0.633}
& \diagbest{1.823}
& \diagbest{0.381} \\

ETTh1
& \diagbest{1.427}
& \diagbest{0.498}
& \diagbest{1.754}
& 0.182 \\

ETTh2
& \diagbest{2.317}
& \diagbest{1.465}
& \diagbest{5.486}
& \diagbest{0.685} \\

ETTm1
& \diagbest{1.427}
& \diagbest{0.478}
& \diagbest{1.746}
& 0.183 \\

ETTm2
& \diagbest{2.317}
& \diagbest{1.465}
& \diagbest{5.512}
& \diagbest{0.689} \\

\bottomrule
\end{tabular}
\end{table}
\definecolor{highlightcolor}{HTML}{9ECED0}
\colorlet{bestlight}{highlightcolor!80}

\begin{table}[t]
\centering
\caption{RevIN ablation results with $F=96$. Better results in each column are highlighted.}
\label{tab:revin_ablation}
\small
\setlength{\tabcolsep}{3pt}
\begin{tabular}{ccccccccc}
\toprule
\multicolumn{1}{c}{Dataset}
& \multicolumn{2}{c}{ETTh2}
& \multicolumn{2}{c}{ETTm2}
& \multicolumn{2}{c}{Solar}
& \multicolumn{2}{c}{Double Pendulum} \\
\cmidrule(lr){1-1}
\cmidrule(lr){2-3}
\cmidrule(lr){4-5}
\cmidrule(lr){6-7}
\cmidrule(lr){8-9}
RevIN
& MSE & MAE
& MSE & MAE
& MSE & MAE
& MSE & MAE \\
\midrule
Yes
& \colorbox{bestlight}{\strut 0.2628}
& \colorbox{bestlight}{\strut 0.3271}
& \colorbox{bestlight}{\strut 0.1528}
& \colorbox{bestlight}{\strut 0.2409}
& \colorbox{bestlight}{\strut 0.1534}
& \colorbox{bestlight}{\strut 0.1981}
& 0.1158
& 0.1753 \\
No
& 0.2968
& 0.3623
& 0.1695
& 0.2671
& 0.1725
& 0.2110
& \colorbox{bestlight}{\strut 0.0678}
& \colorbox{bestlight}{\strut 0.1176} \\
\bottomrule
\end{tabular}
\end{table}
This section provides practical guidance for deciding whether to apply RevIN and decomposition to a given dataset. RevIN improves robustness to temporal distribution shifts in level and scale, while decomposition can preserve prominent trend structure. We select their use by comparing validation loss with and without each step. To offer interpretable, useful preliminary guidance on new datasets, we also examine four diagnostics using only the training and validation splits. For each variable $i$, the Trend diagnostic is computed by fitting an ordinary least squares line over the combined training and validation segment:

\begin{equation}
    \mathrm{Trend}_i
    =
    \frac{\left|\hat{b}_i\right|\left(T_{\mathrm{tv}}-1\right)}{\sigma_i},
\end{equation}
where $\hat{b}_i$ is the fitted slope, $T_{\mathrm{tv}}$ is the number of time steps in the combined segment, and $\sigma_i$ is its standard deviation. This measures the magnitude of the fitted net change in units of the variable's own scale. To capture scale changes between splits symmetrically, we define:
\begin{equation}
    \mathrm{Std.\ Shift}_i
    =
    \max\left(
    \frac{\sigma_{i,\mathrm{val}}}{\sigma_{i,\mathrm{train}}},
    \frac{\sigma_{i,\mathrm{train}}}{\sigma_{i,\mathrm{val}}}
    \right),
\end{equation}
where $\sigma_{i,\mathrm{train}}$ and $\sigma_{i,\mathrm{val}}$ are the corresponding standard deviations, and a value of one indicates no scale change. Mean Shift measures the change in level as
\begin{equation}
    \mathrm{Mean\ Shift}_i
    =
    \frac{\left|\mu_{i,\mathrm{val}}-\mu_{i,\mathrm{train}}\right|}
    {\sigma_{i,\mathrm{train}}},
\end{equation}
where $\mu_{i,\mathrm{train}}$ and $\mu_{i,\mathrm{val}}$ are the corresponding means. Finally, $\mathrm{KS}_i$ denotes the Kolmogorov--Smirnov statistic between the training and validation marginal distributions as a broader measure of distribution shift. Each diagnostic is computed per variable, and the 95th percentile across variables is reported for each dataset.

The resulting diagnostics can help identify when each preprocessing step may be useful. For example, values above $0.2$ for Trend, Mean Shift, and KS, or above $1.2$ for Std. Shift can be treated as noticeable. Applying RevIN when any threshold is exceeded and decomposition when Trend exceeds $0.2$ gives recommendations consistent with our validation-based choices. These thresholds serve as illustrative guidance rather than universal rules, and direct validation comparisons remain more reliable, especially for borderline cases.

Table~\ref{tab:preprocessing_diagnostics} shows that Double Pendulum and Lorenz Coupled are the only datasets below all four thresholds, consistent with our validation-based decisions to omit both preprocessing steps for these datasets and retain them elsewhere. The ablation with $F=96$ in Table~\ref{tab:revin_ablation} further supports the RevIN choices. Specifically, RevIN improves both MSE and MAE on ETTh2, ETTm2, and Solar, while disabling it substantially improves both metrics on Double Pendulum. The decomposition ablation is already provided in Table~\ref{tab:ablation_main}.

\section{Periodicity Analyses}
\label{app:periodicity}

\begin{figure}[htbp]
    \centering
    \includegraphics[width=0.90\textwidth]{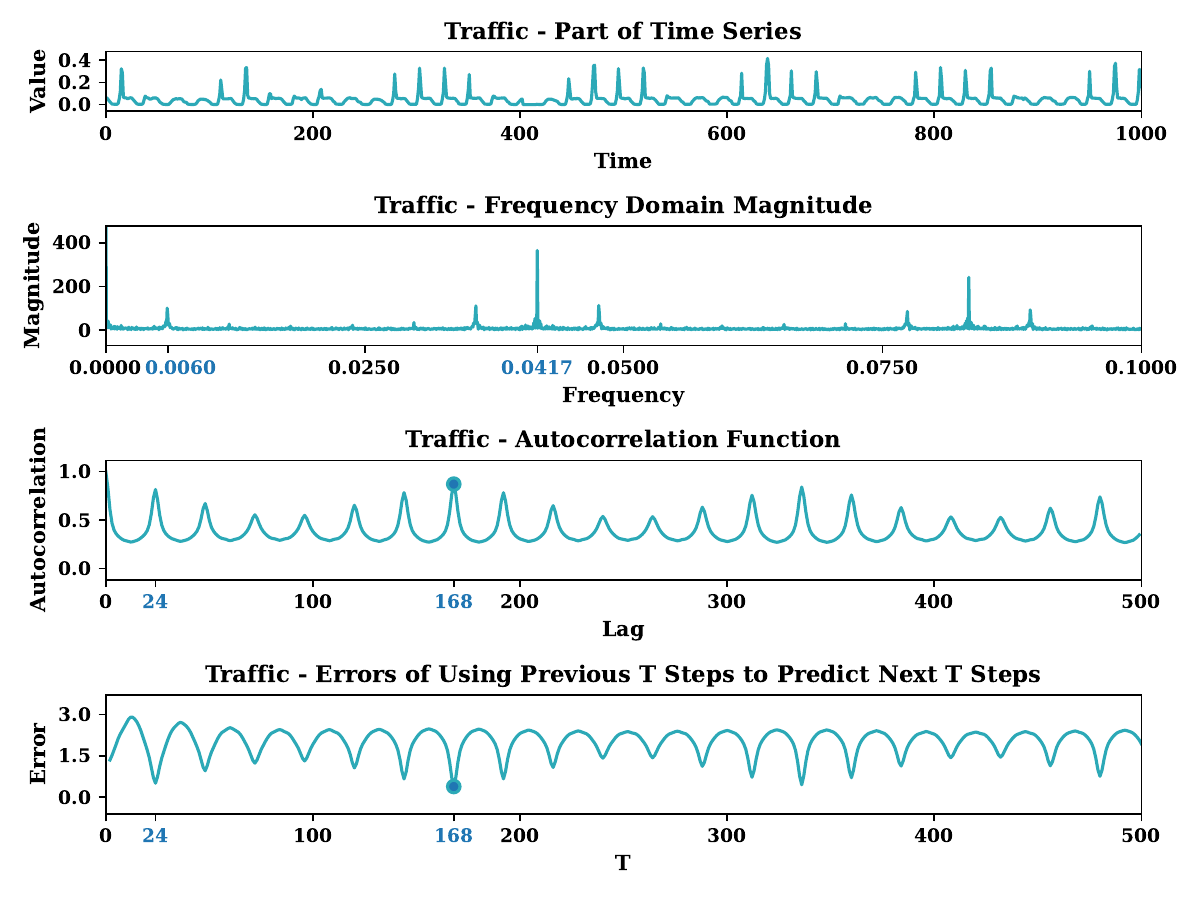}
    \caption{Traffic periodicity analysis, identifying 168 time steps (one week) as the strongest period.}
    \label{fig:traffic}
\end{figure}

\begin{figure}[htbp]
    \centering
    \includegraphics[width=0.90\textwidth]{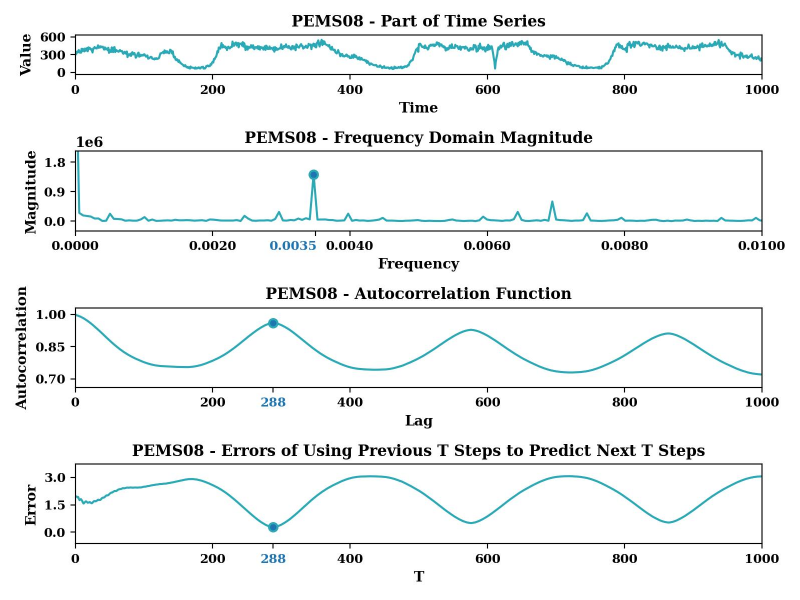}
    \caption{PEMS08 periodicity analysis, identifying 288 time steps (one day) as the strongest period.}
    \label{fig:pems08}
\end{figure}

\begin{figure}[htbp]
    \centering
    \includegraphics[width=0.90\textwidth]{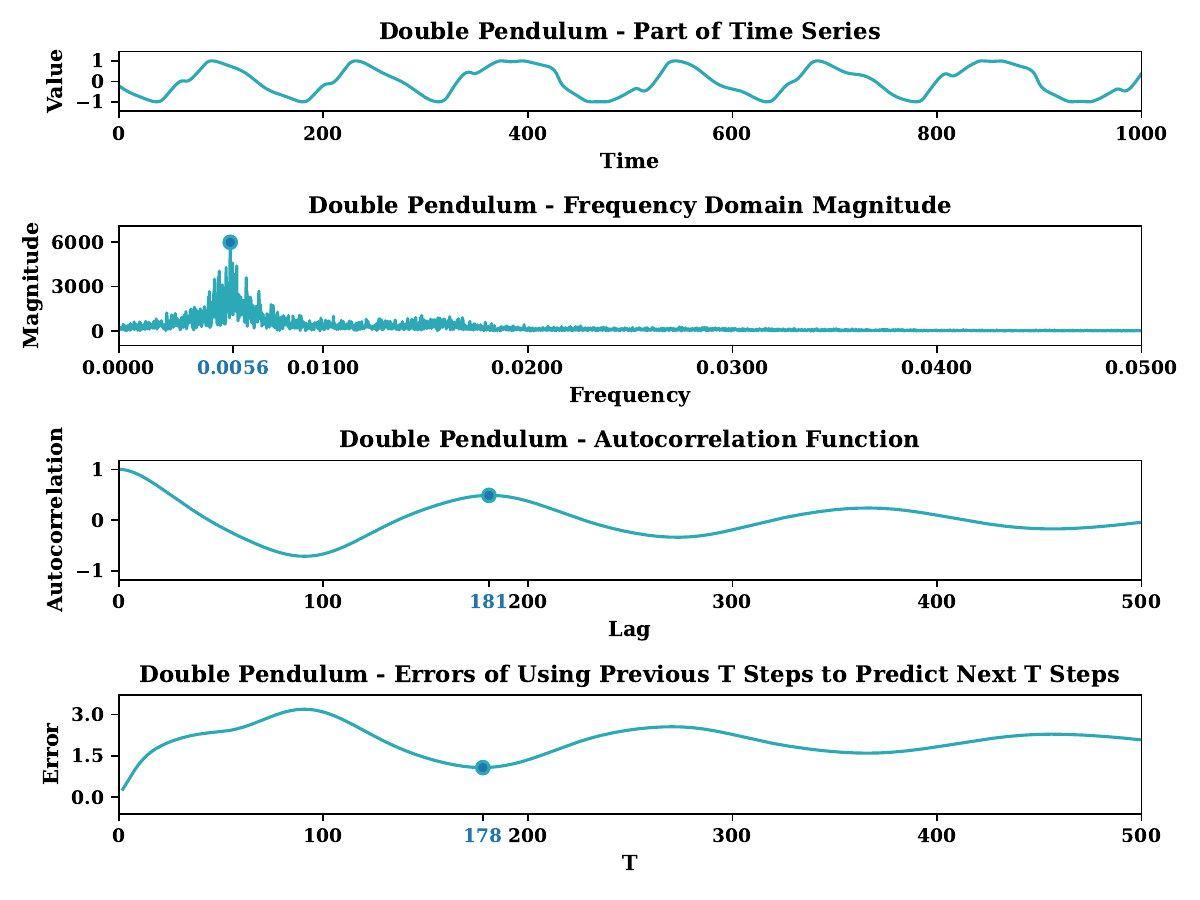}
    \caption{Double Pendulum periodicity analysis. No clear dominant period is identified.}
    \label{fig:dp}
\end{figure}

\begin{figure}[htbp]
    \centering
    \includegraphics[width=0.90\textwidth]{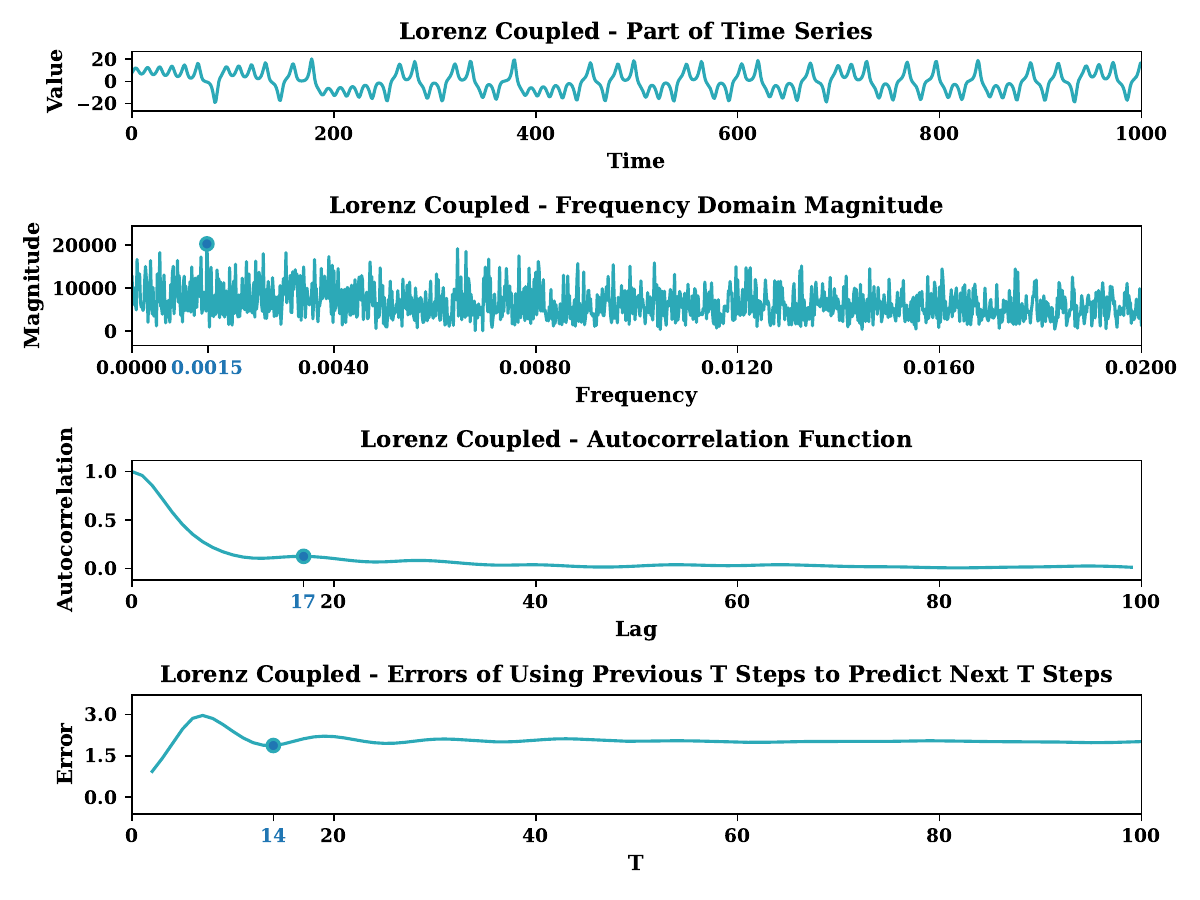}
    \caption{Lorenz Coupled periodicity analysis. No clear dominant period is identified.}
    \label{fig:lorenz_periodicity}
\end{figure}

We analyze periodicity in Traffic, PEMS08, Double Pendulum, and Lorenz Coupled~\citep{lai2018,song2020,gilpin2021} using Fourier transforms, autocorrelation, and lag-copy tests. Traffic and PEMS08 exhibit dominant periods of 168 and 288 time steps, corresponding to one week and one day, respectively. Both exceed the conventional look-back window of 96 steps, supporting evaluation with larger $T$ to capture this periodic context. In contrast, neither chaotic ODE dataset exhibits a clear dominant period. Together with their strong and distinct cross-variable dependencies, this contrasting temporal behavior makes them informative complements to the standard benchmarks.

\end{document}